\pdfoutput=1
\documentclass[11pt]{article}

\usepackage[preprint]{acl}

\usepackage{times}
\usepackage{latexsym}

\usepackage[T1]{fontenc}
\usepackage[utf8]{inputenc}

\usepackage{microtype}

\usepackage{inconsolata}

\usepackage{graphicx}

\usepackage{amsmath,amsfonts,bm}

\def\eqref#1{equation~\ref{#1}}
\def\1{\bm{1}}

\DeclareMathAlphabet{\mathsfit}{\encodingdefault}{\sfdefault}{m}{sl}
\SetMathAlphabet{\mathsfit}{bold}{\encodingdefault}{\sfdefault}{bx}{n}

\usepackage{soul}
\usepackage{url}
\usepackage[utf8]{inputenc} 
\usepackage{caption}
\usepackage{graphicx}
\usepackage{amsmath}
\usepackage{hyperref}
\usepackage[capitalize,noabbrev]{cleveref}
\crefname{equation}{Eq.}{Eqs.}
\Crefname{equation}{Eq.}{Eqs.}
\makeatletter
\AddToHook{cmd/appendix/before}{\def\cref@section@alias{appendix}}
\AddToHook{cmd/appendix/before}{\def\cref@subsection@alias{appendix}}
\makeatother

\usepackage{amsthm}
\newtheorem{theorem}{Theorem}

\usepackage{booktabs}
\usepackage{algorithm}
\usepackage{algorithmicx}
\usepackage{algpseudocode}
\usepackage[switch]{lineno}

\usepackage{amssymb}
\usepackage{array}
\usepackage{tabularx}
\usepackage{setspace}
\usepackage{enumitem}

\usepackage{graphicx}
\usepackage{float} 
\usepackage{adjustbox}
\usepackage{subcaption}
\usepackage{svg}

\usepackage{float}
\usepackage{multirow}
\usepackage{wrapfig}
\usepackage{colortbl}
\usepackage{makecell}
\usepackage{bbding}
\usepackage{placeins}

\usepackage[most]{tcolorbox}
\usepackage{xcolor}
\usepackage{listings}

\lstdefinestyle{promptbox}{
  basicstyle=\ttfamily\small,
  breaklines=true,
  breakatwhitespace=true,
  columns=fullflexible,
  frame=none,
  showstringspaces=false
}

\newcommand{\boxtoken}[1]{\textcolor{purple!80!black}{\texttt{#1}}}
\newcommand{\placeholder}[1]{\textcolor{teal!70!black}{\texttt{\{#1\}}}}

\makeatletter
\AddToHook{cmd/appendix/before}{\def\cref@section@alias{appendix}}
\AddToHook{cmd/appendix/before}{\def\cref@subsection@alias{appendix}}
\makeatother

\usepackage{tikz}
\usetikzlibrary{shapes, arrows.meta, positioning, calc, backgrounds, fit, shadows}

\definecolor{aclblue}{RGB}{78, 121, 167}
\definecolor{aclorange}{RGB}{242, 142, 43}
\definecolor{aclgrey}{RGB}{110, 110, 110}
\definecolor{bgblue}{RGB}{235, 242, 250}
\definecolor{bgorange}{RGB}{255, 245, 235}
\definecolor{categoryblue}{RGB}{217, 234, 250}
\definecolor{avgyellow}{RGB}{255, 255, 204}
\definecolor{graybg}{rgb}{0.9,0.9,0.9}

\newcommand\blfootnote[1]{\begingroup\renewcommand\thefootnote{}\footnote{#1}\addtocounter{footnote}{-1}\endgroup}

\title{From Off-Policy to On-Policy: Enhancing GUI Agents via Bi-level Expert-to-Policy Assimilation}

\author{Zezhou Wang\textsuperscript{1$\ddagger$} \quad Ziyun Zhang\textsuperscript{2} \quad Xiaoyi Zhang\textsuperscript{3$\dagger$} \\
\textbf{Zhuzhong Qian\textsuperscript{1}} \quad \textbf{Yan Lu\textsuperscript{3}} \vspace{2mm} \\
$^1$Nanjing University \quad $^2$Peking University \\
$^3$Microsoft Research Asia
}

\begin{document}
\maketitle
% \begingroup
% \renewcommand\thefootnote{$*$}
% \footnotetext{Corresponding author.}
% \endgroup
\blfootnote{$^{\ddagger}$: Work is done during internship at Microsoft Research Asia. $^{\dagger}$: Project lead. }

\begin{abstract}
Vision-language models are increasingly deployed as computer-use agents (CUAs) that operate desktops and browsers. Top-performing CUAs are framework-based systems that decompose planning and execution, while end-to-end screenshot-to-action policies are easier to deploy but lag behind on benchmarks such as OSWorld-Verified. GUI datasets like OSWorld pose two bottlenecks: they expose only a few hundred interactive, verifiable tasks and environments, and expert trajectories must be gathered by interacting with these environments, making such data hard to scale. We therefore ask how reinforcement learning from verifiable rewards (RLVR) can best exploit a small pool of exist expert trajectories to train end-to-end policies. Na\"ively mixing these off-policy traces into on-policy RLVR is brittle: even after format conversion, expert trajectories exhibit structural mismatch and distribution shift from the learner. We propose \textbf{BEPA} (\textbf{Bi}-Level \textbf{E}xpert-to-\textbf{P}olicy \textbf{A}ssimilation), which turns static expert traces into policy-aligned guidance via self-rolled reachable trajectories under the base policy (LEVEL-1) and a per-task, dynamically updated cache used in RLVR (LEVEL-2). On OSWorld-Verified, BEPA improves UITARS1.5-7B success from 22.87\% to 32.13\% and raises a held-out split from 5.74\% to 10.30\%, with consistent gains on MMBench-GUI and Online-Mind2Web. Our code and data are available at: \url{https://github.com/LEON-gittech/Verl_GUI.git}.
\end{abstract}

\section{Introduction}

GUI agents aim to solve realistic computer-use tasks---from web navigation to full desktop control---under multimodal, long-horizon interactions \cite{Wang2024GUIAgentswithFoundationModels,Nguyen2024GUIAgentSurvey,Zhang2024LLMBrainedGUIAgents,miniweb++,Zhou2023WebArenaAR,Xie2024OSWorldBM,Deng2023Mind2WebTA}. Recent progress largely follows two paradigms. \textbf{Framework-based} systems wrap an LLM with a compositional stack of planners, tools, and executors \cite{Agashe2024AgentSA,xie2025jedi}, multi-agent collaboration \cite{ye2025mobileagentv3fundamentalagentsgui}, and explicit tool/code execution modules \cite{Song2025CoAct1CA}. These frameworks can be highly effective: on OSWorld-Verified, framework-based agents such as Agent S2 and Jedi-7B w/ o3 achieve 33.00--42.40\% success, whereas specialized end-to-end (E2E) policies such as UITARS1.5-7B and OpenCUA-7B are around 23--24\% \cref{tab:osworld}. In contrast, \textbf{E2E} agents train a single policy that maps screenshots and instructions directly to low-level actions \cite{Wang2025OpenCUAOF,Wang2025UITARS2TR}. UI-TARS \cite{Qin2025UITARSPA} develops a native E2E GUI agent via multi-stage post-training on perception, unified action modeling/grounding, and reasoning, and OpenCUA \cite{Wang2025OpenCUAOF} releases an open framework and supervised pipeline that converts human demonstrations into state--action pairs with reflective chain-of-thought for training end-to-end computer-use agents. Despite this progress, E2E policies still lag behind strong framework-based systems on challenging benchmarks such as OSWorld-Verified \cite{Xie2024OSWorldBM}.

Unlike text-only instruction following, where synthetic data can be scaled via Self-Instruct-style generation, GUI datasets and benchmarks such as OSWorld-Verified pose two bottlenecks: they expose only a few hundred interactive, verifiable tasks and environments, and expert trajectories must be gathered by interacting with these environments, typically by running strong framework-based agents and verifying each episode. This makes both tasks and demonstrations hard to scale and yields only a modest pool of high-quality expert traces. This raises a complementary question: \textit{given such a small but valuable pool of expert framework trajectories, how can Reinforcement Learning with Verifiable Rewards (RLVR) best exploit them to train end-to-end policies?}

A natural direction is to use expert executions as off-policy guidance in RLVR. However, directly mixing such traces into learning is brittle. The core difficulty is twofold: a \textbf{structural mismatch} and a \textbf{distribution gap}. Structurally, framework traces interleave multiple roles (planning, execution, and grounding) and often operate in tool-level action spaces (e.g., APIs) that a single end-to-end policy cannot directly imitate. Distributionally, even after format conversion, the resulting trajectories can lie far from the base-policy manifold (\cref{fig:distribution_bias}). In RLVR, where optimization relies on on-policy rollouts and trust-region style updates \cite{Schulman2015TRPO}, this mismatch can cause exploration collapse or unstable optimization. Mixed on-/off-policy training has been explored \cite{Xiong2023IterativePL,Yan2025luffy,zhang2025bread}; LUFFY \cite{Yan2025luffy} instantiates this with mixed-policy GRPO \cite{DeepSeekR1IR} and policy shaping on \emph{raw} off-policy traces, while BREAD \cite{zhang2025bread} introduces expert anchors and branched rollouts that let the student self-roll from expert prefixes to alleviate sparse rewards and distribution shift in textual reasoning. Yet these approaches still treat expert trajectories as largely static guidance defined in the expert’s own action space; Empirically, such static integration strategies do not close the gap on OSWorld-Verified and can even degrade UITARS1.5-7B compared to pure GRPO \cref{tab:main_table}.

We propose \textbf{BEPA} (\textbf{Bi}-Level \textbf{E}xpert-to-\textbf{P}olicy \textbf{A}ssimilation), a plug-and-play RLVR component that turns static, mismatched expert traces into dynamic, policy-aligned guidance (\cref{fig:overview,sec:method}). BEPA operates in two stages: \textbf{LEVEL-1} re-rolls expert plans under the base policy to produce \emph{reachable} trajectories that seed an off-policy cache, and \textbf{LEVEL-2} continuously refreshes this cache using the agent's own emerging successes, injecting cached guidance only upon total exploration failure. Integrated with GRPO in \cref{sec:grpo-integration}, this design lets the agent learn primarily from on-policy exploration while using expert traces as a self-aligned scaffold. The contributions are as follows:
\begin{itemize}[noitemsep,topsep=0pt]
    \item We identify structural and distributional expert-to-policy mismatches in GUI agents: expert trajectories are not directly learnable by the base policy, and naive conversion remains distributionally biased (\cref{fig:distribution_bias,sec:preliminaries}).
    \item We propose \textbf{BEPA}, a bi-level assimilation framework that bootstraps with policy-reachable guidance and maintains dynamic alignment via a self-updating cache integrated into GRPO (\cref{fig:overview,sec:method}).
    \item We demonstrate the effectiveness of BEPA on three benchmarks. On OSWorld-Verified, BEPA improves UITARS1.5-7B from 22.87\% to 32.13\% overall success and from 5.74\% to 10.30\% on a strictly held-out split, while achieving consistent improvements on MMBench-GUI and Online-Mind2Web \cref{tab:main_table,tab:L1,tab:L2,tab:mind2web}. We further provide mechanism and sensitivity analyses linking performance to distribution alignment and optimization dynamics (\cref{subsec:mechanism_analysis,sec:sensitivity,fig:histogram,fig:training_dynamics}).
\end{itemize}

\section{Related Work}\label{sec:related_work}

\subsection{GUI Agents}
GUI agents aim to solve realistic computer-use tasks---from mobile use, web navigation to full desktop control---under multimodal, long-horizon settings \cite{miniweb++,Zhou2023WebArenaAR,Xie2024OSWorldBM,Deng2023Mind2WebTA}. In this paper, we mainly focus on full desktop control scenario. Existing systems largely fall into two paradigms: \textbf{framework-based agents} and \textbf{end-to-end agents}.

\paragraph{Framework-based agents.}
Framework-based agents wrap a LLM with structured control, including hierarchical planner--executor designs \cite{Agashe2024AgentSA,xie2025jedi}, multi-agent collaboration \cite{ye2025mobileagentv3fundamentalagentsgui}, and explicit tool/code execution modules \cite{Song2025CoAct1CA}. Such systems can generate high-quality, executable trajectories, but their outputs are typically produced by multiple specialized roles and interfaces, making them mismatched to a single end-to-end policy.

\paragraph{End-to-end agents.}
End-to-end GUI agents instead train one policy to map observations and instructions directly to low-level actions \cite{Wang2025OpenCUAOF,Qin2025UITARSPA,Wang2025UITARS2TR,gao2024assistgui,gou2025uground}. UI-TARS \cite{Qin2025UITARSPA} develops a native end-to-end GUI agent via multi-stage post-training on perception, unified action modeling/grounding, and reasoning. ARPO \cite{lu2025ARPO} performs end-to-end reinforcement learning with a replay buffer to reuse successful experiences across training iterations. OpenCUA \cite{Wang2025OpenCUAOF} instead focuses on the data and system stack for E2E CUAs, providing an open-source framework with annotation tooling and a supervised pipeline that turns recorded human computer-use demonstrations into training data for screenshot-to-action policies. Despite progress, current E2E agents remain far behind the strongest framework-based systems. This substantial gap---in contrast to the marginal improvements often seen in text-only reasoning benchmarks---makes it meaningful to study how to learn from off-policy expert traces under severe distribution shift, which is precisely the regime targeted by our expert-to-policy assimilation.

\subsection{Reinforcement Learning for LLM Agents}
Reinforcement learning is widely used to improve LLM agents beyond supervised imitation and preference optimization \cite{Ziegler2019FineTuningLMPreferneces,Rafailov2023DirectPO,Ouyang2022TrainingLM,Hu2025OpenReasonerZeroAO}. Classical RLHF often relies on PPO \cite{Schulman2017ProximalPO}, while recent work proposes critic-free, group-based objectives such as GRPO \cite{Shao2024DeepSeekMathPT,DeepSeekR1IR} and its variants \cite{ahmadian2024rloo,liu2025Dr.GRPO,Yu2025DAPOAO,chen2025loop}. Beyond purely on-policy RL, there is growing interest in incorporating stronger experts or offline data: iterative and hybrid preference learning under KL constraints is studied in \cite{Xiong2023IterativePL}; LUFFY \cite{Yan2025luffy} augments GRPO with off-policy teacher rollouts and a mixed-policy objective; and BREAD \cite{zhang2025bread} uses expert anchors and branched rollouts, letting the student self-roll from short expert prefixes to ease sparse rewards and distribution shift in textual reasoning. These methods show that expert-guided RL can be effective when teacher and student operate in a similar action space. In GUI agents, however, the expert action spaces and trajectory distributions could differ substantially from the base policy. BEPA re-rolls expert guidance into policy-reachable trajectories and maintains a dynamically aligned cache within RLVR, enabling stable expert assimilation under such structural and distributional mismatch.

\section{Preliminaries}\label{sec:preliminaries}

Following the standard end-to-end GUI agent formulation \cite{Qin2025UITARSPA,Wang2025UITARS2TR}, we consider a multi-step decision-making setting where an agent interacts with a GUI environment to complete a task specified by a natural-language instruction $x$. At each step $t$, the agent observes a screenshot $s_t \in \mathcal{S}$ and produces a \emph{textual action trace} $a_t \in \mathcal{V}^n$, generated autoregressively by the policy $\pi_\theta(a_t \mid s_t, x)$. Unlike framework-based agents that separate high-level reasoning from low-level grounding, end-to-end GUI agents emit the \emph{entire} reasoning and grounding sequence as a unified textual output. The action space details are provided in \cref{app:action_space}. The environment executes $a_t$ and transitions to $s_{t+1}$. Crucially, we assume access to a deterministic \textbf{verifier} function $R(\tau) \in \{0, 1\}$ that evaluates the correctness of the final system state. Given an instruction $x$, an episode trajectory is defined as $\tau = (x, (s_t, a_t)_{t=1}^{T})$. The sparse binary reward is typically computed at termination: $r_T = R(\tau)$, with $r_t=0$ for $t < T$.

\section{Bi-Level Expert-to-Policy Assimilation}
\label{sec:method}

\begin{figure*}[t]
    \centering
    \includegraphics[width=0.9\textwidth]{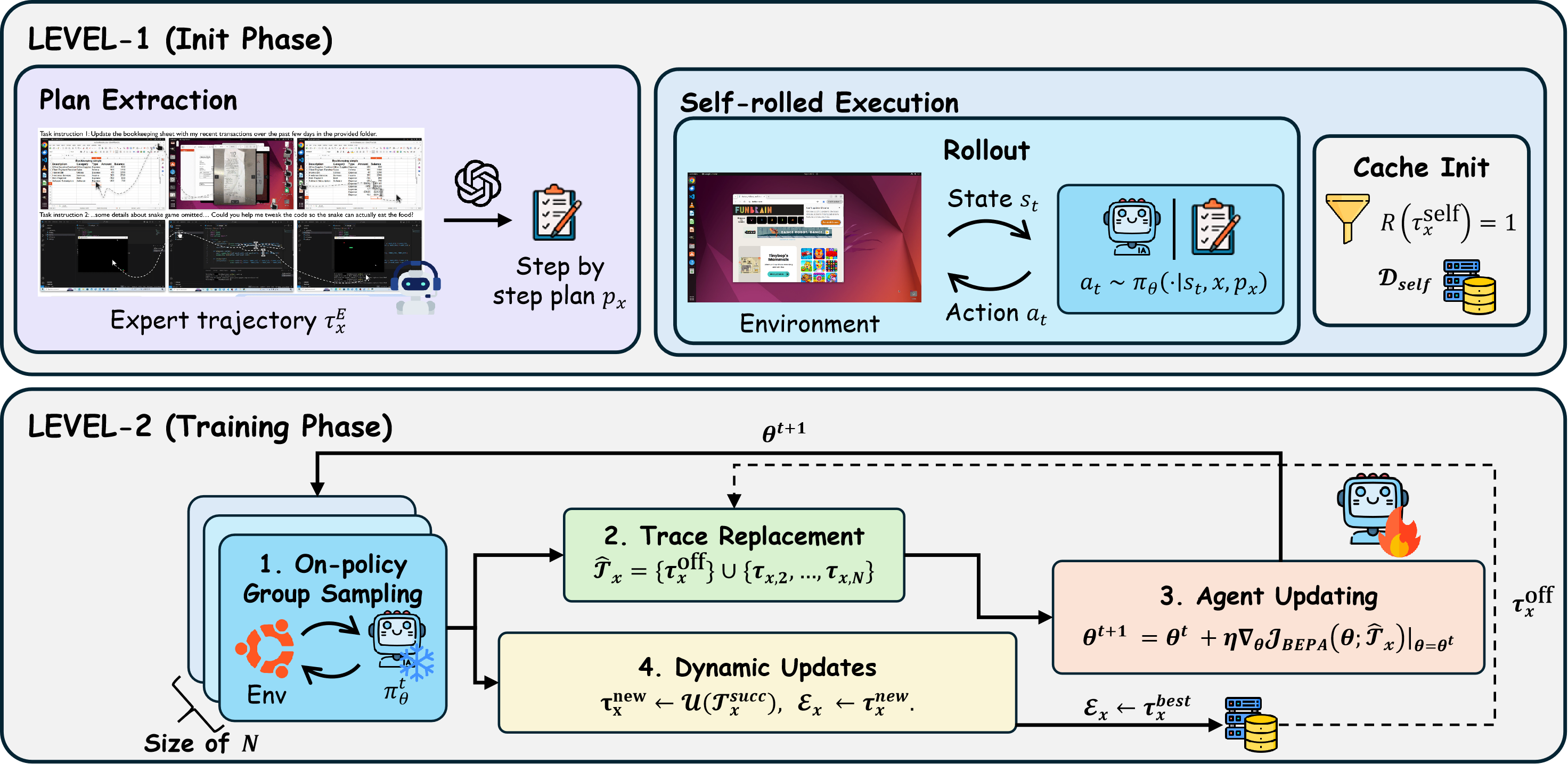}
    \caption{\textbf{BEPA overview.} We exploit strong but mismatched experts via two modular, plug-and-play stages. \textbf{LEVEL-1} initializes a policy-compatible guidance seed by re-rolling expert plans under the base policy. \textbf{LEVEL-2} maintains a self-aligned per-task cache using the agent's own emerging successes, keeping the off-policy guidance within a controllable distribution gap relative to the evolving on-policy manifold. The cached guidance is injected into GRPO only upon total on-policy failure (i.e., when all rollouts in a group fail).}

    \label{fig:overview}
    \vspace{-4mm}
\end{figure*}

\begin{figure*}[t]
    \centering
    \includegraphics[width=0.9\textwidth]{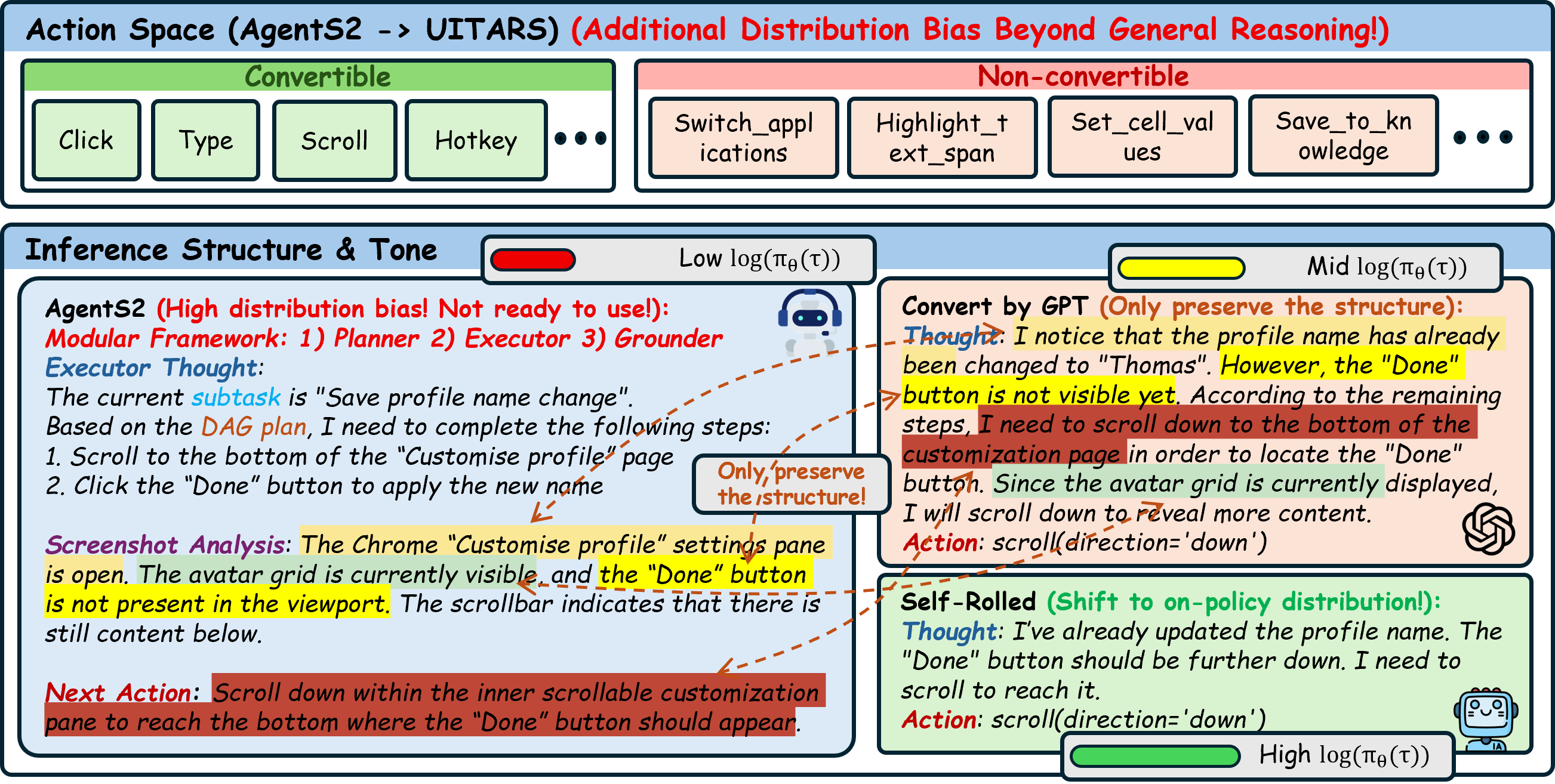}
    \caption{\textbf{Distribution bias from framework experts to end-to-end GUI policies.}
\textbf{Top: action-space mismatch} (Agent S2 includes non-convertible framework actions beyond UI-TARS primitives).
\textbf{Bottom: inference structure mismatch.} Agent S2 traces (left) interleave multi-role artifacts (planner/executor/grounder), yielding low likelihood under the base policy (low $\log \pi_\theta(\tau)$).
Naive conversion (middle) maps format but largely preserves the framework structure, remaining off-manifold.
Self-rolled execution (right) re-generates actions under the base policy (plan-conditioned), producing more policy-compatible, higher-likelihood trajectories.}

    \label{fig:distribution_bias}
    \vspace{-4mm}
\end{figure*}

We propose \textbf{Bi-Level Expert-to-Policy Assimilation} (BEPA), a framework designed to catalyze capability breakthroughs in end-to-end GUI agents by bridging the distributional gap between external experts and the policy's intrinsic manifold (as shown in \cref{fig:distribution_bias}). Rather than passively mixing mismatched data, BEPA utilizes off-policy traces as a structured \emph{guidance scaffold} to steer the agent beyond its initial capability frontier. As shown in \cref{fig:overview}, the framework operates in two complementary stages: \textbf{LEVEL-1: Self-Rolled Execution} transforms alien expert traces into policy-compatible trajectories to initialize a guidance pool; \textbf{LEVEL-2: Self-Aligned Off-Policy Assimilation} dynamically maintains a per-task cache, injecting these guided trajectories into GRPO updates only upon \emph{total on-policy failure} (i.e., when all rollouts in a group fail), thereby turning static expert data into an evolving, policy-aligned guidance signal.

\subsection{LEVEL-1: Self-Rolled Execution}
\label{sec:selfroll}

Directly training on distributionally mismatched expert data often leads to performance degradation due to significant covariate shift. To mitigate this, LEVEL-1 acts as a \emph{reachability (policy-manifold) adapter}: it re-executes expert solutions \emph{under the base policy} with plan conditioning, producing trajectories that are immediately executable and learnable by $\pi_\theta$.

We start from an offline expert trace set $\mathcal{D}_E = \{(x, \tau_x^{E})\}$, where $\tau_x^{E}$ is a successful trajectory from a framework-based agent or a stronger policy. For each pair $(x, \tau_x^{E})$, we first abstract the expert trajectory into a compact natural-language plan $p_x = \phi(\tau_x^{E})$ (e.g., a sequence of subgoals).
During the \emph{initialization phase}, we append $p_x$ to the instruction $x$ and let the base policy $\pi_\theta$ act in the environment: $a_t \sim \pi_\theta(\cdot \mid s_t, x, p_x)$. The environment executes $a_t$, producing a \emph{self-rolled} trajectory $\tau_x^{\text{self}}$. Trajectories that pass the OSWorld verifier ($R(\tau_x^{\text{self}})=1$) form the filtered self-rolled set:
\begin{equation}
    % \small
\mathcal{D}_{\text{self}} = \{(x, \tau_x^{\text{self}}) : R(\tau_x^{\text{self}})=1\}.
\end{equation}

The expert-derived plan $p_x$ serves as a scaffold, guiding $\pi_\theta$ to visit high-reward regions it would rarely explore autonomously. Crucially, since $\tau_x^{\text{self}}$ is generated by $\pi_\theta$ itself, it lies much closer to the policy's manifold than the original $\tau_x^{E}$.

\subsection{LEVEL-2: Off-Policy Assimilation}
\label{sec:assimilation}

While LEVEL-1 converts expert guidance into policy-reachable trajectories, the policy itself continues to evolve during RLVR. LEVEL-2 performs \emph{self-aligned off-policy assimilation} by continually refreshing the cache with the policy's own successes traces, ensuring the off-policy signal evolves alongside the agent.

\paragraph{Self-Aligned Off-Policy Cache.}
We initialize $\mathcal{E}$ from the LEVEL-1 seed set $\mathcal{D}_{\text{self}}$ by setting
\begin{equation}
    % \small
\mathcal{E}_x \triangleq \tau \quad \text{for } (x,\tau)\in\mathcal{D}_{\text{self}},
\end{equation}
and denote the cached trajectory by $\tau_x^{\text{off}} \triangleq \mathcal{E}_x$.

\paragraph{Dynamic Updates.}
During online GRPO training, at each iteration $k$, we collect a group of
$N$ on-policy rollouts $\mathcal{T}_x = \{\tau_{x,i}\}_{i=1}^N$ from
$\pi_{\theta_{\text{old}}}$.
Let $\mathcal{T}_x^{\mathrm{succ}} = \{\tau \in \mathcal{T}_x \mid R(\tau)=1\}$ denote
the subset of successful trajectories.
If the current policy succeeds on the task ($\mathcal{T}_x^{\mathrm{succ}} \neq \varnothing$),
we update the cache using an updating rule
$\mathcal{U}(\mathcal{T}_x^{\mathrm{succ}})$, setting $\tau_x^{\mathrm{new}} \gets \mathcal{U}(\mathcal{T}_x^{\mathrm{succ}})$ and 
$\mathcal{E}_x \leftarrow \tau_x^{\mathrm{new}}$. 
$\mathcal{U}(\cdot)$ is random sampling by default. In summary, LEVEL-2 keeps the guidance signal $\tau_x^{\text{off}}$ concentrated on traces that lie in high-density regions of the evolving policy $\pi_\theta$, thereby reducing covariate shift during trace replacement.

\subsection{Integration with GRPO}
\label{sec:grpo-integration}

We integrate the self-aligned cache into GRPO via conditional \emph{trace replacement}, allowing the agent to learn primarily from its own exploration and injecting off-policy guidance only upon total failure.
For mixing, cached traces are injected under the original instruction $x$ \emph{without} the step-by-step plan prefix used in LEVEL-1 self-rolling, matching the conditioning of on-policy rollouts.

At each training step, we sample a group of on-policy rollouts $\mathcal{T}_x$ (size $N$) from $\pi_{\theta_{\text{old}}}$ and evaluate them using the verifier $R(\cdot)$.
When the agent experiences \emph{total exploration failure} (i.e., $\forall \tau \in \mathcal{T}_x, R(\tau)=0$) and $\mathcal{E}_x \neq \varnothing$, we replace the first failed trajectory with the cached off-policy trajectory $\tau_x^{\text{off}}$:

\begin{equation}
    % \small
\hat{\mathcal{T}}_x = \{\tau_x^{\text{off}}\} \cup \{\tau_{x,2}, \dots, \tau_{x,N}\}.
\label{eq:Tx}
\end{equation}

This injection guarantees that even in failed batches, the optimizer receives at least one positive signal. Given the finalized group $\hat{\mathcal{T}}_x$, we compute advantages using group-wide normalization.
For $\tau_\ell \in \hat{\mathcal{T}}_x$ with reward set $\hat{G} = \{R(\tau) : \tau \in \hat{\mathcal{T}}_x\}$:

\begin{equation}
    % \small
\hat{A}_\ell = \frac{R(\tau_\ell) - \mathrm{mean}(\hat{G})}{\mathrm{std}(\hat{G}) + \epsilon}.
\label{eq:mixed-adv}
\end{equation}

The BEPA objective extends GRPO to accommodate the mixed composition of $\hat{\mathcal{T}}_x$:

\begin{equation}
    % \small
\mathcal{J}_{\mathrm{BEPA}}(\theta)
= \frac{1}{Z} \sum_{\tau \in \hat{\mathcal{T}}_x} \sum_{t=1}^{|\tau|}
L^{\mathrm{CLIP}}(r_t, \hat{A}_\tau),
\label{eq:bepa-obj}
\end{equation}

where $Z = \sum_{\tau} |\tau|$,
$L^{\mathrm{CLIP}}(r,A) = \min[rA,\ \mathrm{clip}(r; 1\pm\epsilon)A]$, and
$r_t = \pi_\theta(a_t\mid s_t) / \pi_{\theta_{\text{old}}}(a_t\mid s_t)$ is the importance ratio. The clipping empirically keeps $\pi_\theta$ within a trust region of $\pi_{\theta_{\text{old}}}$ \cite{Schulman2015TRPO}. The convergence analysis of the resulting mixed-policy GRPO objective is provided in Appendix~\ref{app:convergence_analysis}.

% Trace replacement also admits a simple advantage-space interpretation.
% In a total-failure group without replacement, rewards are all zeros and \cref{eq:mixed-adv} yields near-zero advantages for all trajectories, resulting in a degenerate update.
% With replacement, the reward multiset becomes $\hat{G}=\{1,0,\dots,0\}$ (one success, $N{-}1$ failures), so ignoring $\epsilon$ we obtain
% \begin{equation}
% \hat{A}_{\text{off}} \approx \sqrt{N-1},
% \qquad
% \hat{A}_{\text{fail}} \approx -\frac{1}{\sqrt{N-1}},
% \end{equation}
% i.e., a clear ``success vs.\ failure'' advantage gap while retaining PPO-style trust-region updates through $r_t$ in \cref{eq:bepa-obj}.

\section{Experiments}\label{sec:experiments}
 
\subsection{Experiment Setup}
\paragraph{Dataset \& Benchmarks.}
We conduct RLVR training on the OSWorld-Verified \cite{Xie2024OSWorldBM} and assess cross-domain generalization on MMBench-GUI \cite{Wang2025MMBenchGUIHM} and Online-Mind2Web \cite{deng2023mind2web,xue2025illusionprogressassessingcurrentOnline-Mind2web}. 
\textbf{OSWorld-Verified} comprises 369 real-world tasks requiring file I/O and multi-app workflows. 
Training RL agents on such complex benchmarks is often impeded by extremely sparse reward signals, as many tasks remain effectively unsolvable by current models, yielding negligible feedback for optimization. To mitigate this and ensure \textbf{informative reward signals} during exploration, we curate a subset of ``high-value'' tasks that offer verifiable success trajectories. Specifically, we identify tasks solvable by \textit{either} the external expert (Agent S2) \textit{or} the base policy (UITARS1.5-7B Pass@5), resulting in a pool of 150 tasks where valid supervision is guaranteed. From this pool, we randomly sample 80\% ($120$ tasks) and 8 other tasks to form the \textbf{Training Set} $\mathcal{D}_{\text{train}}$ (128 tasks), ensuring the model focuses on learnable behaviors rather than stalling on intractable scenarios. The remaining 20\% of the solvable pool, combined with other tasks, constitutes the \textbf{Held-out Test Set} $\mathcal{D}_{\text{held\_out}}$ ($241$ tasks), which is used to strictly evaluate the agent's generalization capability to unseen scenarios.
Detailed benchmark information is provided in the \cref{app:benchmarks}.

\paragraph{Baselines.} We evaluate our approach against a comprehensive suite of competitive baselines spanning multiple categories: 1) \textbf{closed-source LLMs} with strong reasoning capabilities, 2) \textbf{open-source GUI agents} specialized for visual interaction, and 3) \textbf{modular agent frameworks} that employ compositional strategies for computer use tasks. 4) \textbf{training methodologies}, we benchmark against GRPO~\cite{Shao2024DeepSeekMathPT} as the pure on-policy baseline, along with several \textbf{expert integration methods} that incorporate converted expert traces $\mathcal{D}_{\text{conv}}$ (the conversion prompt is in \cref{app:prompt_templates}): SFT, RL w/ SFT Loss (incorporating SFT loss during RL training), and SFT+RL (a two-stage process continuing RL after SFT). We also compare with LUFFY \cite{Yan2025luffy}, which augments RLVR with \emph{raw off-policy reasoning traces} via Mixed-Policy GRPO and policy shaping, and Trace Replacement, which differs from LUFFY by replacing trajectories with off-policy traces only upon group failure and calculating the importance ratio using the old policy likelihood rather than a constant (BEPA is based on the trace replacement). See more details in \cref{app:baselines}.

\begin{figure}[t]
\centering
\includegraphics[width=0.45\textwidth]{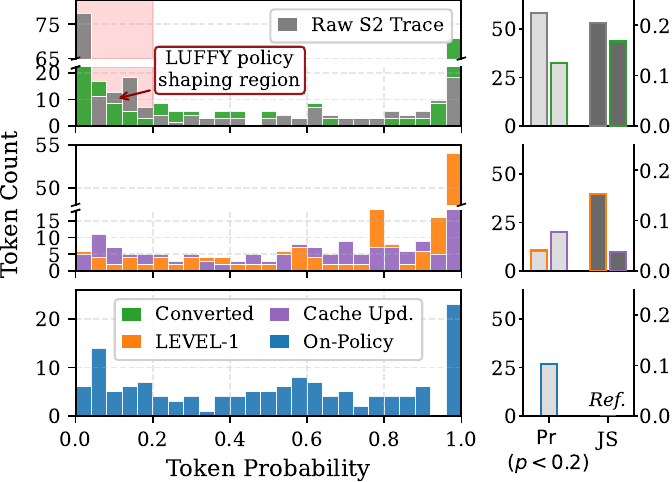}
% \caption{\textbf{Token-probability distributions and alignment statistics (computed on $\mathcal{D}_{\mathrm{upd}}$).}
% \textbf{Top}: After conversion, only a small fraction of tokens fall into LUFFY’s shaping band (0--0.2, shaded); most mass shifts to higher probabilities.
% \textbf{Middle}: LEVEL-1 self-rolled traces concentrate near prob.\ 1 (plan-conditioned); after cache updates, the distribution shifts toward the on-policy reference.
% \textbf{Bottom}: On-policy.
% \textbf{Right}: Tail mass $\Pr(p<0.2)$ and Jensen--Shannon divergence to the on-policy reference $P_{\mathrm{ref}}$ (lower is closer), quantifying reduced distribution gap after cache updates.}

\caption{\textbf{Token-probability distributions and simple diagnostics on $\mathcal{D}_{\mathrm{upd}}$.}
\textbf{Top}: Raw Agent S2 traces (gray) versus converted traces (green). Conversion substantially reduces the extreme low-probability tail ($p<0.2$) but still leaves a large mass in regions atypical for the base policy.
\textbf{Middle}: LEVEL-1 self-rolled guidance (orange) and LEVEL-2 cache-updated traces (purple). LEVEL-1 makes guidance highly reachable (peaked near $p\approx 1$), while LEVEL-2 reshapes the cache toward the on-policy reference, yielding a much smaller JS divergence.
\textbf{Bottom}: On-policy baseline.
\textbf{Right}: Tail mass $\Pr(p<0.2)$ and JS w.r.t.\ the on-policy reference.}

\label{fig:histogram}
% \vspace{-2mm}
\end{figure}

\paragraph{Distribution Analysis Protocol.}
To measure internalization beyond the base policy's existing competence, we construct an \emph{expert-only} task set $\mathcal{D}_{\text{expert\_only}}$ by removing all tasks that the untrained base policy can solve with Pass@5 from the self-rolled success pool $\mathcal{D}_{\text{self}}$, i.e., $\mathcal{D}_{\text{expert\_only}} = \{(x,\tau)\in\mathcal{D}_{\text{self}} \mid \text{Pass@5}_{\text{base}}(x)=0\}$, yielding $|\mathcal{D}_{\text{expert\_only}}|=54$.
We then form an update-triggered subset $\mathcal{D}_{\text{upd}} \subset \mathcal{D}_{\text{expert\_only}}$ of 19 tasks on which BEPA triggers cache updates during training.
On $\mathcal{D}_{\text{upd}}$, we score token probabilities under the base policy, aggregate histograms over $[0,1]$, and in \cref{fig:histogram} report two descriptive statistics: the tail mass $\Pr(p{<}0.2)$ (showing how much probability remains in LUFFY's shaping band after conversion) and the Jensen--Shannon divergence to the on-policy reference~\cite{Menéndez1997THEJD} (a coarse measure of shape similarity to on-policy rollouts rather than a formal on-/off-policy test).

\paragraph{Implementation Details.} 
We use UITARS1.5-7B as the base end-to-end agent for all experiments. The plan extractor $\phi$ we use is GPT-4o. As the expert source, we collect 115 successful OSWorld trajectories generated by Agent S2 and use them for conversion and self-rolling (success rate is 76\%, resulting in $|\mathcal{D}_{\text{self}}|=88$). For reinforcement learning, we adopt GRPO with a rollout group size of $N=8$ and a maximum episode length of $15$ steps, following the OSWorld's 15-step evaluation setting. See more details in the \cref{app:implementation_details}.

\subsection{Main Results}\label{subsec:main_results}

\paragraph{Overall Performance.}
As shown in \cref{tab:main_table}, BEPA achieves $32.13\%$ success on OSWorld-Verified, improving over UITARS1.5-7B ($22.87\%$) by $+9.26$ points ($+40.5\%$ relative) and over GRPO ($23.60\%$) by $+8.53$ points ($+36.1\%$ relative).
BEPA also improves all three splits: on $\mathcal{D}_{\text{train}}$ it reaches $73.23\%$ (vs.\ $55.12\%$ for UITARS1.5-7B and $58.02\%$ for GRPO), and on the strictly held-out set $\mathcal{D}_{\text{held\_out}}$ it improves to $10.30\%$ (vs.\ $5.74\%$ for UITARS1.5-7B and $5.32\%$ for GRPO), indicating better generalization beyond solvable training tasks.
Detailed results on OSWorld-Verified, MMBench-GUI and Online-Mind2Web are provided in \cref{app:full_results}.

\paragraph{Baseline Analysis.}
Na\"ive expert integration strategies consistently underperform.
SFT+RL drops to $14.74\%$ overall and $39.37\%$ on $\mathcal{D}_{\text{train}}$, indicating severe forgetting and weakened exploration.
RL+SFT reaches $20.88\%$ overall and does not improve held-out generalization ($3.53\%$ on $\mathcal{D}_{\text{held\_out}}$).
Trace Replacement ($23.91\%$) and LUFFY ($24.11\%$) remain close to GRPO overall, while showing a large generalization gap (Replacement: $66.50\%\rightarrow 1.29\%$; LUFFY: $65.44\%\rightarrow 2.16\%$ from $\mathcal{D}_{\text{train}}$ to $\mathcal{D}_{\text{held\_out}}$), suggesting that \emph{static} off-policy injection does not robustly transfer to unseen tasks.
Moreover, LUFFY performs similarly to Replacement in our setting because its policy shaping mainly reweights gradients toward low-probability tokens in \emph{raw} off-policy traces; on $\mathcal{D}_{\text{upd}}$, raw Agent S2 traces place $58.33\%$ of tokens in the $p<0.2$ band with JS$\approx0.2040$ to the on-policy reference, whereas after conversion the tail mass and JS drop to $32.48\%$ and $0.1676$ (\cref{fig:histogram}), leaving a much smaller shaping band for LUFFY to act on.

\begin{table}[t]
    \centering
    \small
    \resizebox{0.49\textwidth}{!}{%
    \begin{tabular}{l cccc}
        \toprule
        \textbf{Method} & \textbf{$\mathcal{D}_{\text{expert\_only}}$} & \textbf{$\mathcal{D}_{\text{train}}$} & \textbf{$\mathcal{D}_{\text{held\_out}}$} & \textbf{Overall (\%)} \\
        \midrule
        Agent S2 & - & - & - & 33.00 \\
        \textbf{Jedi-7B w/ o3} & - & - & - & \textbf{42.40} \\
        OpenAI o3 & - & - & - & 9.10 \\
        Doubao-1.5-Think & - & - & - & 31.90 \\
        Claude-4-Sonnet & - & - & - & 31.20 \\
        \midrule
        UITARS-72B-DPO & - & - & - & 24.00 \\
        \textbf{GUI-Owl-7B} & - & - & - & \textbf{32.10} \\
        OpenCUA-7B & - & - & - & $24.30_{\pm 1.40}$ \\
        ARPO & - & - & - & $23.86_{\pm 0.72}$ \\
        UITARS1.5-7B & 18.52 & 55.12 & 5.74 & $22.87_{\pm 0.97}$ \\    
        \midrule
        SFT & 5.56 & 47.77 & 1.65 & $17.65_{\pm 0.55}$ \\
        GRPO & 11.11 & 58.02 & 5.32 & $23.60_{\pm 1.15}$ \\
        RL+SFT & 14.81 & 53.55 & 3.53 & $20.88_{\pm 1.55}$ \\
        SFT+RL & 9.26 & 39.37 & 1.66 & $14.74_{\pm 0.60}$ \\
        Trace Replacement & 18.52 & 66.50 & 1.29 & $23.91_{\pm 2.35}$ \\
        LUFFY & 19.01 & 65.44 & 2.16 & $24.11_{\pm 2.10}$ \\
        \midrule
        LEVEL-1 & 25.93 & 69.20 & 5.05 & $27.30_{\pm 1.45}$ \\
        LEVEL-2 & 29.18 & 71.65 & 7.48 & $29.74_{\pm 0.90}$ \\
        \textbf{BEPA (ours)} & 35.19 & 73.23 & 10.30 & $\mathbf{32.13}_{\pm \mathbf{0.25}}$ \\
        % \textbf{BEPA$^\dagger$ (ours)} & 40.74 & 75.17 & 12.91 & $\mathbf{34.51}_{\pm \mathbf{1.22}}$ \\
        \bottomrule
    \end{tabular}
    }
    \caption{\textbf{Performance on OSWorld-Verified.} We reported the average success rate (\%) on $\mathcal{D}_{\text{expert\_only}}$, $\mathcal{D}_{\text{train}}$ and $\mathcal{D}_{\text{held\_out}}$. For the performance on the whole OSWorld-Verified dataset (Overall), the average success rate (\%) with standard deviation across 3 runs is reported.}
    \label{tab:main_table}
\end{table}

\subsection{Why Static Expert Integration Fails}\label{subsec:static_fail}

\begin{figure}[t]
\centering
\includegraphics[width=0.49\textwidth]{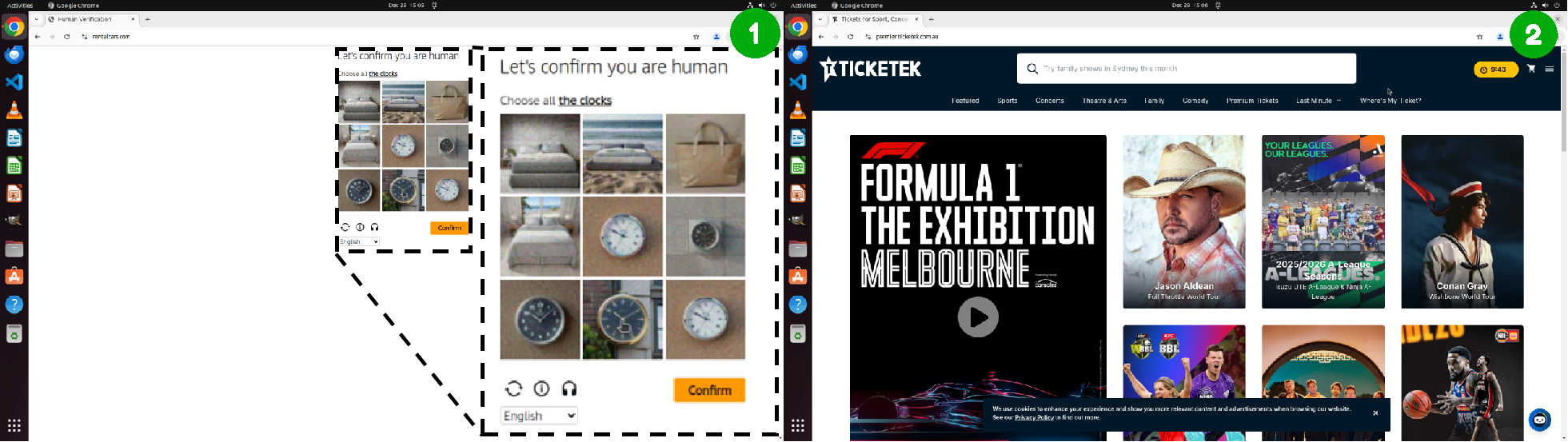}
\caption{\textbf{Dynamic environments in OSWorld.}
(1) \textbf{Pop-ups}: human-verification challenges and pop-up dialogs appear stochastically with the random content.
(2) \textbf{Web Content}: Real time content and random ads.}
\label{fig:dynamic_env}
% \vspace{-2mm}
\end{figure}

\begin{figure*}[ht]
\centering
\begin{subfigure}[b]{0.32\linewidth}
\includegraphics[width=\linewidth]{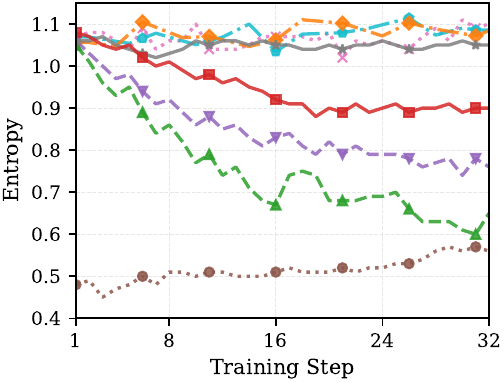}
\caption{Entropy}\label{fig:entropy}
\end{subfigure}
\hfill
\begin{subfigure}[b]{0.32\linewidth}
\includegraphics[width=\linewidth]{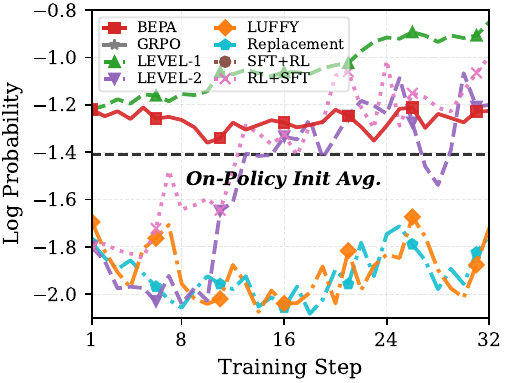}
\caption{Off-policy Avg. Log-prob.}\label{fig:offpolicy_logprob}
\end{subfigure}
\hfill
\begin{subfigure}[b]{0.32\linewidth}
\includegraphics[width=\linewidth]{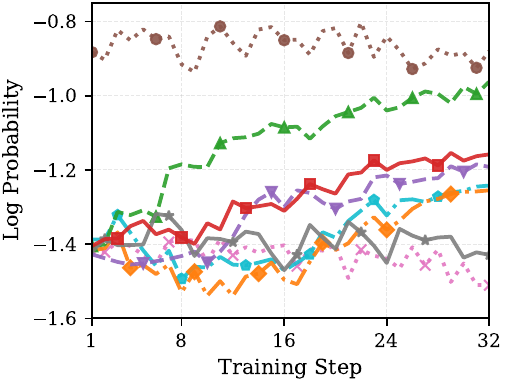}
\caption{On-policy Avg. Log-prob.}\label{fig:onpolicy_logprob}
\end{subfigure}
\caption{\textbf{Training dynamics.}
(a) \textbf{Policy entropy}: SFT+RL collapses during the SFT stage, and LEVEL-1 shows a pronounced entropy drop due to high-confidence plan-conditioned self-rolled guidance; in contrast, LUFFY/Replacement maintain entropy close to the on-policy baseline, while BEPA sustains a moderate entropy profile alongside steadily improving expert-only success.
(b--c) \textbf{Avg.\ log-probability on successful trajectories}: BEPA improves off-policy and on-policy successes in tandem, indicating gradual assimilation without degrading on-policy learning; LUFFY/Replacement exhibit an early dip and only partial recovery on on-policy log-probability, converging to a low plateau, while SFT+RL shows overfitting that harms on-policy improvement.}

\label{fig:training_dynamics}
\vspace{-4mm}
\end{figure*}

\paragraph{SFT vs.\ GRPO in Dynamic GUIs.}
OSWorld-style tasks are highly dynamic: the initial state and valid actions can change across episodes due to human-verification popups, time-sensitive banners, and other transient UI elements (\cref{fig:dynamic_env}).
SFT minimizes token-level cross-entropy on a small converted expert dataset ($125$ successful tasks, $1070$ step-level pairs), encouraging the policy to \emph{replay} expert action sequences under an idealized history rather than re-reading the current screen and adapting to these variations.
By contrast, GRPO-style mixed updates use group-normalized advantages (\cref{eq:mixed-adv,eq:bepa-obj}) and PPO clipping to simultaneously push up successful trajectories, push down failures, and keep updates within a trust region of $\pi_{\theta_{\text{old}}}$, which stabilizes learning on mixed on-/off-policy batches.
Empirically, SFT does increase the log-probability of converted traces (\cref{fig:offpolicy_logprob}), but higher off-policy likelihood alone does not translate into robust on-policy success under distribution shift: SFT+RL collapses entropy during the SFT stage and struggles to recover exploration (\cref{fig:entropy}), and RL+SFT yields only modest overall gains and weak held-out performance (\cref{tab:main_table}).

\paragraph{Static off-policy mixing.}
For LUFFY and naive Replacement, the main issue is not exploration collapse—both maintain entropy close to the on-policy baseline (\cref{fig:entropy})—but \emph{non-internalization} under distribution shift.
They improve off-policy successes only marginally (\cref{fig:offpolicy_logprob}) and can initially hinder on-policy learning, where success log-probability dips before partially recovering to a low plateau (\cref{fig:onpolicy_logprob}), and held-out performance remains low (\cref{tab:main_table}).
This indicates that static guidance is not kept aligned with the policy’s evolving manifold.

\subsection{Mechanism Analysis}\label{subsec:mechanism_analysis}

We analyze why BEPA works beyond final success rates through two complementary
mechanisms: (i) \textbf{native, learnable guidance} from LEVEL-1; and
(ii) \textbf{dynamic alignment} from LEVEL-2.

\begin{figure}[t]
\centering
\includegraphics[width=0.4\textwidth]{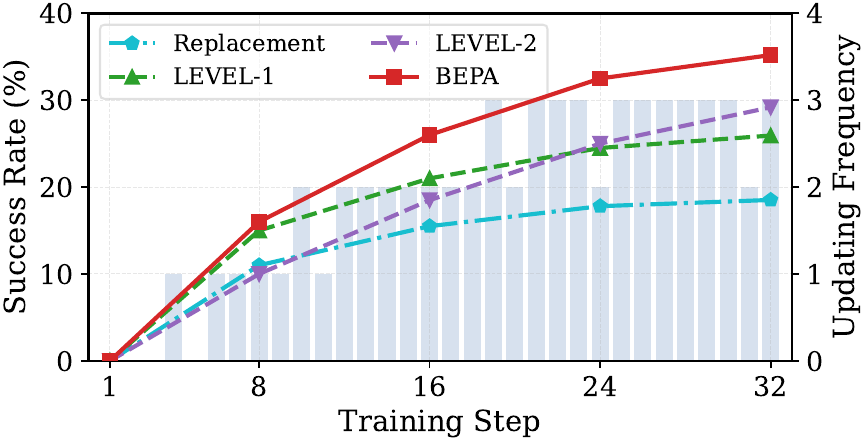}
\caption{\textbf{Expert-to-Policy Assimilation.} The evolution of expert-only success ($\mathcal{D}_{\text{expert\_only}}$, left axis) versus cache updating frequency (right axis). The updating frequency of BEPA rises in tandem with performance, demonstrating that rising competence drives active cache updates, thereby sustaining the effective \textbf{assimilation} of expert guidance into the policy.}
\label{fig:alignment_dynamics}
% \vspace{-2mm}
\end{figure}

\subsubsection{LEVEL-1: Native Guidance}
LEVEL-1 makes expert solutions \emph{reachable} by re-generating trajectories under
the base policy with plan conditioning, avoiding the direct
mismatch of compositional expert traces (\cref{fig:distribution_bias}).
On $\mathcal{D}_{\text{upd}}$, raw Agent S2 traces exhibit a very heavy low-probability
tail and noticeable divergence to the on-policy reference (tail mass $58.33\%$,
JS $0.2040$), and conversion alone only partially mitigates this (converted:
$32.48\%$, JS $0.1676$; \cref{fig:histogram}).
LEVEL-1 self-rolled guidance further increases the base policy’s confidence on
guided successes (tail mass $10.53\%$, JS $0.1525$) and achieves higher average
log-probability than both converted traces and initial on-policy rollouts
(\cref{fig:offpolicy_logprob}), making guidance much more learnable.
Empirically, this yields a strong boost on expert-covered tasks
($25.93\%$ on $\mathcal{D}_{\text{expert}}$), but limited transfer to held-out
tasks ($5.05\%$ on $\mathcal{D}_{\text{held\_out}}$) and a pronounced entropy
drop (\cref{fig:entropy,tab:main_table}), indicating that overly confident
plan-conditioned guidance can narrow exploration if left static.

\subsubsection{LEVEL-2: Dynamic Alignment}
LEVEL-2 keeps guidance synchronized with the evolving policy by refreshing the
cache with the agent's own successful executions (Sec.~\ref{sec:assimilation}),
instead of repeatedly injecting a fixed converted distribution as in static
Replacement/LUFFY.
On $\mathcal{D}_{\text{upd}}$, this yields guidance whose token-probability
histogram is much closer to the on-policy reference: the JS divergence is reduced
to $0.0366$, and the tail mass $\Pr(p<0.2)$ moves from $10.53\%$ (LEVEL-1) toward
the on-policy profile ($19.88\%$ vs.\ $26.95\%$; \cref{fig:histogram}).
LEVEL-2 also improves held-out generalization to $7.48\%$, and BEPA further raises
it to $10.30\%$ on $\mathcal{D}_{\text{held\_out}}$ (\cref{tab:main_table}), while
guided and on-policy success likelihoods improve in tandem
(\cref{fig:offpolicy_logprob,fig:onpolicy_logprob}), indicating that guidance is
being gradually internalized rather than remaining a static off-policy signal.

\subsection{Ablation}\label{subsec:ablation}
The ablations in \cref{tab:main_table} and the learning curves in \cref{fig:alignment_dynamics} match this division of labor.
Replacement saturates early on expert-covered tasks ($18.52\%$ on $\mathcal{D}_{\text{expert}}$) and generalizes poorly ($1.29\%$ on $\mathcal{D}_{\text{held\_out}}$).
LEVEL-1 improves reachability and rises faster on $\mathcal{D}_{\text{expert}}$ ($25.93\%$) but plateaus with limited transfer.
LEVEL-2 delivers sustained improvements ($29.18\%$ on $\mathcal{D}_{\text{expert}}$; $7.48\%$ held-out).
BEPA combines both, achieving the best expert-covered success ($35.19\%$) and held-out generalization ($10.30\%$); moreover, \cref{fig:alignment_dynamics} shows its cache update frequency increases in tandem with expert-only success, directly evidencing self-aligned assimilation.

% \section{Conclusion}
% We identify expert--policy mismatch as a key obstacle to leveraging high-quality framework trajectories for end-to-end GUI policies under RLVR. BEPA addresses this with a bi-level assimilation scheme: LEVEL-1 re-rolls expert plans into policy-reachable trajectories, and LEVEL-2 keeps guidance aligned via a self-updating cache that is injected only upon exploration failure. This design consistently improves screenshot-to-action agents on challenging computer-use benchmarks. More broadly, our results suggest that heterogeneous expert data is most effective when it is progressively \emph{assimilated} into the learner’s evolving distribution rather than statically mixed in.

% \section{Conclusion}
% We identify expert--policy mismatch as a key obstacle to leveraging high-quality framework trajectories for end-to-end GUI policies under RLVR, and show that naive SFT or static off-policy mixing either collapses exploration or fails to internalize guidance under distribution shift.
% BEPA addresses this with a bi-level assimilation scheme: LEVEL-1 quickly turns alien expert traces into high-confidence, policy-compatible self-rolled trajectories, while LEVEL-2 continually re-aligns guidance with the evolving policy by maintaining a self-updating cache that is injected only upon total on-policy failure. More broadly, our results suggest that heterogeneous expert data is most effective when it is progressively \emph{assimilated} into the learner's evolving distribution, rather than statically mixed in at the loss level.

\section{Conclusion}
We identify expert--policy mismatch as a central obstacle to turning high-quality framework trajectories into reliable gains for end-to-end GUI policies under RLVR. BEPA addresses this with a bi-level assimilation scheme: LEVEL-1 converts alien expert traces into high-confidence, policy-compatible self-rolled trajectories, while LEVEL-2 keeps guidance aligned with the evolving policy via a self-updating cache that is injected only when on-policy exploration fails. Overall, our results suggest that heterogeneous expert data is most effective when progressively assimilated into the learner's own distribution, rather than statically mixed at the loss level.

\section{Limitations}
While BEPA delivers consistent gains under the evaluated setups, our study instantiates BEPA only on GUI-based computer-use benchmarks, using a specific backbone together with Agent S2 and GUI-Owl as expert sources and a particular plan-extraction pipeline. Extending BEPA to other domains and interaction modalities (e.g., mobile platforms, productivity suites, or non-GUI environments), as well as to more diverse and automatically mined expert pools, is a natural direction for future work. Our experiments further adopt an RLVR regime with sparse, verifiable rewards; adapting cache refresh and conditional guidance injection to settings with noisier or preference-based feedback, and combining BEPA with alternative reward modeling and credit-assignment schemes, are promising next steps. Finally, BEPA introduces additional machinery in the form of self-rolled trajectories and per-task caches; exploring lighter-weight variants, tighter integration with existing agent frameworks, and broader evaluations across backbones and longer-horizon tasks will help further characterize the scalability and generality of bi-level expert-to-policy assimilation.

\bibliography{custom}

%%%%%%%%%%%%%%%%%%%%%%%%%%%%%%%%%%%%%%%%%%%%%%%%%%%%%%%%%%%%
\newpage
\appendix

% ============================
% Appendix Table of Contents
% ============================
\phantomsection
\section*{Appendix Contents}
\label{sec:appendix_contents}
\addcontentsline{toc}{section}{Appendix Contents}

\begingroup
\small
\setlength{\parindent}{0pt}
\renewcommand{\arraystretch}{1.1}

\begin{tabular}{@{}p{0.80\linewidth}r@{}}
\hyperref[app:bepa_algo]{A.\quad BEPA Algorithm Details} &
  \hyperref[app:bepa_algo]{\pageref{app:bepa_algo}} \\

\quad\hyperref[app:bepa_pseudocode]{A.1\quad Pseudo-code} &
  \hyperref[app:bepa_pseudocode]{\pageref{app:bepa_pseudocode}} \\

\quad\hyperref[app:plan_concat]{A.2\quad Plan Concatenation Example} &
  \hyperref[app:plan_concat]{\pageref{app:plan_concat}} \\[0.4em]

\hyperref[app:convergence_analysis]{B.\quad Convergence Analysis} &
  \hyperref[app:convergence_analysis]{\pageref{app:convergence_analysis}} \\

\hyperref[app:benchmarks]{C.\quad Benchmarks and Evaluation Protocols} &
  \hyperref[app:benchmarks]{\pageref{app:benchmarks}} \\

\quad\hyperref[app:osworld]{C.1\quad OSWorld-Verified} &
  \hyperref[app:osworld]{\pageref{app:osworld}} \\

\quad\hyperref[app:mmbench]{C.2\quad MMBench-GUI} &
  \hyperref[app:mmbench]{\pageref{app:mmbench}} \\

\quad\hyperref[app:mind2web]{C.3\quad Online-Mind2Web} &
  \hyperref[app:mind2web]{\pageref{app:mind2web}} \\[0.4em]

\hyperref[app:baselines]{D.\quad Baselines} &
  \hyperref[app:baselines]{\pageref{app:baselines}} \\[0.4em]

\hyperref[app:implementation_details]{E.\quad Implementation Details} &
  \hyperref[app:implementation_details]{\pageref{app:implementation_details}} \\

\quad\hyperref[app:sft_details]{E.1\quad SFT on Converted Expert Traces} &
  \hyperref[app:sft_details]{\pageref{app:sft_details}} \\

\quad\hyperref[app:rl_sft_details]{E.2\quad RL with SFT Loss (RL+SFT)} &
  \hyperref[app:rl_sft_details]{\pageref{app:rl_sft_details}} \\

\quad\hyperref[app:luffy_details]{E.3\quad LUFFY} &
  \hyperref[app:luffy_details]{\pageref{app:luffy_details}} \\

\quad\hyperref[sec:s2_logprob]{E.4\quad Log-Probability on S2 Raw Traces} &
  \hyperref[sec:s2_logprob]{\pageref{sec:s2_logprob}} \\

\quad\hyperref[subsec:arch]{E.5\quad Codebase: Verl-GUI} &
  \hyperref[subsec:arch]{\pageref{subsec:arch}} \\[0.4em]

\hyperref[app:interface_prompts]{F.\quad Interface and Prompt Templates} &
  \hyperref[app:interface_prompts]{\pageref{app:interface_prompts}} \\

\quad\hyperref[app:action_space]{F.1\quad UI-TARS Action Space} &
  \hyperref[app:action_space]{\pageref{app:action_space}} \\

\quad\hyperref[app:s2_action_space]{F.2\quad Agent S2 Action Space} &
  \hyperref[app:s2_action_space]{\pageref{app:s2_action_space}} \\

\quad\hyperref[app:prompt_templates]{F.3\quad Prompt Templates} &
  \hyperref[app:prompt_templates]{\pageref{app:prompt_templates}} \\[0.4em]

\hyperref[app:full_results]{G.\quad Full Benchmark Results} &
  \hyperref[app:full_results]{\pageref{app:full_results}} \\[0.4em]

\hyperref[app:expert_generalization]{H.\quad Extensibility} &
  \hyperref[app:expert_generalization]{\pageref{app:expert_generalization}} \\[0.4em]

\hyperref[sec:sensitivity]{I.\quad Sensitivity Analysis} &
  \hyperref[sec:sensitivity]{\pageref{sec:sensitivity}} \\[0.4em]

\hyperref[sec:domain_breakdown]{J.\quad Domain Breakdown of BEPA's Gains} &
  \hyperref[sec:domain_breakdown]{\pageref{sec:domain_breakdown}} \\[0.4em]

\hyperref[sec:case_studies]{K.\quad Case Studies} &
  \hyperref[sec:case_studies]{\pageref{sec:case_studies}} \\
\end{tabular}

\endgroup

% \clearpage

% =========================================================
% Appendix A. BEPA Algorithm Details (Aligned Notation)
% =========================================================
% \clearpage
\section{BEPA Algorithm Details}\label{app:bepa_algo}

\subsection{Pseudo-code}\label{app:bepa_pseudocode}
\begin{algorithm}[H]
\caption{LEVEL-1: Self-Rolled Execution (Seed Cache Initialization)}\label{alg:l1_selfroll_aligned}
\small
\begin{algorithmic}[1]
\Require Expert successful traces $\mathcal{D}_E=\{(x,\tau_x^E)\}$; plan extractor $\phi(\cdot)$; base policy $\pi_{\theta}$; verifier $R(\cdot)$
\Ensure Self-rolled seed set $\mathcal{D}_{\text{self}}$; initialized cache $\mathcal{E}$

\State $\mathcal{D}_{\text{self}} \leftarrow \emptyset$; \ \ $\mathcal{E} \leftarrow \emptyset$
\ForAll{$(x,\tau_x^E)\in \mathcal{D}_E$}
    \State $p_x \leftarrow \phi(\tau_x^E)$
    \State Roll out $\pi_{\theta}(\cdot \mid s_t, x, p_x)$ in the environment to obtain $\tau_x^{\text{self}}$
    \If{$R(\tau_x^{\text{self}})=1$}
        \State $\mathcal{D}_{\text{self}} \leftarrow \mathcal{D}_{\text{self}} \cup \{(x,\tau_x^{\text{self}})\}$
        \State $\mathcal{E}_x \leftarrow \tau_x^{\text{self}}$ \Comment{seed cache with policy-compatible successes}
    \EndIf
\EndFor
\State \Return $\mathcal{D}_{\text{self}}, \mathcal{E}$
\end{algorithmic}
\end{algorithm}

\begin{algorithm}[H]
\caption{LEVEL-2: Self-Aligned Off-Policy Assimilation with Conditional Trace Replacement}\label{alg:l2_assim_aligned}
\small
\begin{algorithmic}[1]
\Require Task set $\mathcal{X}$; policy parameters $\theta$; cache $\mathcal{E}$ (seeded by $\mathcal{D}_{\text{self}}$); verifier $R(\cdot)$
\Require Group size $N$; cache update rule $\mathcal{U}(\cdot)$
\Ensure Updated policy parameters $\theta$
\For{iteration $k = 1,2,\ldots$}
    \State Set behavior policy $\pi_{\theta_{\text{old}}} \leftarrow \pi_{\theta}$ \Comment{standard on-policy collection in GRPO}
    \ForAll{sampled tasks $x \sim \mathcal{X}$}
        \State Collect a rollout group $\mathcal{T}_x = \{\tau_{x,i}\}_{i=1}^N$ by sampling $\tau_{x,i} \sim \pi_{\theta_{\text{old}}}(\cdot \mid x)$
        \State $\mathcal{T}_x^{\mathrm{succ}} \leftarrow \{\tau \in \mathcal{T}_x \mid R(\tau) = 1\}$

        \If{$\mathcal{T}_x^{\mathrm{succ}} \neq \varnothing$}
            \State $\tau_x^{\mathrm{new}} \leftarrow \mathcal{U}(\mathcal{T}_x^{\mathrm{succ}})$ \Comment{e.g., random sampling}
            \State $\mathcal{E}_x \leftarrow \tau_x^{\mathrm{new}}$ \Comment{dynamic cache refresh (Sec.~\ref{sec:assimilation})}
            \State $\hat{\mathcal{T}}_x \leftarrow \mathcal{T}_x$
        \Else
            \If{$\mathcal{E}_x \neq \varnothing$}
                \State $\tau_x^{\text{off}} \leftarrow \mathcal{E}_x$
                \State $\hat{\mathcal{T}}_x \leftarrow \{\tau_x^{\text{off}}\} \cup \{\tau_{x,2}, \dots, \tau_{x,N}\}$ \Comment{\cref{eq:Tx}}
            \Else
                \State $\hat{\mathcal{T}}_x \leftarrow \mathcal{T}_x$
            \EndIf
        \EndIf

        \State Compute $\hat{A}$ by \cref{eq:mixed-adv}
    \EndFor
    \State Update $\theta$ by maximizing $\mathcal{J}_{\mathrm{BEPA}}(\theta)$ in \cref{eq:bepa-obj}
\EndFor
\end{algorithmic}
\end{algorithm}

\subsection{Plan Concatenation Example}\label{app:plan_concat}

\begin{tcolorbox}[
    enhanced,
    width=\columnwidth,
    colback=white,
    colframe=black,
    coltitle=white,
    colbacktitle=black,
    fonttitle=\bfseries,
    title=Plan-Conditioned Prompt Example,
    boxrule=0.8pt,
    arc=6pt,
    left=8pt, right=8pt,
    top=6pt, bottom=6pt,
    before skip=4pt,
    after skip=2pt
]
{\small
\setstretch{1.10}
\begin{lstlisting}[style=promptbox]
Can you enable the 'Do Not Track' feature in Chrome to enhance my online privacy?

1. Click on the three-dot menu icon in the upper-right corner of the Chrome window.
2. Select "Settings" from the dropdown menu.
3. Click on "Privacy and security".
4. Scroll down to find the "Send a 'Do Not Track' request..." option.
5. Enable the toggle to send a 'Do Not Track' request.
6. Click "Confirm" to apply the changes.
\end{lstlisting}
}
\end{tcolorbox}

\captionof{figure}{An example of plan conditioning for LEVEL-1: we append an extracted plan to the original instruction.}
\label{fig:plan_injection_example}

\clearpage
\onecolumn

\section{Convergence Analysis}
\label{app:convergence_analysis}

We analyze BEPA as a clipped mixed-policy policy-gradient method based on
\cref{eq:mixed-adv,eq:bepa-obj}, and show that it enjoys the standard
$O(1/\sqrt{K})$ non-convex convergence rate for the GRPO/PPO surrogate.

\subsection{Setup and Assumptions}

Let $r_t(\theta)$ and $w_t(\theta)$ be the usual ratio and clipped ratio:
\begin{equation}
r_t(\theta) =
\frac{\pi_\theta(a_t\mid s_t,x)}{\pi_{\theta_{\text{old}}}(a_t\mid s_t,x)},
\qquad
w_t(\theta) = \mathrm{clip}\big(r_t(\theta);\,1-\epsilon,\,1+\epsilon\big).
\end{equation}
At iteration $k$, BEPA optimizes the clipped GRPO surrogate
\begin{equation}
\mathcal{J}_{\mathrm{BEPA}}(\theta)
= \mathbb{E}_{\tau\sim\mu_k}\!\Bigg[
\frac{1}{Z} \sum_{t=1}^{|\tau|}
L^{\mathrm{CLIP}}\big(r_t(\theta), \hat{A}_\tau\big)
\Bigg],
\label{eq:bepa_surrogate_appendix}
\end{equation}
where $Z = \sum_{\tau}|\tau|$ and
$L^{\mathrm{CLIP}}(r,A)=\min[rA,\mathrm{clip}(r;1\pm\epsilon)A]$.
The expectation is taken with respect to the behavior distribution
$\mu_k$ induced by BEPA at iteration $k$ (a mixture of on-policy rollouts and
cached trajectories; cf.\ Sec.~4.3).
For a trajectory $\tau$ and parameters $\theta$, define the per-trajectory
gradient contribution
\begin{equation}
H_k(\tau;\theta)
:= \frac{1}{Z}\sum_{t}
w_t(\theta)\,\hat A_\tau\,
\nabla_{\theta}\log\pi_{\theta}(a_t\mid s_t,x),
\label{eq:grad_estimator_appendix}
\end{equation}
so the stochastic gradient used by BEPA is
$\hat g(\theta_k)=H_k(\tau;\theta_k)$ with $\tau\sim\mu_k$.

We adopt standard assumptions used in non-convex policy-gradient and
PPO analyses~\cite{reddi2016stochastic,mei2020global,jin2024ppo}:

\begin{itemize}[leftmargin=1.4em, itemsep=0.25em, topsep=0.2em]
\item[(i)] \emph{$L$-smoothness.}
$\mathcal{J}_{\mathrm{BEPA}}(\theta)$ has $L$-Lipschitz gradients:
$\|\nabla \mathcal{J}_{\mathrm{BEPA}}(\theta) - \nabla \mathcal{J}_{\mathrm{BEPA}}(\theta')\|_2
\le L\|\theta-\theta'\|_2$.

\item[(ii)] \emph{Bounded score and advantage.}
There exist constants $G,A_{\max}>0$ such that for all $(s_t,x,a_t)$
and all $\tau$,
\(
\|\nabla_\theta\log\pi_\theta(a_t\mid s_t,x)\|_2 \le G
\)
and
\(
|\hat A_\tau| \le A_{\max}.
\)
In our setting, rewards are $0/1$ and group-normalized advantages are
computed by \cref{eq:mixed-adv} over a rollout group of size $N$.
When trace replacement triggers (all on-policy rollouts fail and the
task has a cached success), the reward multiset is
$\hat G = \{1,0,\dots,0\}$, which yields
$\hat A_{\text{succ}} = \sqrt{N-1}$ and
$\hat A_{\text{fail}} = -1/\sqrt{N-1}$,
so we may take $A_{\max} = \sqrt{N-1}$ (e.g., $\sqrt{7}$ for $N{=}8$).
In all other cases, rewards lie in $\{0,1\}$ and \cref{eq:mixed-adv}
produces advantages with smaller magnitude.

\item[(iii)] \emph{Clipped importance weights.}
By construction,
\(
|w_t(\theta)| \le 1+\epsilon
\)
for all $t,\theta$.
\end{itemize}

\paragraph{Unbiasedness for the surrogate.}
We emphasize that we analyze the clipped surrogate
$\mathcal{J}_{\mathrm{BEPA}}$ in \cref{eq:bepa_surrogate_appendix},
not the unclipped environment return.
At iteration $k$, $\mu_k$ is fixed by the data-collection procedure
(using $\pi_{\theta_{\text{old}}}$ and the current cache), so
\begin{equation}
\mathcal{J}_{\mathrm{BEPA}}(\theta)
=
\mathbb{E}_{\tau\sim\mu_k}\big[F(\theta,\tau)\big],
\qquad
F(\theta,\tau)
:= \frac{1}{Z}\sum_t
L^{\mathrm{CLIP}}\big(r_t(\theta),\hat A_\tau\big).
\end{equation}
Differentiating under the expectation gives
\begin{align}
\nabla \mathcal{J}_{\mathrm{BEPA}}(\theta)
&=
\mathbb{E}_{\tau\sim\mu_k}\big[\nabla_\theta F(\theta,\tau)\big]
=
\mathbb{E}_{\tau\sim\mu_k}\big[H_k(\tau;\theta)\big],
\end{align}
so $\hat g(\theta_k)=H_k(\tau;\theta_k)$ is an unbiased estimator of
$\nabla \mathcal{J}_{\mathrm{BEPA}}(\theta_k)$ \emph{for any} behavior distribution
$\mu_k$, including mixtures with injected cached trajectories.
This is exactly the same notion of unbiasedness used in standard
policy-gradient theory, but applied to the clipped surrogate objective:
we do not claim monotonic improvement guarantees for the true
environment return.

\subsection{Variance Bound}

Using (ii)–(iii) and Cauchy–Schwarz, for any trajectory $\tau$ and
iteration $k$ we have
\begin{align}
\|H_k(\tau;\theta)\|_2^2
&=
\Bigg\|
\frac{1}{Z} \sum_{t}
w_t(\theta)\, \hat A_\tau\,
\nabla_\theta \log \pi_\theta(a_t\mid s_t,x)
\Bigg\|_2^2
\nonumber\\
&\le
\frac{1}{Z^2} \sum_{t}
|w_t(\theta)|^2\,|\hat A_\tau|^2\,
\big\|\nabla_\theta \log \pi_\theta(a_t\mid s_t,x)\big\|_2^2
\nonumber\\
&\le
\frac{1}{Z^2} \sum_{t}
(1+\epsilon)^2 A_{\max}^2 G^2
\;\le\; \sigma^2,
\label{eq:Hk_bound}
\end{align}
for some $\sigma^2 = O((1+\epsilon)^2 G^2 A_{\max}^2)$ that does not
depend on the behavior distribution $\mu_k$.
Taking expectation over $\tau \sim \mu_k$ gives a uniform second-moment
bound
\begin{equation}
\mathbb{E}_{\tau \sim \mu_k}\big[\|\hat g(\theta_k)\|_2^2\big]
=
\mathbb{E}_{\tau \sim \mu_k}\big[\|H_k(\tau;\theta_k)\|_2^2\big]
\le \sigma^2,
\label{eq:variance_bound}
\end{equation}
for all iterations $k$, regardless of how $\mu_k$ mixes on-policy and
cached trajectories.

\subsection{Convergence of BEPA}

\begin{theorem}[Convergence of BEPA]
\label{thm:bepa_convergence}
Let $\{\theta_k\}_{k=0}^{K-1}$ be the parameter sequence generated by
BEPA with learning rate $\alpha_k = c/\sqrt{K}$.
Under assumptions \textnormal{(i)}–\textnormal{(iii)}, there exist
constants $C_1,C_2>0$ such that
\begin{align}
\frac{1}{K}
\sum_{k=0}^{K-1}
\mathbb{E}\big[\|\nabla \mathcal{J}_{\mathrm{BEPA}}(\theta_k)\|_2^2\big]
\;\le\;&\;
\frac{C_1}{\sqrt{K}}\big(\mathcal{J}_{\mathrm{BEPA}}(\theta^*)
- \mathcal{J}_{\mathrm{BEPA}}(\theta_0)\big)
\nonumber\\
&\;+\; \frac{C_2}{\sqrt{K}}\,\sigma^2,
\label{eq:final_rate_appendix}
\end{align}
where $\theta^*$ is an optimal solution of the surrogate objective.
Thus BEPA converges to a first-order stationary point of
$\mathcal{J}_{\mathrm{BEPA}}(\theta)$ at rate $O(1/\sqrt{K})$.
\end{theorem}

\paragraph{Proof sketch.}
By $L$-smoothness of $\mathcal{J}_{\mathrm{BEPA}}(\theta)$ and the
update $\theta_{k+1} = \theta_k - \alpha_k \hat g(\theta_k)$, we have
\begin{align}
\mathcal{J}_{\mathrm{BEPA}}(\theta_{k+1})
&\le
\mathcal{J}_{\mathrm{BEPA}}(\theta_k)
- \alpha_k \big\langle \nabla \mathcal{J}_{\mathrm{BEPA}}(\theta_k),
\hat g(\theta_k) \big\rangle
+ \frac{L\alpha_k^2}{2}\,\|\hat g(\theta_k)\|_2^2.
\end{align}
Taking expectation and using
$\mathbb{E}[\hat g(\theta_k)] = \nabla
\mathcal{J}_{\mathrm{BEPA}}(\theta_k)$ gives
\begin{align}
\mathbb{E}[\mathcal{J}_{\mathrm{BEPA}}(\theta_{k+1})]
&\le
\mathbb{E}[\mathcal{J}_{\mathrm{BEPA}}(\theta_k)]
- \alpha_k\,\mathbb{E}\big[\|\nabla
\mathcal{J}_{\mathrm{BEPA}}(\theta_k)\|_2^2\big]
\nonumber\\
&\quad
+ \frac{L\alpha_k^2}{2}\,
\mathbb{E}\big[\|\hat g(\theta_k)\|_2^2\big].
\end{align}
Applying the variance bound \cref{eq:variance_bound} and rearranging
yields
\begin{equation}
\alpha_k\,\mathbb{E}\big[\|\nabla
\mathcal{J}_{\mathrm{BEPA}}(\theta_k)\|_2^2\big]
\;\le\;
\mathbb{E}[\mathcal{J}_{\mathrm{BEPA}}(\theta_k)]
- \mathbb{E}[\mathcal{J}_{\mathrm{BEPA}}(\theta_{k+1})]
+ C\,\alpha_k^2\sigma^2,
\label{eq:per_step_ineq_appendix}
\end{equation}
for some constant $C>0$ (e.g., $C=L/2$).
Summing \cref{eq:per_step_ineq_appendix} over $k=0,\dots,K-1$ gives
\begin{align}
\sum_{k=0}^{K-1}\alpha_k\,
\mathbb{E}\big[\|\nabla
\mathcal{J}_{\mathrm{BEPA}}(\theta_k)\|_2^2\big]
&\le
\mathbb{E}[\mathcal{J}_{\mathrm{BEPA}}(\theta_0)]
- \mathbb{E}[\mathcal{J}_{\mathrm{BEPA}}(\theta_K)]
+ C\,\sigma^2 \sum_{k=0}^{K-1}\alpha_k^2
\nonumber\\
&\le
\mathcal{J}_{\mathrm{BEPA}}(\theta_0)
- \mathcal{J}_{\mathrm{BEPA}}(\theta^*)
+ C\,\sigma^2 \sum_{k=0}^{K-1}\alpha_k^2,
\label{eq:sum_alpha_grad_appendix}
\end{align}
since $\mathcal{J}_{\mathrm{BEPA}}(\theta_K)
\ge \mathcal{J}_{\mathrm{BEPA}}(\theta^*)$.
With constant step size $\alpha_k = \alpha = c/\sqrt{K}$ we have
\begin{equation}
\sum_{k=0}^{K-1}\alpha_k
= K\alpha = c\sqrt{K},
\qquad
\sum_{k=0}^{K-1}\alpha_k^2
= K\alpha^2 = c^2.
\label{eq:alpha_sums_appendix}
\end{equation}
Dividing \cref{eq:sum_alpha_grad_appendix} by $\sum_{k=0}^{K-1}\alpha_k$
and using \cref{eq:alpha_sums_appendix} yields
\begin{align}
\frac{\sum_{k=0}^{K-1}\alpha_k
\mathbb{E}\big[\|\nabla
\mathcal{J}_{\mathrm{BEPA}}(\theta_k)\|_2^2\big]}
{\sum_{k=0}^{K-1}\alpha_k}
&\le
\frac{\mathcal{J}_{\mathrm{BEPA}}(\theta_0)
- \mathcal{J}_{\mathrm{BEPA}}(\theta^*)}{c\sqrt{K}}
+ \frac{C\,\sigma^2 c^2}{c\sqrt{K}}
\nonumber\\
&=
\frac{C_1}{\sqrt{K}}\big(\mathcal{J}_{\mathrm{BEPA}}(\theta_0)
- \mathcal{J}_{\mathrm{BEPA}}(\theta^*)\big)
+ \frac{C_2}{\sqrt{K}}\,\sigma^2,
\end{align}
for suitable constants $C_1,C_2>0$.
Since $\alpha_k$ is constant, the left-hand side is simply the average
gradient norm:
\begin{equation}
\frac{\sum_{k=0}^{K-1}\alpha_k
\mathbb{E}\big[\|\nabla
\mathcal{J}_{\mathrm{BEPA}}(\theta_k)\|_2^2\big]}
{\sum_{k=0}^{K-1}\alpha_k}
=
\frac{1}{K}\sum_{k=0}^{K-1}
\mathbb{E}\big[\|\nabla
\mathcal{J}_{\mathrm{BEPA}}(\theta_k)\|_2^2\big],
\end{equation}
which gives exactly \cref{eq:final_rate_appendix}.

\paragraph{Remark (Injected success, group-wise normalization, and experience replay).}
When a single cached success is injected into $N{-}1$ failures (\cref{eq:Tx}),
the reward multiset is $\{1,0,\dots,0\}$ and \cref{eq:mixed-adv} produces one
positive advantage and $N{-}1$ negative advantages with
$\hat A_{\text{succ}} = \sqrt{N-1}$,
$\hat A_{\text{fail}} = -1/\sqrt{N-1}$,
and $\sum_{\tau\in\hat{\mathcal{T}}_x}\hat A_\tau = 0$.
Thus the group-wise normalization acts as a data-dependent baseline:
the injected success and failed rollouts are contrasted within the same
group, but the overall scale is bounded by $A_{\max}$ and the
second-moment bound \cref{eq:variance_bound} still holds.
This design biases each failed group toward the successful trajectory
(as intended) while keeping the update within the PPO trust region via
the clipping in $w_t(\theta)$.

Moreover, LEVEL-2 refreshes the cache with \emph{the agent’s own successful
rollouts}, so cached trajectories increasingly coincide with recent
on-policy experiences.
In this sense, the role of the cache gradually shifts from assimilating
external expert traces to providing a bounded-staleness \emph{experience
replay} buffer over the agent’s own successes \cite{lin1992self,mnih2015human,schaul2016prioritized}.
Under the clipped mixed-policy GRPO update, such experience replay is
compatible with the variance and convergence guarantees above and
connects BEPA to a long line of empirically validated replay-based
methods in reinforcement learning.

\twocolumn

% =========================================================
% Appendix D. Benchmarks and Evaluation Protocols
% (Use \cref{} for all figure/table references)
% =========================================================

\section{Benchmarks and Evaluation Protocols}\label{app:benchmarks}

\subsection{OSWorld-Verified}\label{app:osworld}
\paragraph{Benchmark.}
We conduct RLVR training on OSWorld-Verified only, an execution-based real-computer benchmark built upon OSWorld \cite{Xie2024OSWorldBM}, consisting of 369 diverse computer-use tasks with per-task setup and evaluation scripts. 
We summarize dataset statistics in \cref{tab:osworld_stats} and report main results in \cref{tab:main_table} (with per-domain breakdown in \cref{tab:osworld}). MMBench-GUI and Online-Mind2web are used as held-out evaluation benchmarks.

\paragraph{Evaluation Protocol.}
Each OSWorld-Verified task is executed in a fresh virtual-machine snapshot with its official initial-state config; the agent interacts with the desktop by emitting textual UITARS actions, which are mapped to \texttt{pyautogui} mouse/keyboard operations in the VM (\emph{cf.} \cite{Xie2024OSWorldBM}).
An episode terminates when the agent outputs \texttt{finished} or reaches the 15-step limit.
After termination, OSWorld runs the example-specific execution-based evaluation script to retrieve relevant files/UI state and returns a binary success signal $R(\tau)\in\{0,1\}$ (we count a task as solved iff $R(\tau)=1$).
All success rates reported on OSWorld-Verified are the fraction of tasks with $R(\tau)=1$, averaged over three random seeds.

\subsection{MMBench-GUI}\label{app:mmbench}
\paragraph{Benchmark.}
MMBench-GUI is a hierarchical, multi-platform benchmark designed to systematically assess GUI agents across four progressive levels of difficulty: GUI Content Understanding (L1), Element Grounding (L2), Task Automation (L3), and Task Collaboration (L4). We evaluate generalization on MMBench-GUI \cite{Wang2025MMBenchGUIHM}, a hierarchical multi-platform GUI benchmark spanning four levels (L1--L4) over multiple platforms, and reporting both success rate (SR) and Efficiency--Quality Area (EQA).
We provide benchmark statistics in \cref{tab:mmbench_stats} and full results in \cref{tab:L1,tab:L2}. Only L1 and L2 results are reported as L3 and L4 task configurations are not publicly available at the time of writing.

\paragraph{Evaluation Protocol.}
For L1 (GUI Content Understanding), we follow the official protocol of MMBench-GUI~\cite{Wang2025MMBenchGUIHM}: the agent receives a single GUI screenshot and a multiple-choice question with options, and must select one option; we compute accuracy (SR) per platform and difficulty, and aggregate scores as a weighted average across platforms as in the original benchmark.
For L2 (GUI Element Grounding), the agent is given a screenshot and a textual instruction describing a target element, and must output a click position; a prediction is counted as correct if the point lies inside the annotated bounding box, and we report accuracy over all evaluated elements, averaged across platforms and instruction types (Basic / Advanced) following the MMBench-GUI evaluation metric.

\subsection{Online-Mind2Web}\label{app:mind2web}
\paragraph{Benchmark.}
Online-Mind2Web \cite{xue2025illusionprogressassessingcurrentOnline-Mind2web} is a benchmark designed to evaluate the real-world performance of web agents on live websites, featuring 300 tasks across 136 popular sites in diverse domains with reliable LLM-as-a-Judge (WebJudge) automactic evaluation. Based on the number of steps required by human annotators, tasks are divided into three difficulty levels: Easy (1-5 steps, 83 tasks), Medium (6-10 steps, 143 tasks), and Hard (11+ steps, 74 tasks). Dataset statistics are summarized in \cref{tab:mind2web_stats}, and full results are reported in \cref{tab:mind2web}. We follow the official Online-Mind2Web evaluation procedure: each task is executed in a real browser session with a fixed step budget and is scored by task success (SR) based on the benchmark's execution-based evaluator.

\paragraph{Evaluation Protocol.}
We follow the official Online-Mind2Web evaluation setup with WebJudge, an LLM-as-a-judge pipeline.
For each task, the agent interacts in a real browser session under a fixed step budget; the task description, full action sequence, and up to 50 screenshots are logged and fed to WebJudge.
WebJudge first extracts key requirements from the description (especially comparative terms such as ``cheapest'' or ``most recent''), then filters screenshots by scoring each frame and retaining only informative ones, and finally decides success or failure based on the filtered screenshots, action history, and key requirements.
We adopt the benchmark’s strict criteria on correct filter application, verifiability from screenshots/actions, and exact satisfaction of range and submission conditions.
Following \cite{xue2025illusionprogressassessingcurrentOnline-Mind2web}, we use \texttt{o4-mini} as the judge and run evaluation in parallel (up to 60 workers); success rate (SR) is defined as the proportion of tasks judged as \textsc{SUCCESS} by WebJudge.

\begin{table}[t!]
\centering
\resizebox{0.3\textwidth}{!}{%
\begin{tabular}{lr}
\toprule
\textbf{Domain} & \textbf{\# Examples} \\
\midrule
Chrome & 46 \\
GIMP & 26 \\
LibreOffice Calc & 47 \\
LibreOffice Impress & 47 \\
LibreOffice Writer & 23 \\
Multi-Apps & 101 \\
OS & 24 \\
Thunderbird & 15 \\
VLC & 17 \\
VS Code & 23 \\
\midrule
\textbf{Total} & \textbf{369} \\
\bottomrule
\end{tabular}
}
% Caption 放在底部
\caption{Statistics of the OSWorld-verified dataset across different domains.}
\label{tab:osworld_stats}
\end{table}

\begin{table}[t!]
\centering
\resizebox{0.48\textwidth}{!}{%
\begin{tabular}{lccccccc}
\toprule
& Windows & MacOS & Linux & iOS & Android & Web & Overall \\
\midrule
% L1 Section
% Multirow 跨越 6 行：3个标题行 + 3个数据行
\multirow{6}{*}{\textbf{L1}} 
 & \multicolumn{7}{c}{\cellcolor{graybg}\textbf{L1 - Easy}} \\
 & 271 & 84 & 196 & 115 & 307 & 221 & 1194 \\
 & \multicolumn{7}{c}{\cellcolor{graybg}\textbf{L1 - Medium}} \\
 & 271 & 84 & 196 & 115 & 307 & 221 & 1194 \\
 & \multicolumn{7}{c}{\cellcolor{graybg}\textbf{L1 - Hard}} \\
 & 271 & 84 & 196 & 115 & 307 & 221 & 1194 \\

\midrule
% L2 Section
% Multirow 跨越 4 行：2个标题行 + 2个数据行
\multirow{4}{*}{\textbf{L2}} 
 & \multicolumn{7}{c}{\cellcolor{graybg}\textbf{L2 - Basic}} \\
 & 271 & 345 & 191 & 314 & 356 & 310 & 1787 \\
 & \multicolumn{7}{c}{\cellcolor{graybg}\textbf{L2 - Advanced}} \\
 & 272 & 346 & 196 & 330 & 335 & 308 & 1787 \\

\midrule
% L3 & L4 Section
\textbf{L3} & 145 & 35 & 268 & - & 116 & 155 & 719 \\
\textbf{L4} & 35 & 35 & 101 & - & 30 & 47 & 248 \\

\midrule
\textbf{Total} & \textbf{1536} & \textbf{1013} & \textbf{1344} & \textbf{989} & \textbf{1758} & \textbf{1483} & \textbf{8123} \\
\bottomrule
\end{tabular}%
}
\caption{Statistics of the evaluation data in MMBench-GUI. }
\label{tab:mmbench_stats}
\end{table}

\begin{table}[t!]
\centering
\resizebox{0.45\textwidth}{!}{%
\begin{tabular}{lr}
\toprule
\textbf{Task Domain} & \textbf{Distribution (\%)} \\
\midrule
Shopping \& E-Commerce & 17.6 \\
Entertainment \& Media & 13.2 \\
Travel \& Transportation & 11.8 \\
Education & 11.0 \\
Technology & 8.8 \\
Government \& Services & 8.1 \\
Health \& Medical & 6.6 \\
Housing \& Real Estate & 5.9 \\
Finance \& Investment & 5.9 \\
Other & 5.9 \\
Jobs \& Careers & 2.9 \\
Food \& Recipes & 2.2 \\
\midrule
\textbf{Total Websites} & \textbf{136} \\
\bottomrule
\end{tabular}
}
\caption{Statistics distribution of task domains in the Online-Mind2Web dataset.}
\label{tab:mind2web_stats}
\end{table}

\section{Baselines}\label{app:baselines}
\paragraph{Baselines on OSWorld-Verified.} We compare our approach with a comprehensive suite of competitive baselines on the OSWorld benchmark: (1) Closed-Source LLMs: Claude 3.5 Sonnet, OpenAI o3, Doubao-1.5-Pro \cite{Guo2025Seed15VLTR}, and others, representing state-of-the-art generalist reasoning capabilities. (2) Open-Source GUI Agents: The UI-TARS family \cite{Qin2025UITARSPA}, ARPO \cite{lu2025ARPO}, OpenCUA series \cite{Wang2025OpenCUAOF}, and GUI-Owl-7B \cite{ye2025mobileagentv3fundamentalagentsgui}, which are specialized models for GUI interaction. (3) Agent Frameworks: Agent S2 \cite{Agashe2024AgentSA}, a compositional framework employing a Mixture-of-Grounding technique and Proactive Hierarchical Planning to delegate cognitive tasks; and Jedi \cite{xie2025jedi}, which utilizes multi-scale models trained on large-scale synthetic grounding data to enhance agentic capabilities. (4) Training Methodologies: We benchmark against various strategies including GRPO \cite{Shao2024DeepSeekMathPT}, SFT (training on the converted expert traces, and the conversion prompt is shown in Appendix), RL w/ SFT Loss (using SFT loss during RL training), and SFT+RL (a two-stage training process that continues RL training after SFT). We also compare with LUFFY \cite{Yan2025luffy}, which augments RLVR with off-policy reasoning traces via Mixed-Policy GRPO and policy shaping, and Trace Replacement, which differs from LUFFY by replacing trajectories with off-policy traces only upon group failure and calculating the importance ratio using the old policy likelihood rather than a constant (we build BEPA based on the simple trace replacement). For all trace replacement based methods, we use the converted expert traces $\mathcal{D}_{\text{conv}}$ as the off-policy data.

\paragraph{Baselines on MMBench-GUI.} For the MMBench-GUI benchmark, we assess cross-platform generalization using: (1) Proprietary Models: GPT-4o and Claude 3.7; (2) Native GUI Models: Qwen2.5-VL-72B \cite{Bai2025Qwen25VLTR}, Aguvis-72B \cite{Xu2024AguvisUP}, and the UI-TARS series (1.5-7B and 72B-DPO) ; and (3) Modular Agents: Planner-Grounder combinations such as GPT-4o paired with UGround-V1-7B \cite{Qian2025UGroundTU} or UI-TARS-1.5-7B.

\paragraph{Baselines on Online-Mind2Web.} We categorize these baselines into three groups: (1) Proprietary End-to-End Agents, including the industry-leading OpenAI Operator , Google Computer Use, and Claude 3.7 Computer Use, which interact with the browser via raw pixel or accessibility tree inputs; (2) Native GUI Agents, such as UITARS1.5-7B \cite{Qin2025UITARSPA} and ACT-1 \cite{Adept2022ACT1}, which are fine-tuned specifically for grounding and action generation on GUI screenshots; and (3) Modular Agent Frameworks, which decouple planning and execution, including SeeAct \cite{Zheng2024SeeAct}, Agent-E \cite{Abuelsaad2024AgentE}, Navigator (LaVague) \cite{LaVague2025Navigator}, and Browser Use \cite{Muller2024BrowserUse}.

% =========================================================
% Appendix B. Baseline Implementation Details
% =========================================================

\section{Implementation Details}\label{app:implementation_details}

The learning rate is set to $1\times10^{-6}$, and training is conducted with a batch size of $16$ over $256$ parallel virtual environments for $4$ epochs. Following DAPO \cite{Yu2025DAPOAO}, we use asymmetric clipping thresholds with $\epsilon_{\text{low}}=0.2$ and $\epsilon_{\text{high}}=0.3$ to balance exploration and exploitation. We remove the KL divergence regularization term and therefore do not need a reference model during optimization. The optimizer is AdamW \cite{Loshchilov2017Adamw} with a learning rate of $1\times10^{-6}$, and the training-time sampling temperature is fixed to $1.0$. 

\subsection{SFT on Converted Expert Traces}\label{app:sft_details}
We train the SFT baseline on the \emph{converted} expert trace set $\mathcal{D}_{\text{conv}}$, consisting of $125$ successful OSWorld tasks (one converted successful trajectory per task). We then construct step-level instruction--response pairs from the same converted expert traces.
Concretely, for each expert task $x$ and each step $t$ in its converted trajectory, we create one training pair $(u_{x,t}, y_{x,t})$, where $u_{x,t}$ is the step input (instruction plus the interaction context at step $t$) and $y_{x,t}$ is the corresponding target response (the expert step output).
This yields $1070$ step-level samples in total.

We perform supervised fine-tuning for $3$ epochs with the token-level cross-entropy loss. The learning rate is $1\times10^{-6}$ and the batch size is $16$.

\subsection{RL with SFT Loss (RL+SFT)}\label{app:rl_sft_details}
During GRPO training, at each iteration we may sample some expert tasks within the RL batch.
We then select the SFT samples \emph{only} from those expert tasks appearing in the current RL batch, and compute an auxiliary cross-entropy loss on their corresponding step-level pairs.

\paragraph{Joint objective and weighting.}
We optimize a combined objective with equal weights for the RL and SFT terms:
\begin{equation}
\mathcal{J}(\theta) \;=\; \mathcal{J}_{\mathrm{GRPO}}(\theta) \;+\; \lambda \, \mathcal{L}_{\mathrm{SFT}}(\theta),
\end{equation}
with $\lambda=1$. All GRPO hyperparameters follow Sec.~\ref{sec:experiments}.

\subsection{LUFFY}\label{app:luffy_details}
\paragraph{Algorithm.}
LUFFY augments GRPO with \emph{Mixed-Policy GRPO} and \emph{policy shaping}.
For each task, a rollout group includes both on-policy trajectories and off-policy expert trajectories, no matter if the rollout traces all fail; advantages are computed by group-wise normalization over the union of rewards from on-/off-policy samples.

\paragraph{Off-policy importance weighting.}
In principle, the off-policy branch uses an importance ratio
$r_{t}=\pi_{\theta}(a_t\mid s_t)/\pi_{\phi}(a_t\mid s_t)$.
In practice, LUFFY sets $\pi_{\phi}=1$ for off-policy data (i.e., it does not compute teacher token probabilities) to avoid tokenizer/probability incompatibilities and to simplify using existing datasets; correspondingly, the off-policy branch does not apply the PPO-style clipping.

\paragraph{Policy shaping.}
To mitigate entropy collapse and under-training on low-probability but critical tokens in off-policy traces, LUFFY applies a shaping transform to the off-policy weight:
$f(w)=\frac{w}{w+\gamma}$, with $\gamma=0.1$ by default .
Shaping is applied only to the off-policy branch; the on-policy branch follows GRPO.

\subsection{Log-Probability on S2 Raw Traces}
\label{sec:s2_logprob}

For the token-probability analysis in \cref{fig:histogram}, we evaluate
S2's raw traces under the base policy.
S2 is a multi-agent, multi-turn system, but UI-TARS expects a single
\texttt{system+user} input and a single assistant response.
We therefore reconstruct, for each step in an S2 trace, a synthetic
single-turn Worker call and compute the log-probability of the
corresponding S2 \texttt{executor\_plan}.

Concretely, each step provides:
(i) the global task instruction;
(ii) the current subtask and its description;
(iii) DAG-derived context (which subtasks are done and which are
remaining);
(iv) the current reflection (if any); and
(v) the current screenshot.
We instantiate S2's Worker procedural memory prompt with this context
and use the current subtask/reflection plus screenshot as the user
message (see \cref{fig:s2_worker_prompt} and an input/output example in \cref{fig:s2_worker_example}).
The S2 \texttt{executor\_plan} at that step is treated as the assistant
response.

\clearpage
\onecolumn
\begin{tcolorbox}[
    enhanced,
    breakable,
    enhanced jigsaw,
    colback=white,
    colframe=black,
    coltitle=white,
    colbacktitle=black,
    fonttitle=\bfseries,
    title=S2 Worker Prompt Template,
    boxrule=0.8pt,
    arc=6pt,
    top=6pt,
    bottom=6pt,
    left=8pt,
    right=8pt,
]
\small
\setstretch{1.10}

\textbf{System Message (Worker Procedural Memory).}

You are an expert in graphical user interfaces and Python code.
You are responsible for executing the current subtask:
\verb|`{current_subtask}`|
of the larger goal:
\verb|`{task_instruction}`|.

IMPORTANT:
The subtasks \verb|['{done_tasks}']| have already been done.
The future subtasks \verb|['{future_tasks}']| will be done later.
You must \textbf{only} perform the current subtask
\verb|`{current_subtask}`| and must not attempt future subtasks.

You are working in \verb|{platform}|.
You are provided with:
\begin{itemize}[leftmargin=1.2em]
    \item A screenshot of the current time step.
    \item The history of your previous interactions with the UI (summarized in reflection).
    \item Access to the following class and methods to interact with the UI:
\begin{lstlisting}[basicstyle=\ttfamily\small, frame=single]
class Agent:
    {agent_api_methods}
\end{lstlisting}
\end{itemize}

Your response should be formatted with the following sections:
\begin{itemize}[leftmargin=1.2em]
    \item \textbf{(Previous action verification)}:
    analyze whether the previous action was successful.
    \item \textbf{(Screenshot Analysis)}:
    describe the current state of the desktop and open applications.
    \item \textbf{(Next Action)}:
    decide on the next action in natural language.
    \item \textbf{(Grounded Action)}:
    output a single Python call using the \texttt{Agent} API:
\begin{lstlisting}[basicstyle=\ttfamily\small, frame=single]
```python
agent.xxx(...)
```

\end{lstlisting}
\end{itemize}

\vspace{0.4em}
\textbf{User Message (per step).}

For the first step of a subtask:
\begin{lstlisting}[basicstyle=\ttfamily\small, frame=single]
Remember only complete the subtask: {current_subtask}.
You can use this extra information for the current subtask:
{current_subtask_info}.
Text Buffer = [].
\end{lstlisting}

For subsequent steps (with reflection):
\begin{lstlisting}[basicstyle=\ttfamily\small, frame=single]
You may use this reflection on the previous action and trajectory:
{reflection_text}
Text Buffer = [].
\end{lstlisting}

In both cases, the user message is paired with the current screenshot as the vision input.

\end{tcolorbox}

\captionof{figure}{S2 Worker prompt template used to reconstruct single-turn
inputs for log-probability evaluation of S2 raw traces under UI-TARS1.5-7B.}
\label{fig:s2_worker_prompt}
\twocolumn

\clearpage
\onecolumn
\begin{tcolorbox}[
    enhanced,
    breakable,
    enhanced jigsaw,
    colback=white,
    colframe=black,
    coltitle=white,
    colbacktitle=black,
    fonttitle=\bfseries,
    title=Example Input/Output for Raw S2 Trace (Step 1),
    boxrule=0.8pt,
    arc=6pt,
    top=6pt,
    bottom=6pt,
    left=8pt,
    right=8pt,
]
\small
\setstretch{0.7}

\textbf{System Message.}

You are an expert in graphical user interfaces and Python code.
You are responsible for executing the current subtask:
\texttt{\textcolor{blue}{`Focus Chrome window`}}
of the larger goal:
\texttt{\textcolor{blue}{`Can you make my computer bring back the last tab I shut down?`}}.

IMPORTANT:
The subtasks \texttt{\textcolor{blue}{['']}} have already been done.
The future subtasks
\texttt{\textcolor{blue}{['Restore last closed tab']}}
will be done in the future by me.
You must only perform the current subtask:
\texttt{\textcolor{blue}{`Focus Chrome window`}}.
Do not try to do future subtasks.

You are working in \texttt{\textcolor{blue}{linux}}.
You must only complete the subtask provided and not the larger goal.

You are provided with:
\begin{itemize}[leftmargin=1.2em,itemsep=0.2em]
    \item A screenshot of the current time step.
    \item The history of your previous interactions with the UI.
    \item Access to an \texttt{Agent} API to interact with the UI.
\end{itemize}

Your response should be formatted with the following sections:

\begin{itemize}[leftmargin=1.2em,itemsep=0.2em]
    \item \textbf{(Previous action verification)}:
    analyze whether the previous action was successful.
    \item \textbf{(Screenshot Analysis)}:
    describe the current state of the desktop and open applications.
    \item \textbf{(Next Action)}:
    decide on the next action in natural language.
    \item \textbf{(Grounded Action)}:
    output a single Python call using the \texttt{Agent} API, e.g.
\begin{lstlisting}[basicstyle=\ttfamily\small, frame=single]
```python
agent.click("...", 1, "left")
```
\end{lstlisting}
\end{itemize}

\vspace{0.6em}
\textbf{User Message (instance-specific).}

\begin{lstlisting}[basicstyle=\ttfamily\small, frame=single]
Remember only complete the subtask: Focus Chrome window
You can use this extra information for completing the current subtask:
- Click an empty spot inside the Chrome window
  - Purpose: make Chrome the active application.
Text Buffer = [].
\end{lstlisting}

\vspace{0.4em}
\textbf{Screenshot.}

\begin{itemize}[leftmargin=1.2em,itemsep=0.2em]
    \item Input image: \texttt{\textcolor{blue}{step 0's screenshot}}
\end{itemize}

\vspace{0.6em}
\textbf{\textcolor{blue}{Ground-Truth Assistant Response (\texttt{executor\_plan}).}}

\begin{lstlisting}[basicstyle=\ttfamily\small, frame=single, breaklines=true]
(Previous action verification)
No previous action has been taken yet, so nothing to verify.

(Screenshot Analysis)
The desktop shows an Ubuntu environment. A Google Chrome window is
present in the center displaying the Airbnb webpage; it appears to be
the active application (the top bar reads "Google Chrome"). To ensure it
is definitely in focus for later steps, we should click anywhere within
the webpage that doesn't interact with page elements.

(Next Action)
Click an empty spot inside the current Chrome window (e.g., a blank
portion of the Airbnb page) to guarantee Chrome is focused.

(Grounded Action)
```python
agent.click("A blank area in the middle of the Airbnb webpage within the Google Chrome window, away from any buttons or links", 1, "left")
```
\end{lstlisting}

\end{tcolorbox}

\captionof{figure}{Example single-turn input/output used to compute
token-level log-probabilities for an S2 Worker step under
UI-TARS1.5-7B (Step 1 of the ``Focus Chrome window'' subtask).}
\label{fig:s2_worker_example}
\twocolumn

\begin{table*}[t]
\centering
\small
\begin{tabular}{lll}
\toprule
\textbf{Environment} & \textbf{Action} & \textbf{Definition} \\
\midrule

\textbf{Shared} & Click(x, y) & Clicks at coordinates (x, y). \\
& Drag(x1, y1, x2, y2) & Drags from (x1, y1) to (x2, y2). \\
& Scroll(x, y, direction) & Scrolls at (x, y) in the given direction. \\
& Type(content) & Types the specified content. \\
& Wait() & Pauses for a brief moment. \\
& Finished() & Marks the task as complete. \\
& CallUser() & Requests user intervention. \\

\midrule
\textbf{Desktop} & Hotkey(key) & Presses the specified hotkey. \\
& LeftDouble(x, y) & Double-clicks at (x, y). \\
& RightSingle(x, y) & Right-clicks at (x, y). \\

\midrule
\textbf{Mobile} & LongPress(x, y) & Long presses at (x, y). \\
& PressBack() & Presses the ``back'' button. \\
& PressHome() & Presses the ``home'' button. \\
& PressEnter() & Presses the ``enter'' key. \\

\bottomrule
\end{tabular}
\caption{The UITARS action space for different platforms. For OSWorld, the agent typically uses shared and desktop actions.}
\label{tab:action-space}
\end{table*}

\begin{table*}[t]
\centering
\small
\begin{tabular}{l p{0.36\linewidth} p{0.40\linewidth}}
\toprule
\textbf{Agent Action} & \textbf{Description} & \textbf{Arguments} \\
\midrule
click & Click on an element. &
\texttt{element\_description}, \texttt{num\_clicks}, \texttt{button\_type}, \texttt{hold\_keys} \\
type & Type text into an element. &
\texttt{element\_description}, \texttt{text}, \texttt{overwrite}, \texttt{enter} \\
scroll & Scroll within an element. &
\texttt{element\_description}, \texttt{clicks}, \texttt{shift} \\
hotkey & Press a hotkey combo. &
\texttt{keys} \\
hold\_and\_press & Hold keys and press others. &
\texttt{hold\_keys}, \texttt{press\_keys} \\
drag\_and\_drop & Drag and drop between elements. &
\texttt{element\_description\_1}, \texttt{element\_description\_2}, \texttt{hold\_keys} \\
save\_to\_knowledge & Save data to a per-task memory. &
\texttt{text} \\
switch\_applications & Switch to another app. &
\texttt{app\_name} \\
highlight\_text\_span & Highlight a text span. &
\texttt{starting\_phrase}, \texttt{ending\_phrase} \\
set\_cell\_values & Set tabular cell values. &
\texttt{cell\_values}, \texttt{app\_name}, \texttt{sheet\_name} \\
wait & Wait for some time. &
\texttt{time} \\
done & Mark subtask as success. &
None \\
fail & Mark subtask as failure. &
None \\
\bottomrule
\end{tabular}
\caption{\textbf{Agent S2 action space, descriptions, and arguments.}}
\label{tab:s2-action-space}
\end{table*}

\subsection{Codebase: Verl-GUI} \label{subsec:arch}

As part of the new assets released with this work, we propose \texttt{Verl-GUI}, a highly scalable distributed training framework for long-horizon, multi-turn vision-language GUI agent training. Our \texttt{Verl-GUI} is built upon the veRL framework~\cite{sheng2025hybridflow} and extends it with several features to enable scalable reinforcement learning for agentic tasks involving complex computer environments. Verl-GUI supports both synchronous (default) and asynchronous training modes, comprising approximately 212K lines of code across 714 Python files, representing an 83\% increase over the original veRL framework (116K lines, 543 files). Key capabilities of our framework include:
\paragraph{Heterogeneous Cluster Architecture.}
Verl-GUI completely separates trainer and rollout into independent Ray clusters in both sync and async modes, enabling deployment across heterogeneous compute resources. Nodes with high-bandwidth interconnects (InfiniBand, NVLink) are allocated for training where gradient synchronization demands intensive inter-GPU communication, while nodes with standard PCIe connectivity suffice for rollout where environment interactions are largely independent.

\paragraph{Multiple Storage Backends.}
The framework supports Azure Blob Storage, NAS, and local filesystems through a unified abstraction layer, enabling seamless deployment across cloud and on-premise infrastructure.

\paragraph{Async Task Queue.}
In asynchronous mode, Verl-GUI dynamically maintains a task queue for the rollout cluster to consume, enabling decoupled and non-blocking task processing where rollout and training proceed independently.

\paragraph{K-round Rollout Processing.}
The system intelligently splits batches across multiple rounds when the trainer's global batch size exceeds rollout cluster capacity. For example, if the trainer requires 128 samples but the rollout cluster can only process 64 concurrently, the system automatically executes $K=2$ rounds, enabling flexible scaling regardless of cluster size mismatches.

\paragraph{Scalable Parallel Environment Execution.}
The number of concurrent environments scales with rollout cluster compute capacity (i.e., $\text{max\_envs} = \text{num\_gpus} \times \text{batch\_per\_gpu}$), with Ray-based orchestration and automatic Docker cleanup.

\paragraph{Service-oriented Rollout Orchestration.}
Verl-GUI employs modular components including CheckpointManager, EnvWorkerPool, RolloutService, and ValidationAdapter for clean separation of concerns.

% AsyncVerl Implementation Details for Paper
% Usage: \input{asyncverl_implementation.tex}
% \onecolumn
% AsyncVerl Implementation Details for Paper
% Usage: \input{asyncverl_implementation.tex}

\begin{figure*}[t]
    \centering
    \includegraphics[width=0.9\textwidth]{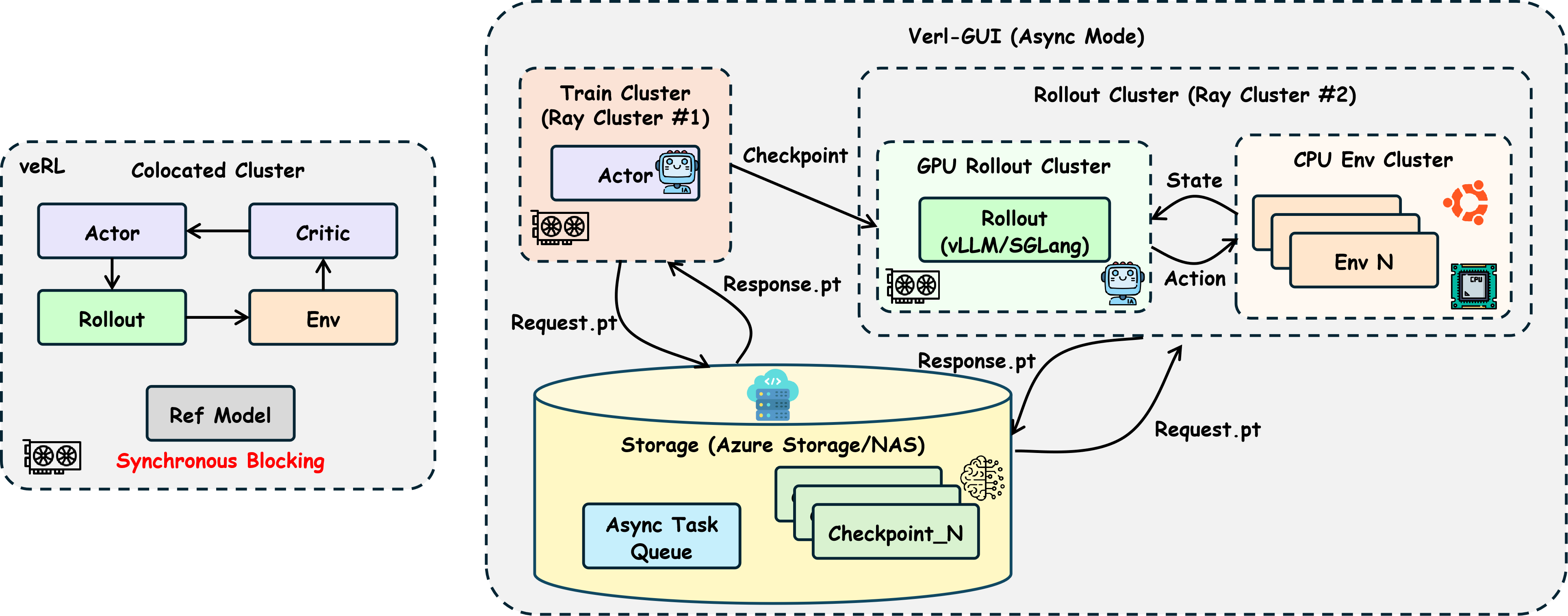}
    \caption{\textbf{GUI agent training framework comparison.} \textbf{Left:} veRL uses a single colocated cluster where all workers (Actor, Critic, Rollout, Env) share resources and communicate synchronously, causing blocking during environment interactions. \textbf{Right:} Verl-GUI separates training and rollout into independent Ray clusters in both sync and async modes, enabling heterogeneous hardware allocation (IB/NVLink nodes for training, PCIe nodes for rollout). The rollout cluster further divides into GPU sub-cluster (for model inference) and CPU sub-cluster (for Docker-based environment execution). In async mode, an async task queue dynamically maintains tasks for decoupled consumption.}
    \label{fig:verl_gui_architecture}
\end{figure*}

% \twocolumn

\section{Interface and Prompt Templates}\label{app:interface_prompts}

\subsection{UI-TARS Action Space}\label{app:action_space}
We adopt the UI-TARS action interface throughout training and evaluation.
\Cref{tab:action-space} lists the complete action space and the parameterization used in our implementation.

\subsection{Agent S2 Action Space}\label{app:s2_action_space}
We provide the expert's action space.
\Cref{tab:s2-action-space} lists the full action space together with natural-language descriptions and argument specifications.

\subsection{Prompt Templates}\label{app:prompt_templates}
We use a unified prompt format across all methods unless otherwise stated.

\paragraph{System prompt.}
The system prompt that constrains the model to output \texttt{Thought} and \texttt{Action} is shown in \cref{fig:prompt_sys}.

\paragraph{Plan summarization prompt.}
To abstract an expert trajectory into a compact natural-language plan $p_x=\phi(\tau_x^{E})$ for LEVEL-1 self-rolling (Sec.~\ref{sec:selfroll}), we use the plan summarization template in \cref{fig:prompt_plan_summary}.

\paragraph{Trace conversion prompt.}
To convert framework-produced expert traces into the end-to-end policy format (i.e., instruction--response pairs compatible with UI-TARS), we use the conversion prompt in \cref{fig:prompt_convertion}.
Converted traces form $\mathcal{D}_{\text{conv}}$ for SFT/RL+SFT baselines (Appendix~\ref{app:sft_details}--\ref{app:rl_sft_details}), and also serve as the expert source $\mathcal{D}_E$ used in BEPA initialization (Sec.~\ref{sec:selfroll}).

\clearpage
\onecolumn
\begin{center}
\begin{tcolorbox}[
    enhanced,
    breakable,
    enhanced jigsaw,
    width=\textwidth,   % 双栏总宽度
    colback=white,
    colframe=black,
    coltitle=white,
    colbacktitle=black,
    fonttitle=\bfseries,
    title=System Prompt Template for UITARS,
    boxrule=0.8pt,
    arc=6pt,
    top=6pt,
    bottom=6pt,
    left=8pt,
    right=8pt
]
{\small
\setstretch{1.10}

You are a GUI agent. You are given a task and your action history, with screenshots.
You need to perform the next action to complete the task.

\vspace{0.6em}
\textbf{Output Format}
\begin{quote}
\verb|Thought: ...|\\
\verb|Action: ...|
\end{quote}

\vspace{0.4em}
\textbf{Action Space}
\begin{lstlisting}[basicstyle=\ttfamily\small, frame=single]
click(start_box='<|box_start|>(x1,y1)<|box_end|>')
left_double(start_box='<|box_start|>(x1,y1)<|box_end|>')
right_single(start_box='<|box_start|>(x1,y1)<|box_end|>')
drag(start_box='<|box_start|>(x1,y1)<|box_end|>',
     end_box='<|box_start|>(x3,y3)<|box_end|>')
hotkey(key='')
type(content='xxx')  # Use \', \" and \n in content
scroll(start_box='<|box_start|>(x1,y1)<|box_end|>',
       direction='down or up or right or left')
wait()
finished(content='xxx')
\end{lstlisting}

\vspace{0.4em}
\textbf{Note}
\begin{itemize}[leftmargin=1.4em]
    \item Use English in both the \texttt{Thought} and \texttt{Action} parts.
    \item In the \texttt{Thought} part, write a brief plan and end with
    one sentence that clearly summarizes your next action and its target element.
\end{itemize}

\vspace{0.4em}
\textbf{User Instruction}
\begin{quote}
\verb|{instruction}|
\end{quote}

} % end small + setstretch
\end{tcolorbox}
\end{center}

\captionof{figure}{The system prompt used to instruct the GUI agent to produce thought and action.}
\label{fig:prompt_sys}
\twocolumn

\clearpage
\onecolumn
\begin{tcolorbox}[
    enhanced,
    breakable,
    enhanced jigsaw,
    colback=white,
    colframe=black,
    coltitle=white,
    colbacktitle=black,
    fonttitle=\bfseries,
    title=Prompt Template for S2$\rightarrow$UI-TARS Trace Conversion,
    boxrule=0.8pt,
    arc=6pt,
    top=6pt,
    bottom=6pt,
    left=8pt,
    right=8pt,
]
\small                 % 字体调小
\setstretch{1.10}       % 行间距调节，1.08~1.15 都很舒服

\textbf{You are an expert at converting S2 GUI automation traces into UI-TARS format.}

\vspace{0.4em}
\textbf{Your Task:} Transform S2's \placeholder{executor\_plan}, \placeholder{reflection}, and
\placeholder{pyautogui\_action} into UI-TARS format with both natural language \texttt{Thought}
and action notation.

\vspace{0.4em}
\textbf{Input (S2 Format):}
\begin{itemize}[leftmargin=1.2em]
    \item \texttt{executor\_plan}: screenshot analysis and next-action planning.
    \item \texttt{reflection}: reflection on current progress (may be \texttt{null}).
    \item \texttt{pyautogui\_action}: the executed PyAutoGUI code (e.g.,\\
    \texttt{pyautogui.click(1265, 245, clicks=1, button='left')}).
\end{itemize}

\vspace{0.2em}
\textbf{Output (UI-TARS Format):}
\begin{lstlisting}[basicstyle=\ttfamily\small, frame=single]
Thought: {{natural language reasoning in 1-3 sentences}}
Action: {{UI-TARS action notation}}
\end{lstlisting}

\vspace{0.2em}
\textbf{UI-TARS Action Space:} you must output actions using the following templates:
\begin{lstlisting}[basicstyle=\ttfamily\small, frame=single]
click(start_box='<|box_start|>(x1,y1)<|box_end|>')
left_double(start_box='<|box_start|>(x1,y1)<|box_end|>')
right_single(start_box='<|box_start|>(x1,y1)<|box_end|>')
drag(start_box='<|box_start|>(x1,y1)<|box_end|>',
     end_box='<|box_start|>(x3,y3)<|box_end|>')
hotkey(key='key1 key2 ...')
type(content='xxx')  # Use \', \", and \n with escaping.
scroll(start_box='<|box_start|>(x1,y1)<|box_end|>',
       direction='down or up or right or left')
wait()  # Sleep for 5s and take a screenshot
finished(content='xxx')
\end{lstlisting}

\vspace{0.2em}
\textbf{Critical Requirements:}
\begin{enumerate}[leftmargin=1.2em]
    \item \textbf{Box tokens:} you \emph{must} include
    \boxtoken{<|box\_start|>} and \boxtoken{<|box\_end|>} around
    coordinates in \texttt{start\_box} and \texttt{end\_box}.
    \item \textbf{Coordinate format:}
    \boxtoken{<|box\_start|>(x,y)<|box\_end|>} where \texttt{x,y} are numbers.
    \item \textbf{Hotkey:}
    space-separated keys (e.g., \texttt{'ctrl c'}, \texttt{'ctrl shift b'},
    or a single key like \texttt{'pagedown'}).
    \item \textbf{Scroll:} \texttt{start\_box} is optional---include it only
    if the PyAutoGUI action specifies coordinates.
    \item \textbf{Escaping:} use \texttt{\textbackslash'} for single quotes,
    \texttt{\textbackslash"} for double quotes, and
    \texttt{\textbackslash n} for newlines.
\end{enumerate}

\vspace{0.2em}
\textbf{Thought Guidelines:}
\begin{enumerate}[leftmargin=1.2em]
    \item Use first-person perspective (\emph{``I see...'',
    ``I notice...'',
    ``Let me...''}).
    \item Be natural and conversational, not robotic.
    \item Keep it concise.
    \item Focus on what you observe and what you plan to do next.
\end{enumerate}

\vspace{0.2em}
\textbf{Example (simplified):}
Given \placeholder{executor\_plan}, \placeholder{reflection}, and
\placeholder{pyautogui\_action}, output:
\begin{lstlisting}[basicstyle=\ttfamily\small, frame=single]
Thought: I see the Chrome window open on the Puzzle Game 2048 page.
        To access more options, I should open the browser menu.
Action: click(start_box='<|box_start|>(1265,245)<|box_end|>')
\end{lstlisting}

\end{tcolorbox}
\captionof{figure}{Prompt for converting Agent S2's traces into UI-TARS-style thoughts and actions.}
\label{fig:prompt_convertion}
\twocolumn
\clearpage
\onecolumn
\begin{tcolorbox}[
    enhanced,
    breakable,
    enhanced jigsaw,
    colback=white,
    colframe=black,
    coltitle=white,
    colbacktitle=black,
    fonttitle=\bfseries,
    title=Prompt Template for Plan Summary,
    boxrule=0.8pt,
    arc=6pt,
    top=6pt,
    bottom=6pt,
    left=8pt,
    right=8pt,
]
{\small
\setstretch{1.10}

\textbf{You are an expert at converting technical GUI automation traces into clear, actionable step-by-step plans.}

\vspace{0.4em}
\textbf{Your Task:}
Analyze the complete execution trace and generate a clean, effective step-by-step plan.
The trace contains \emph{all} actions that were performed, including exploratory actions, mistakes, and corrections.
Your job is to extract only the \emph{effective} actions that contributed to task completion.

\vspace{0.4em}
\textbf{Critical Requirements:}
\begin{enumerate}[leftmargin=1.4em]
    \item \textbf{Filter out exploratory / trial-and-error steps:} remove actions that
    \begin{itemize}[leftmargin=1.2em]
        \item were exploratory attempts that did not work,
        \item were mistakes that needed correction,
        \item were redundant or duplicated,
        \item were verification steps that did not contribute to progress.
    \end{itemize}
    \item \textbf{Keep only effective steps:} include actions that
    \begin{itemize}[leftmargin=1.2em]
        \item successfully moved the task forward,
        \item were necessary for completion,
        \item represent the optimal path to achieve the goal.
    \end{itemize}
    \item \textbf{Consolidate related actions:} if multiple execution steps accomplish one logical action, combine them into a single step.
\end{enumerate}

\vspace{0.4em}
\textbf{Output Format:}
Generate a clean, numbered step-by-step plan that:
\begin{enumerate}[leftmargin=1.4em]
    \item starts with a brief task description,
    \item lists each action as a simple, clear step,
    \item uses action verbs (click, type, select, drag, press, etc.),
    \item removes technical details and purposes,
    \item combines related sub-steps when logical,
    \item maintains the essential flow of actions,
    \item contains \emph{only} effective actions (no trial-and-error or exploration).
\end{enumerate}

\vspace{0.2em}
\textbf{Example:}
\begin{lstlisting}[basicstyle=\ttfamily\small, frame=single]
Task: Create a desktop shortcut for the current website

1. Press Ctrl+L to focus the address bar
2. Type "mathsisfun.com/games/2048.html" and press Enter
3. Click the three-dot Chrome menu button (top-right)
4. Click "Cast, save and share"
5. Click "Create shortcut..."
6. Click the "Create" button in the dialog
\end{lstlisting}

\vspace{0.4em}
\textbf{Now convert the following trace into a step-by-step plan:}
} % end small + setstretch

\end{tcolorbox}

\captionof{figure}{Prompt for converting S2 execution traces into a concise step-by-step plan.}
\label{fig:prompt_plan_summary}
\twocolumn

\section{Full Benchmark Results}\label{app:full_results}

We reports full quantitative results for each benchmark used in our study for three benchmarks. \cref{tab:osworld} reports per-domain success rates (\%) on OSWorld-Verified for all compared methods. 
For MMBench-GUI, we report detailed results for each level in \cref{tab:L1,tab:L2}.
\cref{tab:mind2web} reports success rates (SR \%) for all methods evaluated on Online-Mind2Web.

\begin{table*}[t]
\centering
\begin{adjustbox}{max width=\textwidth}
\begin{tabular}{l|cccccccccc|>{\columncolor{avgyellow}}c}
\toprule
\textbf{Method} & \textbf{Chrome} & \textbf{GIMP} & \textbf{Calc} & \textbf{Impress} & \textbf{Writer} & \textbf{Multi} & \textbf{OS} & \textbf{Thunderbird} & \textbf{VLC} & \textbf{VSCode} & \textbf{Overall} \\
\midrule
\rowcolor{categoryblue}
\multicolumn{12}{c}{\textbf{\textit{Agentic Framework}}} \\
Agent S2 & 29.06 & 42.69 & 13.16 & 27.38 & 36.69 & \textbf{23.62} & \textbf{67.28} & 64.87 & 42.41 & 52.65 & 33.00 \\
Jedi-7B w/ o3 & \textbf{52.09} & \textbf{65.38} & \textbf{29.79} & \textbf{42.53} & \textbf{65.22} & 20.43 & 50.00 & \textbf{73.33} & \textbf{47.06} & 56.52 & \textbf{42.40} \\
\midrule
\rowcolor{categoryblue}
\multicolumn{12}{c}{\textbf{\textit{General Model}}} \\
OpenAI o3 & 6.52 & 11.54 & 0.00 & 0.00 & 4.35 & 11.83 & 37.50 & 6.67 & 10.88 & 13.04 & 9.10 \\
Doubao-1.5-Think-Vision & 47.83 & 50.00 & 25.53 & 36.15 & 43.48 & 6.45 & 33.33 & 66.67 & 35.29 & 47.83 & \textbf{31.90} \\
Claude-4-Sonnet & 36.87 & 46.15 & 17.02 & 36.17 & 43.48 & 9.68 & 37.50 & 66.67 & 38.47 & 60.87 & 31.20 \\
\midrule
\rowcolor{categoryblue}
\multicolumn{12}{c}{\textbf{\textit{Specialized Model}}} \\
UITARS-72B-DPO & 37.60 & 61.54 & 8.70 & 25.53 & 30.43 & 4.92 & 33.33 & 40.00 & 17.65 & 39.13 & 24.00 \\
Computer-Use-Preview & 36.87 & 34.62 & 10.64 & 25.45 & 30.43 & 10.75 & 45.83 & 46.67 & 29.41 & 47.83 & 26.00 \\
GUI-Owl-7B & 41.22 & \textbf{65.38} & 17.02 & 19.06 & \textbf{52.17} & 9.68 & \textbf{50.00} & \textbf{66.67} & 29.41 & \textbf{65.22} & \textbf{32.10} \\
OpenCUA-7B & 36.87 & 50.00 & 10.64 & 36.15 & 26.09 & 6.52 & 30.43 & 53.33 & 29.41 & 43.48 & 24.30 \\
ARPO & 22.02 & 50.00 & 12.77 & 19.57 & 39.12 & 8.11 & 37.50 & 40.00 & 23.53 & 39.13 & 22.97 \\
UITARS1.5-7B & 22.73 & 50.00 & 8.51 & 25.33 & 39.12 & 7.08 & 34.78 & 33.33 & 23.53 & 56.52 & 23.66 \\
\midrule
\rowcolor{categoryblue}
\multicolumn{12}{c}{\textbf{\textit{UITARS1.5-7B+}}} \\
SFT & 29.75 & 37.84 & 10.46 & 14.66 & 25.65 & 4.23 & 25.66 & 19.68 & 11.58 & 42.77 & 18.00 \\
GRPO & 28.15 & 53.18 & 15.84 & 20.81 & 36.98 & 9.75 & 27.74 & 21.27 & 25.72 & 60.11 & 24.50 \\
RL+SFT & 26.63 & 46.23 & 14.80 & 19.62 & 30.36 & 8.22 & 26.15 & 20.08 & 24.29 & 52.31 & 22.00 \\
SFT+RL & 21.63 & 30.94 & 4.70 & 7.12 & 19.22 & 4.74 & 41.44 & 14.74 & 33.22 & 24.03 & 15.20 \\
Trace Replacement & 29.16 & 58.35 & 12.90 & 15.07 & 35.17 & 10.38 & 33.71 & 40.45 & 23.80 & 67.77 & 25.50 \\
LUFFY & 28.93 & 36.16 & 11.43 & 25.72 & 40.86 & 5.78 & 51.82 & 35.83 & 34.51 & 70.08 & 25.80 \\
\textbf{BEPA} (ours) & \textbf{47.80} & \textbf{69.20} & 10.64 & \textbf{31.90} & 34.80 & 8.60 & \textbf{54.20} & 40.00 & \textbf{29.41} & \textbf{73.90} & \textbf{32.40} \\
\bottomrule
\end{tabular}
\end{adjustbox}
\caption{\textbf{Performance comparison on OSWorld-Verified.} We report the best single-run success rates (\%) across 10 application domains. Methods are grouped into four categories: agentic frameworks with compositional architectures, general-purpose LLMs, specialized GUI models, and UITARS1.5-7B with various training strategies. BEPA achieves the highest overall performance among all UITARS1.5-7B+ variants, demonstrating the effectiveness of bi-level expert-to-policy assimilation.}
\label{tab:osworld}
\end{table*}
\begin{table*}[t!]
\centering
\resizebox{0.8\textwidth}{!}{%
\begin{tabular}{lccccccc}
\toprule
Model & Windows & macOS & Linux & iOS & Android & Web & Overall \\
\midrule
\rowcolor{graybg} \multicolumn{8}{c}{\textbf{Easy Level}} \\
GPT-4o & 62.47 & 62.38 & 67.89 & 58.52 & 56.41 & 58.51 & 60.16 \\
Claude-3.5 & 41.34 & 41.61 & 50.04 & 42.03 & 38.96 & 41.79 & 41.54 \\
Claude-3.7 & 34.66 & 39.37 & 49.05 & 42.76 & 37.45 & 40.80 & 39.08 \\
Qwen-Max-VL & 69.05 & 69.91 & 72.51 & 70.82 & 63.09 & 69.46 & 68.15 \\
Qwen2.5-VL-72B & 65.86 & 73.02 & 75.23 & 67.24 & 58.09 & 72.08 & 66.98 \\
UI-TARS-72B-DPO & 41.59 & 35.16 & 28.52 & 31.08 & 52.25 & 35.33 & 40.18 \\
InternVL3-72B & 74.67 & 79.16 & 78.72 & \textbf{83.57} & 80.10 & 81.18 & 79.15 \\
% LUFFY: Small boost from In-domain
LUFFY & 84.19 & 82.38 & 88.01 & 81.37 & 83.94 & 88.67 & 85.18 \\
% Base Model
UI-TARS1.5-7B & 83.82 & 82.28 & 84.03 & 81.22 & 83.64 & 85.12 & 83.69 \\
% GRPO: Slightly better than LUFFY
GRPO & 84.44 & 82.63 & 88.26 & 81.62 & 84.19 & 88.92 & 85.43 \\
% BEPA: Largest gap (+6.7% Overall), massive In-domain gains (+10%)
\textbf{BEPA (Ours)} & \textbf{88.94} & \textbf{87.13} & \textbf{94.26} & \textbf{86.12} & \textbf{88.69} & \textbf{94.92} & \textbf{90.46} \\

\midrule
\rowcolor{graybg} \multicolumn{8}{c}{\textbf{Medium Level}} \\
GPT-4o & 56.33 & 59.70 & 63.13 & 54.06 & 57.69 & 54.98 & 57.24 \\
Claude-3.5 & 39.28 & 45.97 & 47.63 & 44.57 & 42.03 & 34.33 & 41.26 \\
Claude-3.7 & 39.34 & 42.28 & 39.23 & 39.45 & 36.05 & 36.17 & 38.39 \\
Qwen-Max-VL & 63.40 & 66.90 & 73.85 & 68.02 & 63.66 & 64.59 & 65.44 \\
Qwen2.5-VL-72B & 66.29 & 72.73 & 72.63 & 59.27 & 66.24 & 68.24 & 67.45 \\
UI-TARS-72B-DPO & 38.83 & 37.14 & 41.60 & 41.72 & 54.74 & 31.55 & 41.77 \\
InternVL3-72B & 71.46 & 79.88 & 78.58 & 78.43 & 81.36 & 78.67 & 77.89 \\
% LUFFY
LUFFY & 95.33 & 83.97 & 89.88 & 80.67 & 89.09 & 83.89 & 88.50 \\
% Base
UI-TARS1.5-7B & 95.16 & 84.07 & 86.90 & 80.72 & 88.99 & 81.34 & 87.49 \\
% GRPO
GRPO & 95.58 & 84.22 & 90.13 & 80.92 & 89.34 & 84.14 & 88.75 \\
% BEPA: Medium gap (+4.4% Overall), solid In-domain gains (+7-8%)
\textbf{BEPA (Ours)} & \textbf{98.28} & \textbf{86.92} & \textbf{94.13} & \textbf{83.62} & \textbf{92.04} & \textbf{88.14} & \textbf{91.91} \\

\midrule
\rowcolor{graybg} \multicolumn{8}{c}{\textbf{Hard Level}} \\
GPT-4o & 60.69 & 60.38 & 52.42 & 45.27 & 50.93 & 50.83 & 53.49 \\
Claude-3.5 & 37.40 & 42.70 & 34.07 & 40.86 & 36.96 & 38.11 & 37.55 \\
Claude-3.7 & 32.99 & 34.48 & 31.97 & 39.20 & 36.96 & 38.92 & 35.65 \\
Qwen-Max-VL & 66.64 & 67.59 & 65.80 & 60.23 & 58.78 & 65.34 & 63.69 \\
Qwen2.5-VL-72B & 70.68 & 68.91 & 70.98 & 57.59 & 53.94 & 68.10 & 64.56 \\
UI-TARS-72B-DPO & 31.48 & 35.87 & 24.19 & 36.33 & 58.13 & 19.94 & 35.78 \\
InternVL3-72B & 75.08 & 76.19 & 77.44 & 70.37 & 75.73 & 78.11 & 75.70 \\
% LUFFY
LUFFY & 90.00 & 90.24 & 96.91 & 80.72 & 89.78 & 93.22 & 90.80 \\
% Base
UI-TARS1.5-7B & 90.03 & 90.54 & 95.43 & 80.97 & 89.88 & 92.17 & 90.44 \\
% GRPO
GRPO & 90.25 & 90.49 & 97.16 & 80.97 & 90.03 & 93.47 & 91.05 \\
% BEPA: Smallest gap (+2.2% Overall), still best due to In-domain saturation
\textbf{BEPA (Ours)} & \textbf{91.65} & \textbf{91.89} & \textbf{98.80} & \textbf{82.37} & \textbf{91.43} & \textbf{95.47} & \textbf{92.60} \\
\bottomrule
\end{tabular}%
}
\caption{\textbf{Performance on L1-GUI Content Understanding.} We report the success rates (\%) (SR) of various models across all platforms and the weighted average success rate.}
\label{tab:L1}
\end{table*}

\begin{table*}[t!]
\centering
\resizebox{\textwidth}{!}{%
\begin{tabular}{lcccccccccccccc}
\toprule
\multirow{2}{*}{Model} &
\multicolumn{2}{c}{Windows} &
\multicolumn{2}{c}{MacOS} &
\multicolumn{2}{c}{Linux} &
\multicolumn{2}{c}{iOS} &
\multicolumn{2}{c}{Android} &
\multicolumn{2}{c}{Web} &
\multicolumn{2}{c}{Avg.} \\
\cmidrule(lr){2-3}\cmidrule(lr){4-5}\cmidrule(lr){6-7}\cmidrule(lr){8-9}\cmidrule(lr){10-11}\cmidrule(lr){12-13}\cmidrule(lr){14-15}
& Basic & Adv. & Basic & Adv. & Basic & Adv. & Basic & Adv. & Basic & Adv. & Basic & Adv. & Basic & Adv. \\
\midrule
GPT-4o & 1.48 & 1.10 & 8.69 & 4.34 & 1.05 & 1.02 & 5.10 & 3.33 & 2.53 & 1.41 & 3.23 & 2.92 & 3.68 & 2.35 \\
Claude-3.7 & 1.48 & 0.74 & 12.46 & 7.51 & 1.05 & 0.00 & 13.69 & 10.61 & 1.40 & 1.40 & 3.23 & 2.27 & 5.55 & 3.75 \\
Qwen-Max-VL & 43.91 & 36.76 & 58.84 & 56.07 & 53.93 & 30.10 & 77.39 & 59.09 & 79.49 & 70.14 & 74.84 & 58.77 & 64.73 & 51.82 \\
Aguvis-7B-720P & 37.27 & 21.69 & 48.12 & 33.27 & 33.51 & 25.00 & 67.52 & 65.15 & 60.96 & 50.99 & 61.61 & 45.45 & 51.50 & 40.26 \\
ShowUI-2B & 9.23 & 4.41 & 24.06 & 10.40 & 25.13 & 11.73 & 28.98 & 19.70 & 17.42 & 8.73 & 22.90 & 12.66 & 21.29 & 11.27 \\
OS-Atlas-Base-7B & 36.90 & 18.75 & 44.35 & 21.68 & 31.41 & 13.27 & 74.84 & 48.79 & 69.60 & 46.76 & 61.29 & 35.39 & 53.06 & 30.77 \\
UGround-V1-7B & 66.79 & 38.97 & 71.30 & 48.55 & 56.54 & 31.12 & 92.68 & 70.91 & \textbf{93.54} & 70.99 & 88.71 & 64.61 & 78.26 & 54.19 \\
InternVL3-72B & 70.11 & 42.64 & 75.65 & 52.31 & 59.16 & 41.33 & 93.63 & 80.61 & 92.70 & 78.59 & 90.65 & 65.91 & 80.32 & 60.23 \\
Qwen2.5-VL-72B & 55.72 & 33.82 & 49.86 & 30.06 & 40.31 & 20.92 & 56.05 & 28.18 & 55.62 & 25.35 & 68.39 & 45.78 & 54.33 & 30.69 \\
Qwen2.5-VL-7B & 31.37 & 16.54 & 31.30 & 21.97 & 21.47 & 12.24 & 66.56 & 55.15 & 35.11 & 35.21 & 40.32 & 32.47 & 37.69 & 28.93 \\
% Base
UI-TARS1.5-7B & 76.75 & 45.59 & 77.39 & 52.89 & 67.02 & 40.31 & 92.04 & 70.00 & 93.26 & 71.83 & 87.74 & 61.36 & 83.72 & 58.57 \\
% 72B Reference
UI-TARS-72B-DPO & 78.60 & \textbf{51.84} & 80.29 & \textbf{62.72} & 68.59 & \textbf{51.53} & 90.76 & \textbf{81.21} & 92.98 & \textbf{80.00} & 88.06 & \textbf{68.51} & 84.50 & \textbf{67.49} \\
% LUFFY: Slight boost only
LUFFY & 76.95 & 45.49 & 77.59 & 52.79 & 71.72 & 43.01 & 92.24 & 69.90 & 93.46 & 71.73 & 92.44 & 64.06 & 85.18 & 59.26 \\
% GRPO: Boost in Linux/Web, flat elsewhere
GRPO & 77.25 & 45.79 & 77.89 & 53.09 & 72.02 & 43.31 & 92.54 & 70.20 & 93.76 & 72.03 & 92.74 & 64.36 & 85.48 & 59.56 \\
% BEPA: Massive In-domain boost, Solid OOD boost
\textbf{BEPA (Ours)} & \textbf{80.25} & 47.59 & \textbf{80.89} & 54.89 & \textbf{76.02} & 45.81 & \textbf{95.54} & 72.00 & \textbf{96.76} & 73.83 & \textbf{96.74} & 66.86 & \textbf{88.76} & 61.56 \\
\bottomrule
\end{tabular}%
}
\caption{\textbf{Performance on the L2-GUI Element Grounding.} `Adv.' stands for advanced. `Avg.' reports the weighted average score across platforms for each mode (Basic/Adv.).}
\label{tab:L2}
\end{table*}

\section{Extensibility}\label{app:expert_generalization}

\paragraph{Setup.}
To evaluate how BEPA generalizes across expert sources and base models, we instantiate BEPA with different combinations of experts and backbones on OSWorld-Verified.
On UITARS1.5-7B, we use Agent~S2, GUI-Owl-7B, or a hybrid configuration (S2+Owl), where for each task we \emph{prefer} the self-rolled success from Agent~S2 when available and fall back to the GUI-Owl trace otherwise (i.e., S2 takes precedence on tasks where both experts succeed). For GUI-Owl-7B's step by step plan extraction, we use the same plan summarization prompt as in \cref{fig:prompt_plan_summary}.
On OpenCUA-7B, we additionally train BEPA with Agent~S2 as the expert (\cref{tab:expert_generality}).

\paragraph{Expert diversity on UITARS1.5-7B.}
With Agent~S2 as the expert, BEPA reaches $32.40\%$ overall success, improving over GRPO (UITARS) at $24.50\%$.
Swapping the expert to GUI-Owl-7B still yields a strong $31.55\%$ overall success, with notable gains on Writer and TB (Thunderbird), indicating that BEPA can effectively exploit a different GUI-specialized expert.
Combining Agent~S2 and GUI-Owl (S2+Owl) further boosts performance to $34.71\%$ overall, capturing complementary strengths: S2 remains strong on Chrome, Impress, OS, and VSCode, while Owl provides substantial gains on Writer, Multi-app, TB, and VLC.

\paragraph{Portability across base policies.}
BEPA also transfers to a different base model.
On OpenCUA-7B, BEPA with Agent~S2 as the expert attains $31.98\%$ overall success, a large improvement over the OpenCUA baseline at $24.30\%$ and comparable to BEPA on UITARS1.5-7B.
Across both backbones, BEPA consistently turns stronger experts into tangible, per-domain gains on OSWorld-Verified, demonstrating that our bi-level assimilation mechanism is not tied to a specific base model or a single expert source.

\begin{table*}[t]
\centering
\begin{adjustbox}{max width=\textwidth}
\begin{tabular}{l|cccccccccc|>{\columncolor{avgyellow}}c}
\toprule
\textbf{Method} & \textbf{Chrome} & \textbf{GIMP} & \textbf{Calc} & \textbf{Impress} & \textbf{Writer} & \textbf{Multi} & \textbf{OS} & \textbf{TB} & \textbf{VLC} & \textbf{VSCode} & \textbf{Overall} \\
\midrule
\rowcolor{categoryblue}
\multicolumn{12}{c}{\textbf{\textit{Base Models \& Baselines}}} \\
UITARS-1.5-7B & 22.73 & 50.00 & 8.51 & 25.33 & 39.12 & 7.08 & 34.78 & 33.33 & 23.53 & 56.52 & 23.66 \\
OpenCUA-7B  & 36.87 & 50.00 & 10.64 & 26.09 & 36.15 & 6.52 & 30.43 & 53.33 & 29.41 & 43.48 & 24.30 \\
GRPO (UITARS) & 28.15 & 53.18 & 15.84 & 20.81 & 36.98 & 9.75 & 27.74 & 21.27 & 25.72 & 60.11 & 24.50 \\
\midrule
\rowcolor{categoryblue}
\multicolumn{12}{c}{\textbf{\textit{BEPA Variants (UITARS-1.5-7B)}}} \\
BEPA (w/ Agent S2) & \textbf{47.80} & \textbf{69.20} & 10.64 & \textbf{34.80} & 31.90 & 8.60 & \textbf{54.20} & 40.00 & 29.41 & \textbf{73.90} & 32.40 \\
BEPA (w/ GUI-Owl) & 36.96 & 61.54 & \textbf{12.77} & 25.53 & \textbf{52.17} & \textbf{12.80} & 41.67 & 60.00 & 35.29 & 60.87 & 31.55 \\
BEPA (S2 + Owl) & 46.50 & 68.00 & 12.50 & 33.50 & 50.00 & 9.95 & 52.00 & 58.00 & \textbf{36.00} & 72.00 & \textbf{34.71} \\
\midrule
\rowcolor{categoryblue}
\multicolumn{12}{c}{\textbf{\textit{BEPA Variants (OpenCUA-7B)}}} \\
BEPA (OpenCUA + S2) & 43.48 & 57.69 & 14.89 & 31.91 & 43.48 & 10.78 & 45.83 & \textbf{66.67} & 32.00 & 52.17 & 31.98 \\
\bottomrule
\end{tabular}
\end{adjustbox}
\caption{\textbf{Extensibility.} We report the success rates (\%) on OSWorld-Verified. The results show BEPA's effectiveness across different base models (UITARS, OpenCUA) and its ability to scale with hybrid expert guidance (S2 + Owl).}
\label{tab:expert_generality}
\end{table*}

\begin{table}[t!]
\centering
\resizebox{0.48\textwidth}{!}{%
\begin{tabular}{lcccc}
\toprule
Agent & Easy & Medium & Hard & Average SR \\
\midrule
Navigator & 90.1 & 76.2 & 71.1 & 78.7 \\
Google Computer Use & 77.1 & 71.3 & 55.4 & 69.0 \\
Operator & 83.1 & 58.0 & 43.2 & 61.3 \\
ACT-1 (2025-08-23) & 81.9 & 54.5 & 35.1 & 57.3 \\
Claude Computer Use 3.7 & 90.4 & 49.0 & 32.4 & 56.3 \\
ACT-1 (2025-07-16) & 65.1 & 46.2 & 23.0 & 45.7 \\
SeeAct & 60.2 & 25.2 & 8.1 & 30.7 \\
Browser Use & 55.4 & 26.6 & 8.1 & 30.0 \\
Claude Computer Use 3.5 & 56.6 & 20.3 & 14.9 & 29.0 \\
Agent-E & 49.4 & 26.6 & 6.8 & 28.0 \\
\midrule
% Base Model
UITARS1.5-7B & 46.91 & 18.88 & 5.26 & 23.28 \\
% LUFFY: Weak generalization (+1.5%)
LUFFY & 48.91 & 20.68 & 5.66 & 24.79 \\
% GRPO: Weak generalization (+1.5%)
GRPO & 49.41 & 20.38 & 5.76 & 24.81 \\
% BEPA: Strong generalization (+8.1%), surpassing SeeAct/Browser Use
\textbf{BEPA (Ours)} & \textbf{53.42} & \textbf{24.17} & \textbf{8.92} & \textbf{28.52} \\
\bottomrule
\end{tabular}%
}
\caption{\textbf{Performance comparison on Online-Mind2Web.} We report the success rates (\%) (SR) of various agents across three difficulty levels and the weighted average success rate. BEPA demonstrates superior generalization on unseen websites compared to baselines, achieving a substantial gain over the base model (+5.24\%) and outperforming several modular frameworks.}
\label{tab:mind2web}
\end{table}

\section{Sensitivity Analysis}
\label{sec:sensitivity}

We study the robustness of BEPA to two key hyperparameters:
(i) the rollout group size $N$ used in GRPO, and
(ii) the LEVEL-2 cache update rule $\mathcal{U}$.
We report the average success rate and standard deviation across 3 runs for all experiments.

\begin{figure}[t]
    \centering
    \includegraphics[width=0.95\linewidth]{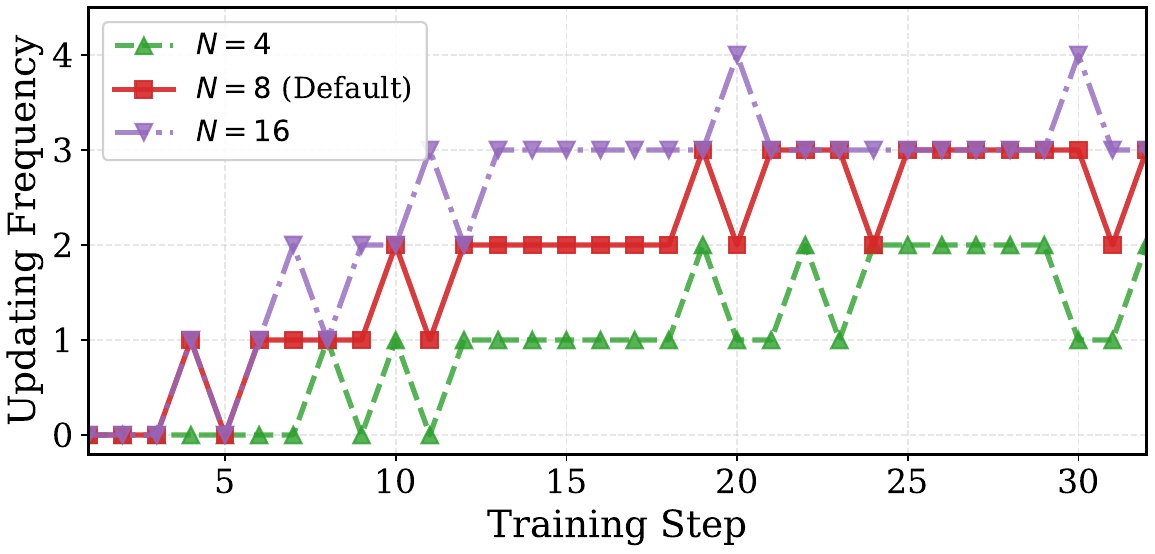}
    \caption{\textbf{Effect of rollout group size $N$ on LEVEL-2 updates.}
    We plot the per-step cache-updating frequency for different group
    sizes ($N{=}4,8,16$).
    Larger $N$ leads to more frequent cache refreshes, especially in
    later stages of training, indicating faster discovery of successful
    rollouts and more aggressive expert-to-policy assimilation.}
    \label{fig:sensitivity_updatefreq}
\end{figure}

\paragraph{Rollout Group Size $N$.}
The rollout group size $N$ controls the exploration width per task before group-wise normalization.
Table~\ref{tab:sensitivity_combined}(a) demonstrates that increasing $N$ from $4$ to $16$ monotonically improves the average success rate on OSWorld-Verified ($29.85_{\pm 1.12}\% \rightarrow 33.40_{\pm 0.58}\%$).
Larger groups increase the probability of discovering rare successful trajectories on hard tasks, which subsequently triggers more frequent LEVEL-2 cache updates (as seen in Figure~\ref{fig:sensitivity_updatefreq}).
Although $N{=}16$ yields the highest performance, we adopt $N{=}8$ ($32.13_{\pm 0.25}\%$) as the default setting to balance computational cost and final performance.

\paragraph{Cache Update Rule $\mathcal{U}$.}
We compare three LEVEL-2 cache update rules $\mathcal{U}(\mathcal{T}_x^{\mathrm{succ}})$ for selecting a new cached trajectory: \textit{Random} (default), \textit{Highest LogProb}, and \textit{Shortest Step}.
\textit{Highest LogProb} prioritizes trajectories with higher likelihood under the current policy, a heuristic often linked to better quality in self-evolving LLMs~\cite{huang2023selfimprove,huang2025rzero}, while \textit{Shortest Step} favors execution efficiency from the GUI agent's perspective.
As shown in Table~\ref{tab:sensitivity_combined}(b), while both heuristic strategies achieve slightly higher average success rates—\textit{Highest LogProb} at $33.12_{\pm 0.45}\%$ and \textit{Shortest Step} at $33.55_{\pm 0.62}\%$—compared to \textit{Random} ($32.13_{\pm 0.25}\%$), these gains are marginal and not statistically significant given the overlapping standard deviations.
This implies that the primary driver of BEPA's success is the \textbf{mechanism of dynamic updating} itself—which progressively aligns off-policy guidance with the on-policy manifold—rather than the specific criterion used to filter successes.
Since any successful rollout from the current policy already provides valid on-manifold guidance, we retain \textbf{Random} as the default $\mathcal{U}$ for its simplicity and robustness.

\begin{table}[h]
\centering
\begin{subtable}[t]{0.45\textwidth}
    \centering
    \resizebox{0.9\linewidth}{!}{
    \begin{tabular}{lc}
    \toprule
    \textbf{Group Size} & \textbf{Success Rate (\%)} \\
    \midrule
    $N=4$ & $29.85_{\pm 1.12}$ \\
    $N=8$ (Default) & $32.13_{\pm 0.25}$ \\
    $N=16$ & $\mathbf{33.40}_{\pm 0.58}$ \\
    \bottomrule
    \end{tabular}
    }
    \caption{Rollout group size $N$.}
    \label{tab:sensitivity_n}
\end{subtable}
\hfill
\begin{subtable}[t]{0.45\textwidth}
    \centering
    \resizebox{0.9\linewidth}{!}{
    \begin{tabular}{lc}
    \toprule
    \textbf{Update Rule $\mathcal{U}$} & \textbf{Success Rate (\%)} \\
    \midrule
    Shortest Step & $\mathbf{33.55}_{\pm 0.62}$ \\
    Highest LogProb & $33.12_{\pm 0.45}$ \\
    Random (Default) & $32.13_{\pm 0.25}$ \\
    \bottomrule
    \end{tabular}
    }
    \caption{Cache update rule $\mathcal{U}$.}
    \label{tab:sensitivity_mech}
\end{subtable}
\caption{\textbf{Sensitivity on OSWorld-Verified.}
Average overall success rate with standard deviation across 3 runs under different rollout group sizes $N$ (left) and LEVEL-2 cache update rules $\mathcal{U}$ (right).}
\label{tab:sensitivity_combined}
\end{table}

\section{Domain Breakdown of BEPA's Gains}
\label{sec:domain_breakdown}

\begin{figure*}[t]
    \centering
    \includegraphics[width=0.8\textwidth]{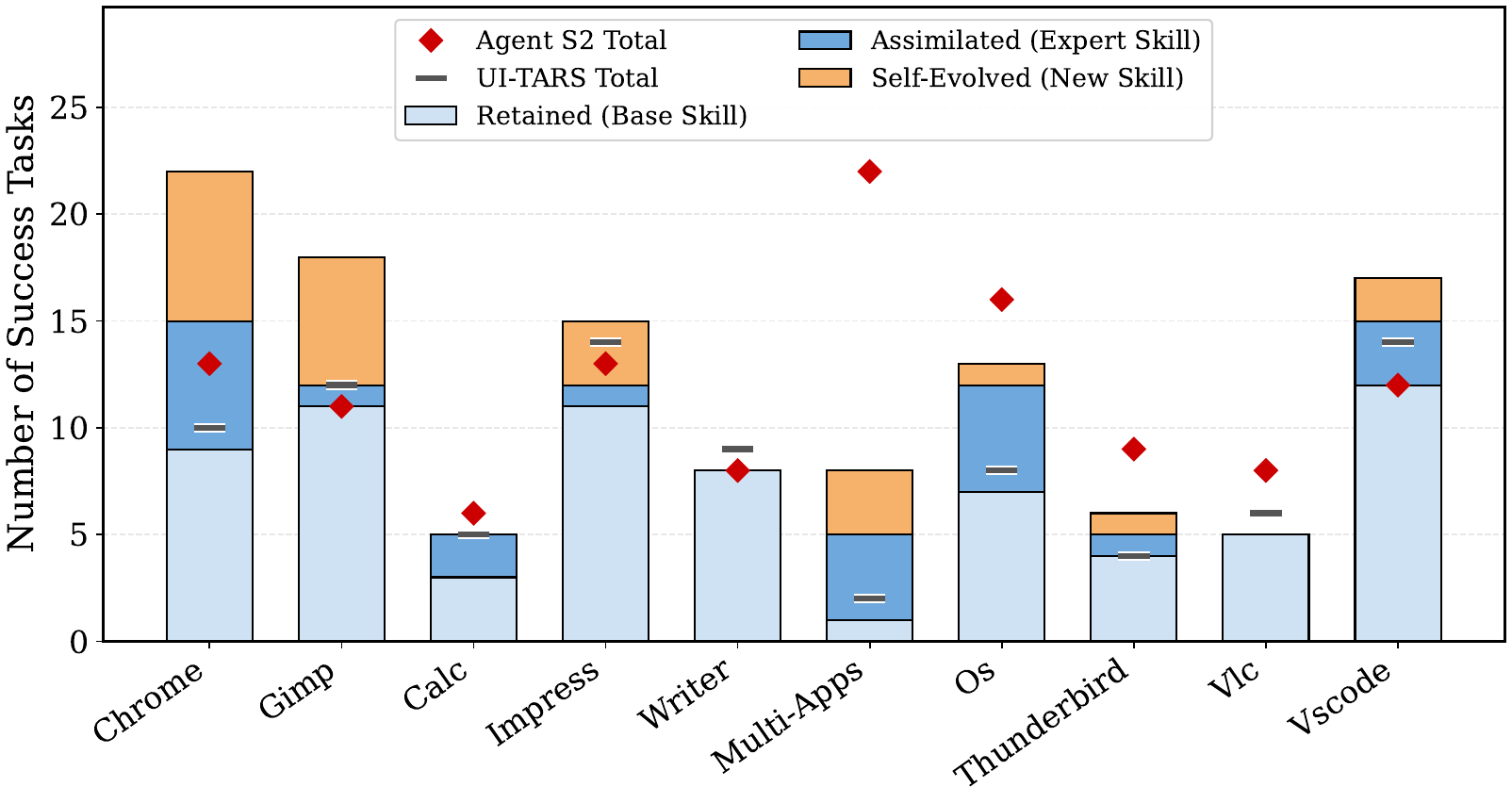}
    % Domain breakdown figure
% Domain breakdown
\caption{\textbf{Domain breakdown of BEPA's gains.}
For each domain, BEPA successes are decomposed into retained tasks (also solved by UITARS1.5-7B), assimilated tasks (solved by Agent~S2 but not by UITARS1.5-7B), and self-evolved tasks (solved only by BEPA).
Red diamonds and gray ticks denote the total successes of Agent~S2 and UITARS1.5-7B, respectively.}
    \label{fig:domain_breakdown}
\end{figure*}

To understand where BEPA improves over the base UITARS1.5-7B policy and the Agent~S2 expert, we decompose OSWorld-Verified successes by their source (\cref{fig:domain_breakdown}). For each domain, we count tasks that BEPA (i) shares with UITARS1.5-7B (retained), (ii) newly solves that are also solved by Agent~S2 but not UITARS1.5-7B (assimilated), and (iii) solves while both baselines fail (self-evolved). BEPA closes much of the gap to Agent~S2 on several domains: for example, on Chrome it solves 22 tasks versus 13 for Agent~S2 and 10 for UITARS1.5-7B, combining strong retention with substantial assimilation and a non-trivial number of self-evolved successes. In contrast, multi-application workflows remain challenging: BEPA reaches only 8 successes where Agent~S2 solves 22, suggesting that when both the base policy and expert traces are far from the target distribution, bi-level assimilation alone is insufficient.

\section{Case Studies}
\label{sec:case_studies}

\begin{figure*}[t]
    \centering
    \includegraphics[width=\textwidth]{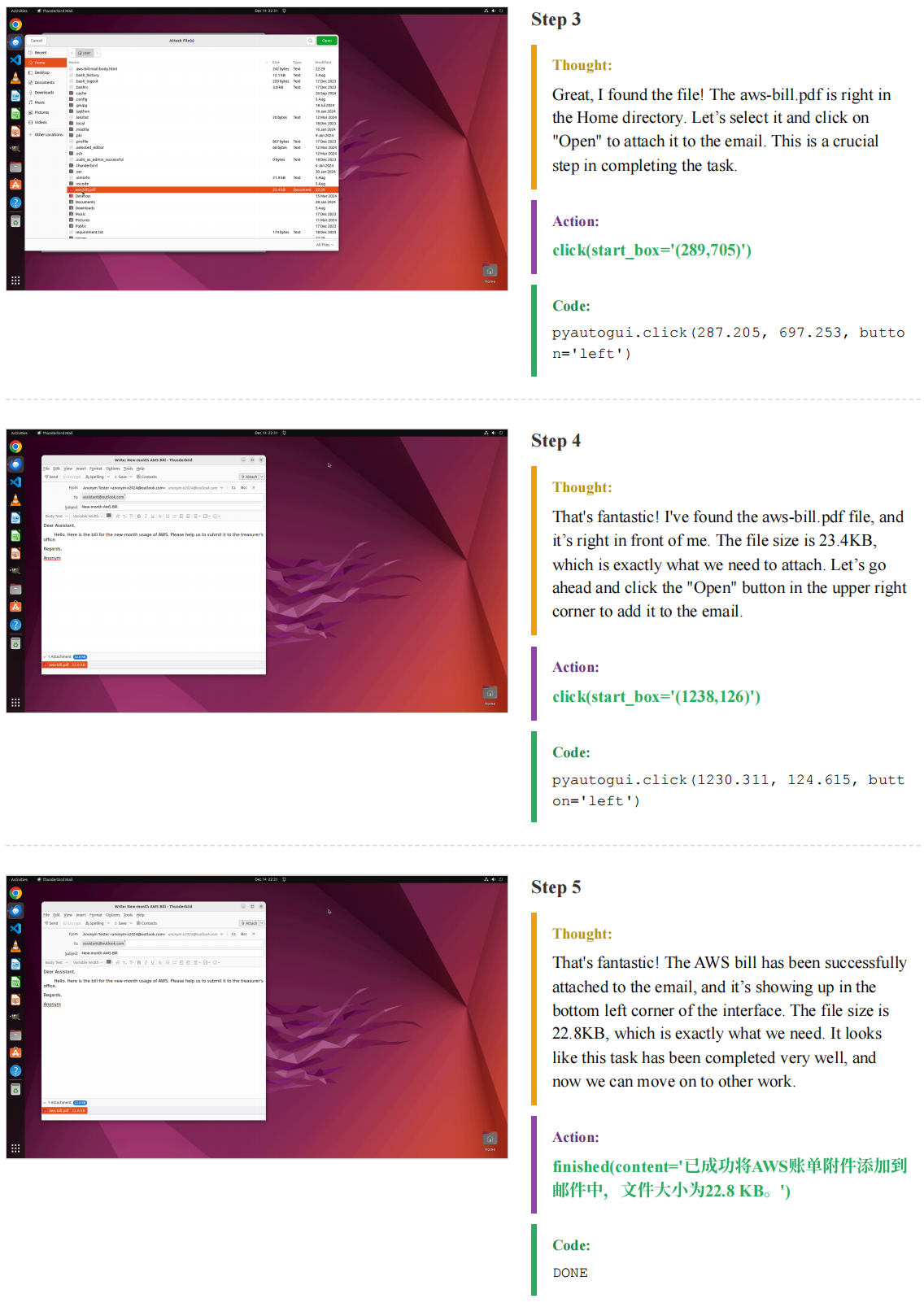}
    % GIMP case: Horizontal Flip
\caption{Case study on attaching a file in Thunderbird, comparing Agent~S2, a converted trace, and a self-rolled trace (BEPA~L1). All three succeed in five steps, but the converted trace inherits S2's hotkeys and template-like chain-of-thought, including references to an external ``plan,'' yielding actions and thoughts that are off the UITARS1.5-7B distribution. The self-rolled trace instead receives only a short expert plan and executes it via on-policy clicks and its own step-wise thoughts, producing a more learnable action and thought distribution.}

    \label{fig:distribution_case}
\end{figure*}

\paragraph{Self-rolling as Learnable Guidance.}
\Cref{fig:distribution_case} compares Agent~S2, a converted trace, and a self-rolled trace on the Thunderbird task ``Attach my test file to the email and send.'' All three complete the task in five steps, but the converted trace directly copies three expert hotkeys (Ctrl+Shift+A, Ctrl+L, Enter) together with S2-style template thoughts (e.g., ``I see~...'', ``I notice~...'',
``The task has been completed successfully as per the plan.''), where the ``plan'' refers to S2's global plan rather than UITARS1.5-7B's own reasoning. This produces actions and thoughts that are valid but lie off the policy's natural distribution. In contrast, the self-rolled trace only conditions on a short expert plan (click Attach, go to Home, select \texttt{aws-bill.pdf}, click Open) and lets the policy choose its own clicks and narrative. The resulting trajectory follows a menu-based path with concise, observation-grounded thoughts, illustrating how plan-conditioned self-rolling rewrites expert guidance into a form that is easier for the end-to-end policy to absorb.

\begin{figure*}[t]
    \centering
    \includegraphics[width=\textwidth]{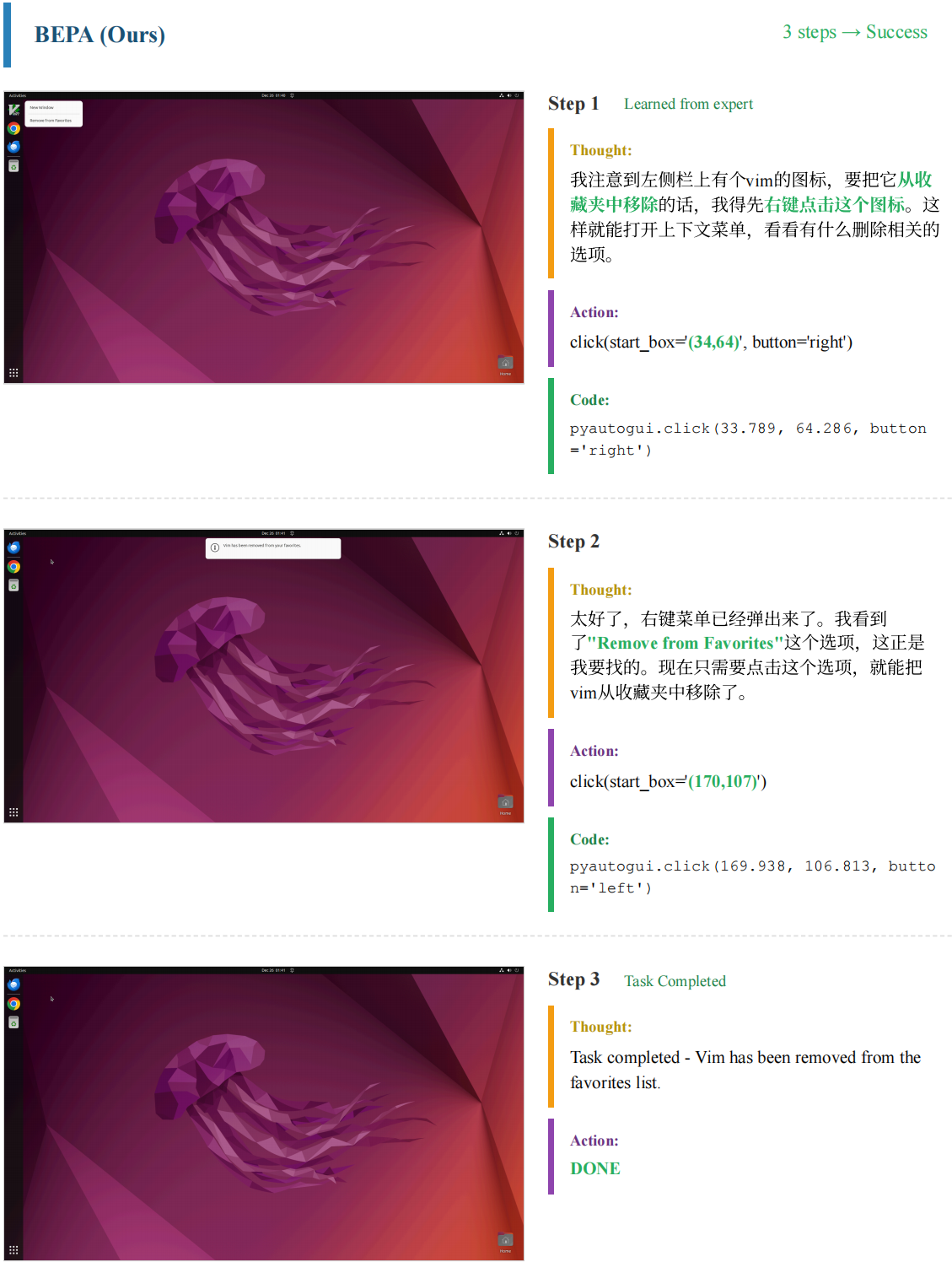}
    
% Emergence case study (Remove Vim from Favorites)
% OS case: Remove Vim from Favorites
\caption{Case study on the OS task ``Remove Vim from the favorites list.''
UITARS1.5-7B misinterprets ``favorites'' and explores irrelevant settings, while Agent~S2 and BEPA both ground it as the dock and quickly remove Vim from the favorites bar. Note that although we emphasized in the system prompt that the agent should output English, the base agent UITARS1.5-7B may still occasionally generate Chinese content due to the characteristics of UITARS's training data, etc.}
\label{fig:emergence_case}
\end{figure*}

\paragraph{Assimilation from Off-policy Traces.}
\Cref{fig:emergence_case} shows how BEPA acquires a new capability from off-policy guidance on the OS task ``Remove Vim from the favorites list.''
UITARS1.5-7B fails to interpret ``favorites'' correctly and never manipulates the dock, whereas the Agent~S2 expert solves the task by right-clicking the Vim icon in the dock and selecting ``Remove from Favorites.''
After training with BEPA, the end-to-end policy solves the same task by following a similar high-level strategy, even though it could not solve it before.
This example illustrates LEVEL-2’s role in turning successful expert trajectories into lasting new skills beyond the base policy.

\clearpage

\begin{figure*}[t]
    \centering
    \includegraphics[width=\textwidth]{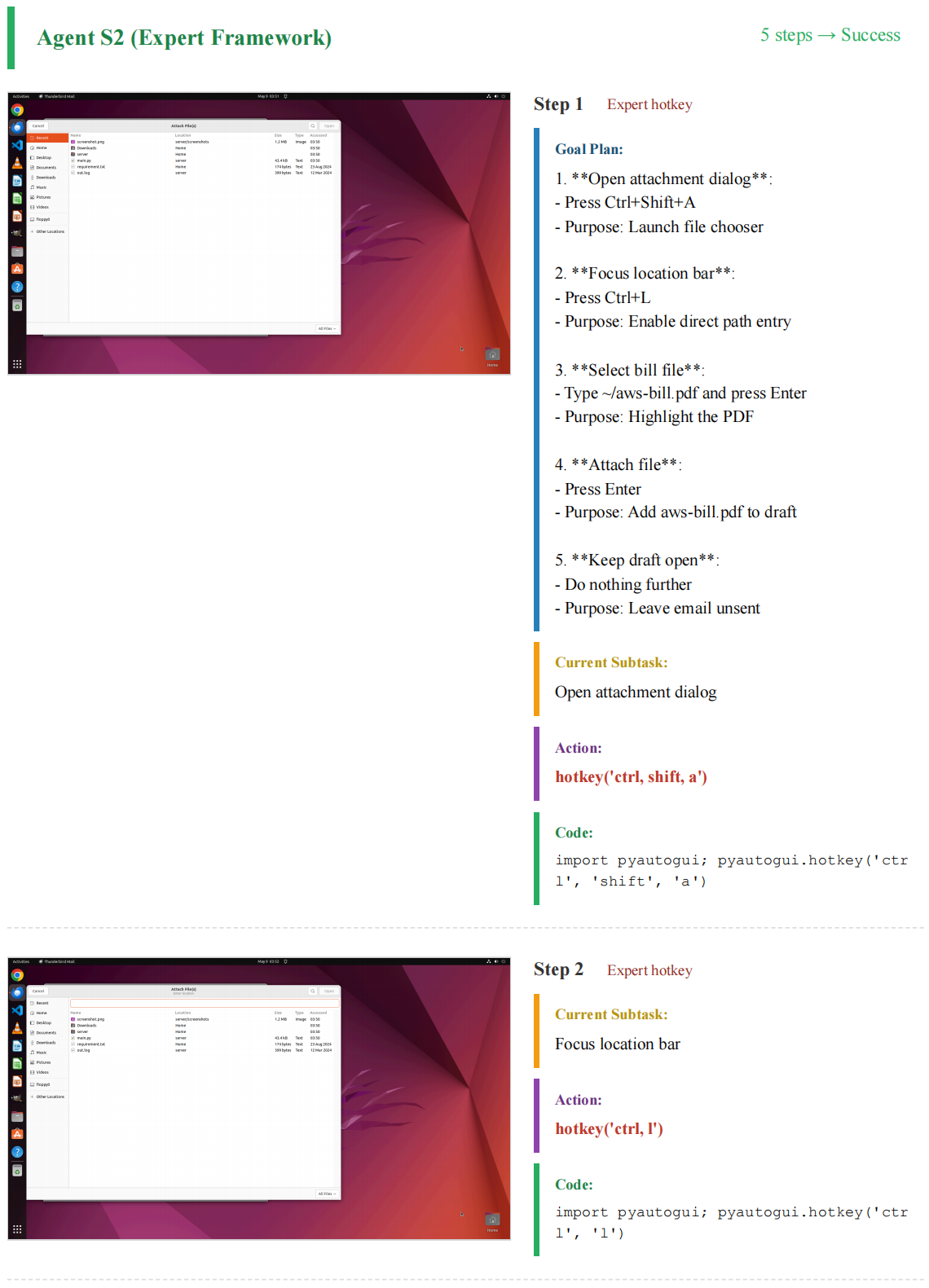}
\end{figure*}
\begin{figure*}[t]
    \centering
    \includegraphics[width=\textwidth]{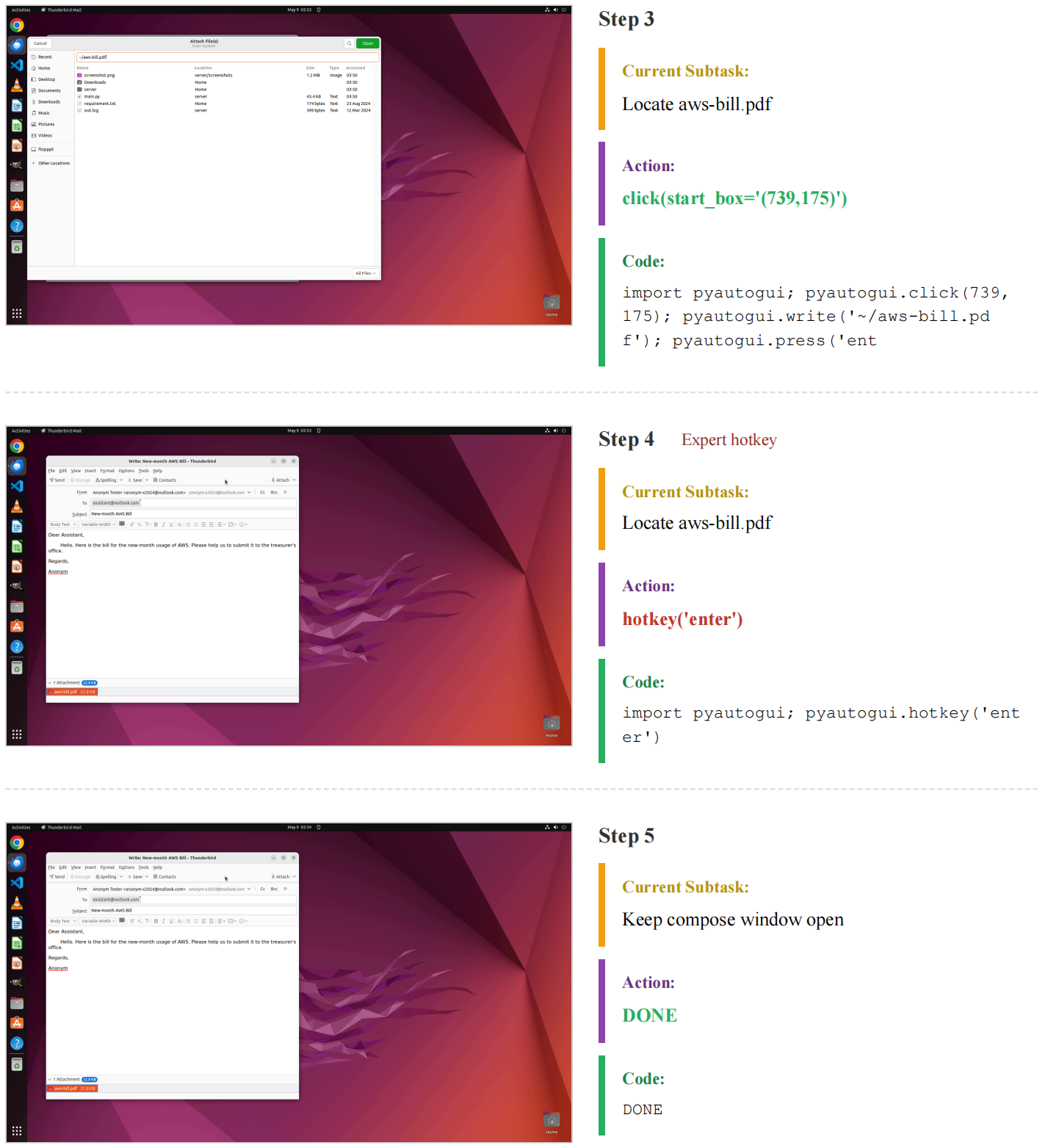}
\end{figure*}
\begin{figure*}[t]
    \centering
    \includegraphics[width=\textwidth]{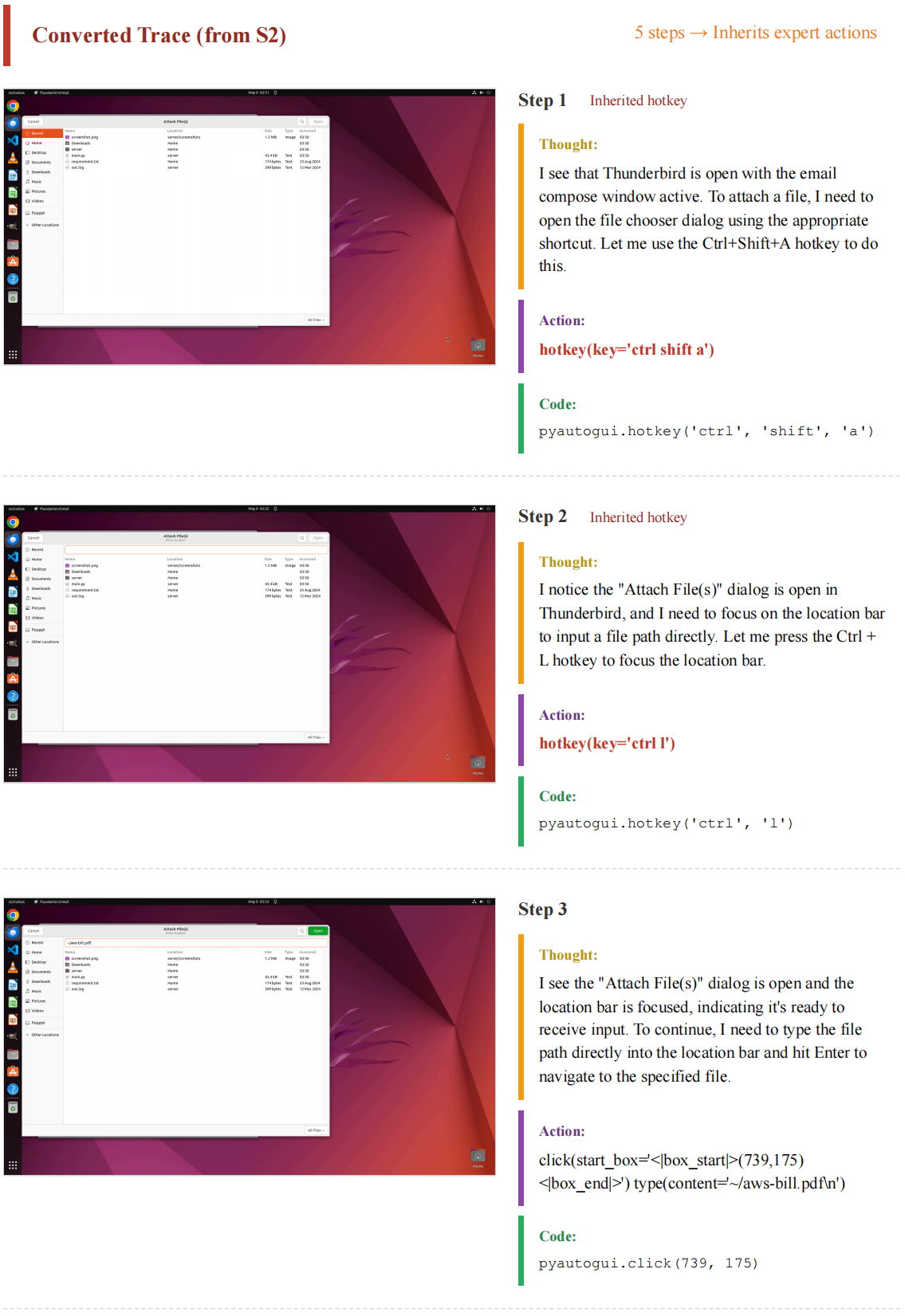}
\end{figure*}
\begin{figure*}[t]
    \centering
    \includegraphics[width=\textwidth]{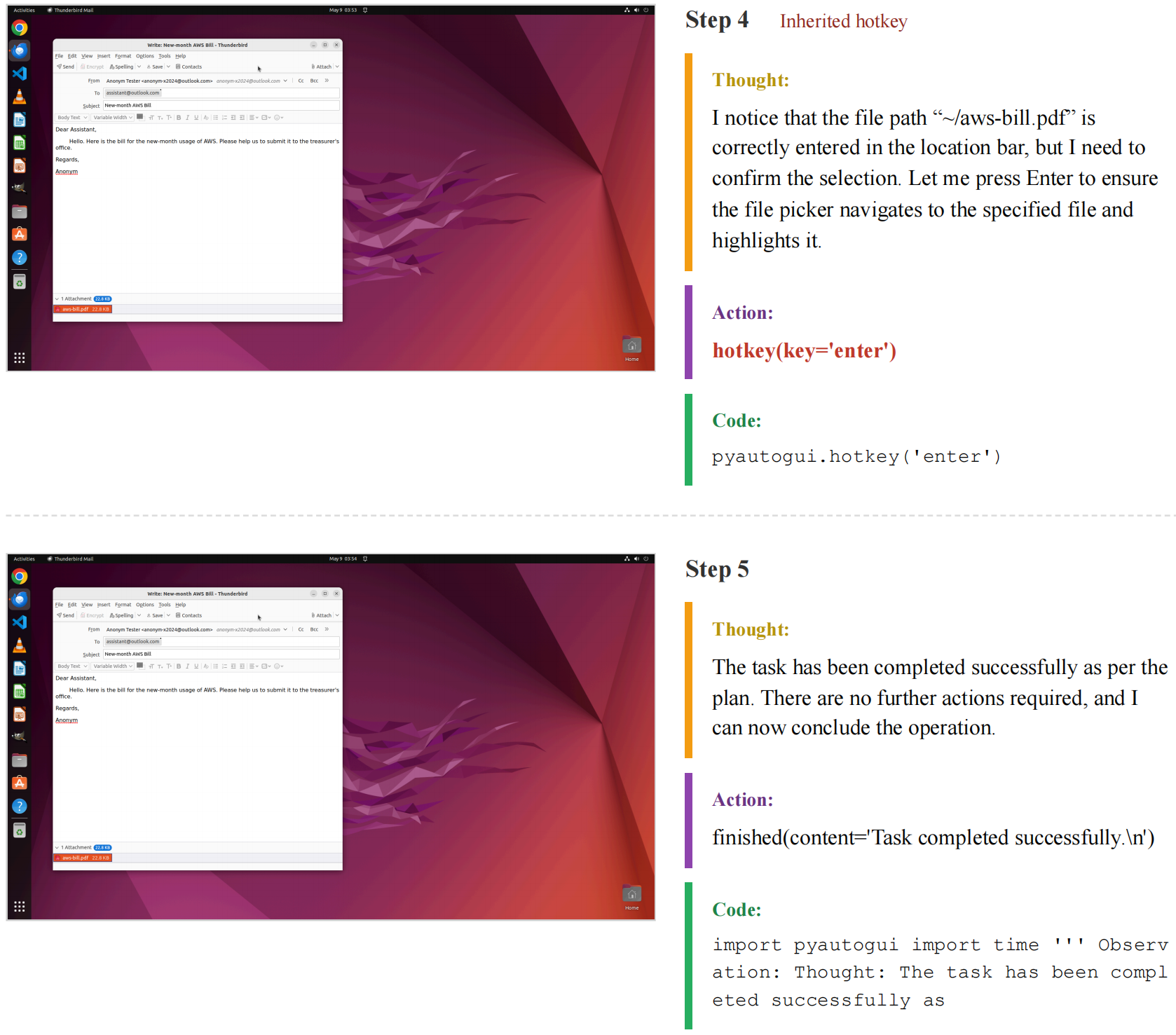}
\end{figure*}
\begin{figure*}[t]
    \centering
    \includegraphics[width=\textwidth]{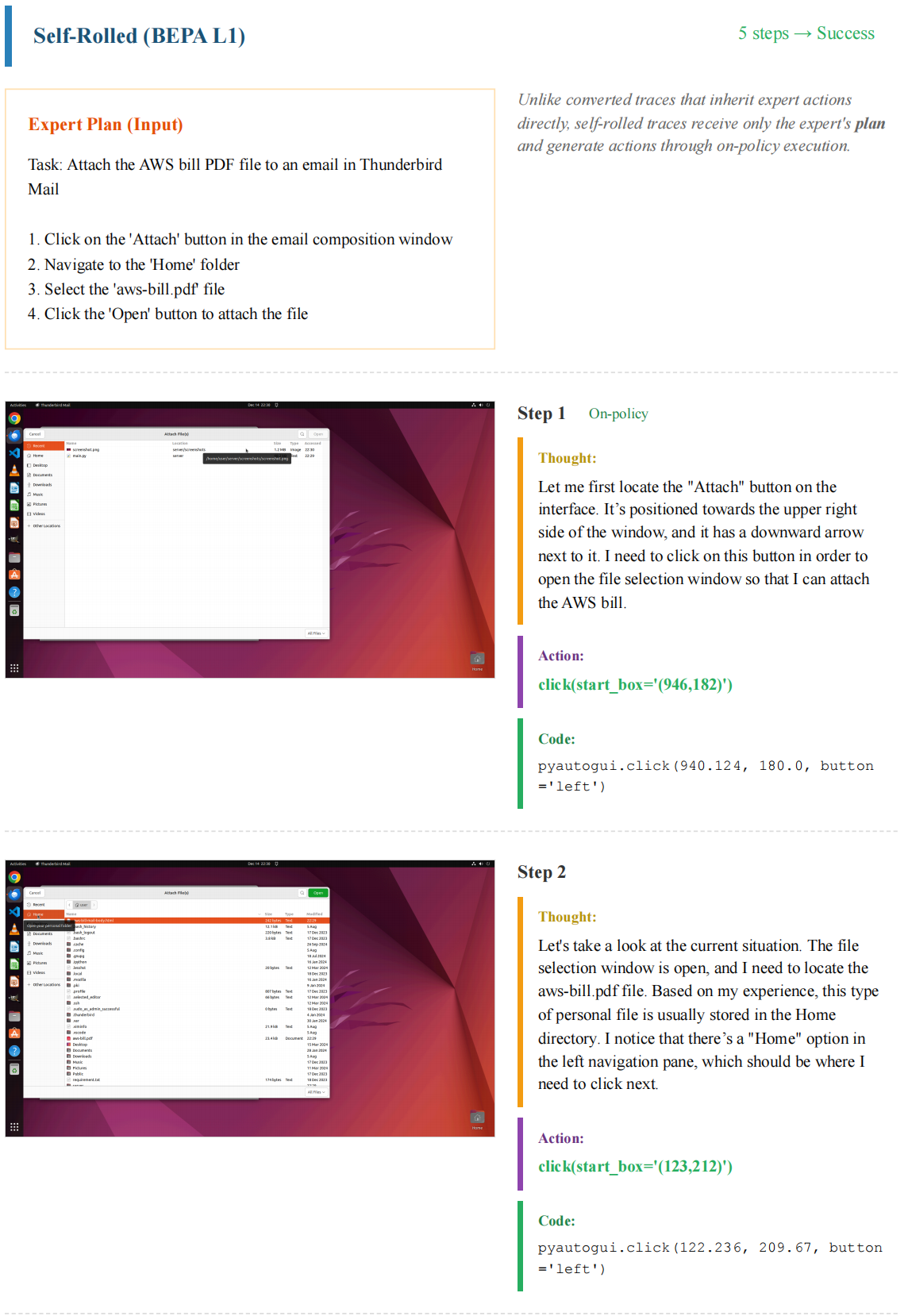}
\end{figure*}

\begin{figure*}[t]
    \centering
    \includegraphics[width=\textwidth]{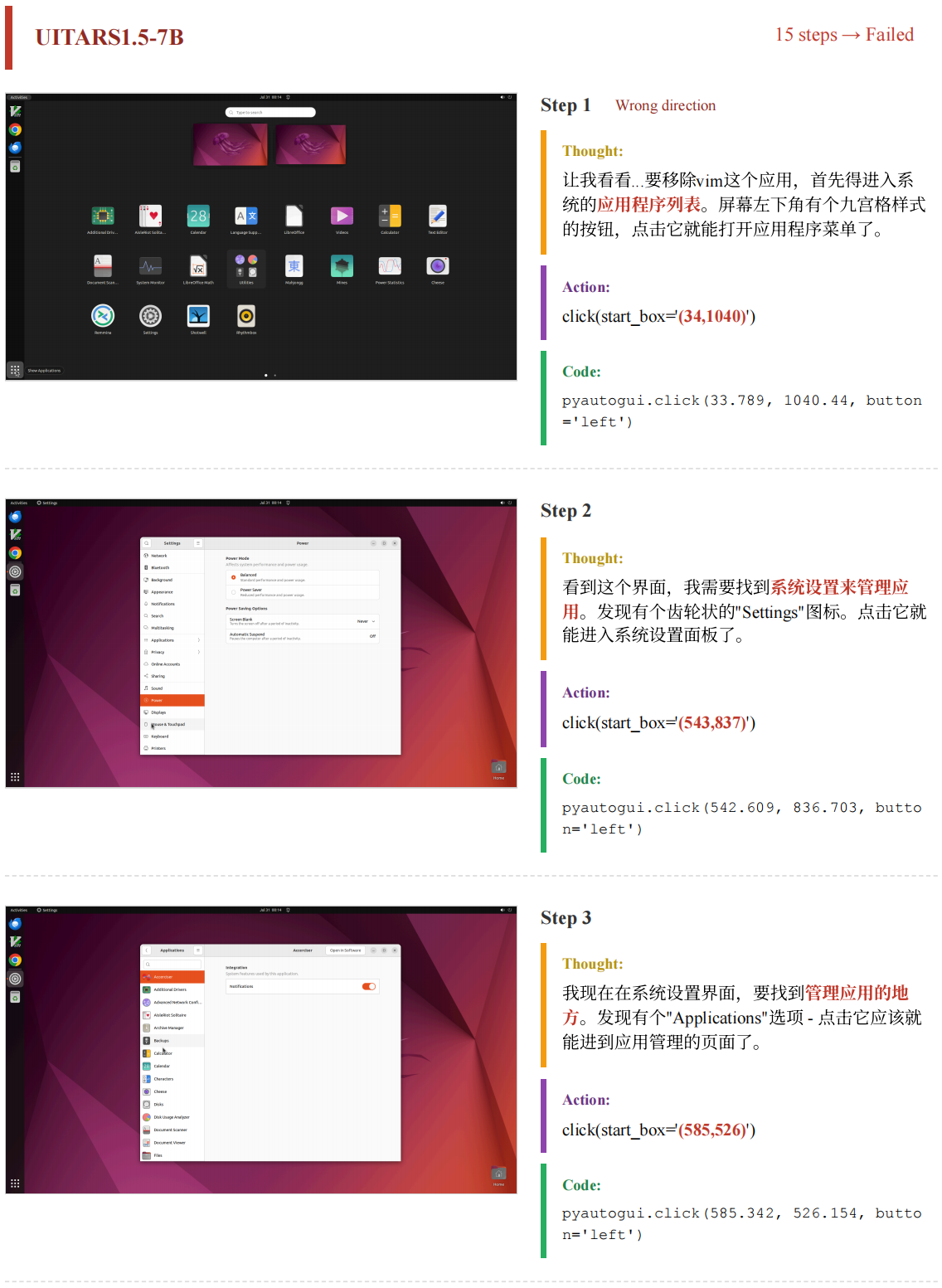}
\end{figure*}
\begin{figure*}[t]
    \centering
    \includegraphics[width=\textwidth]{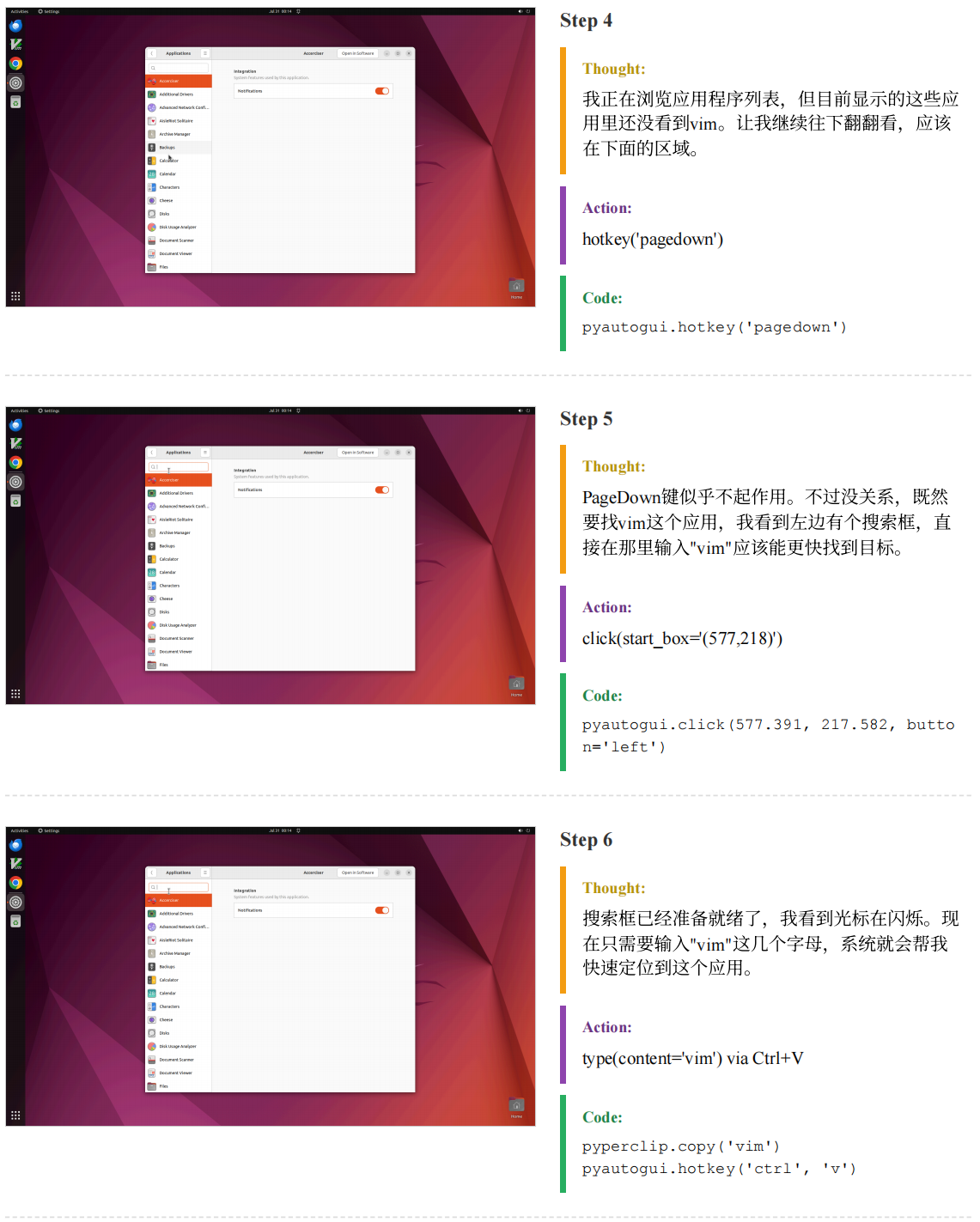}
\end{figure*}
\begin{figure*}[t]
    \centering
    \includegraphics[width=\textwidth]{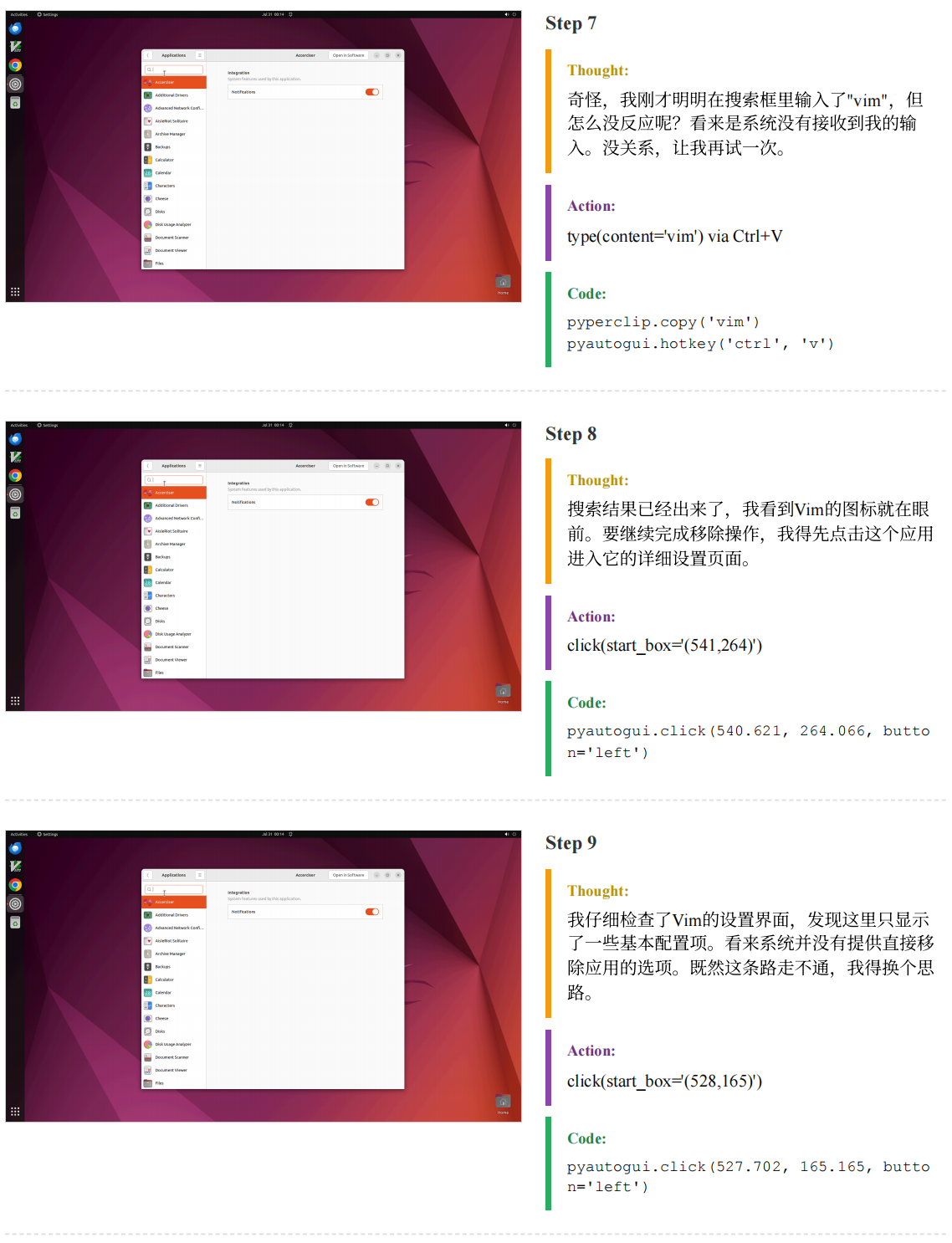}
\end{figure*}
\begin{figure*}[t]
    \centering
    \includegraphics[width=\textwidth]{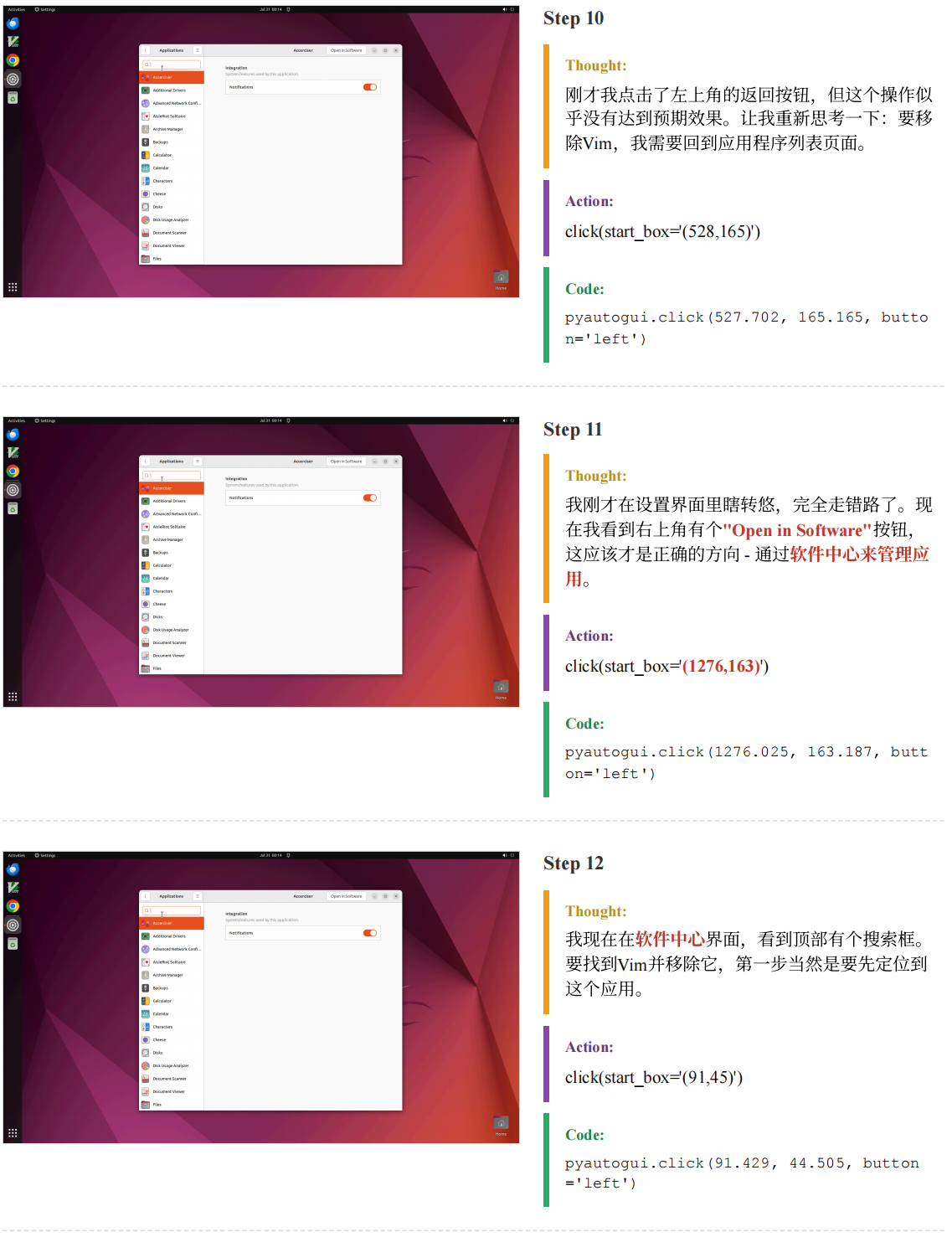}
\end{figure*}
\begin{figure*}[t]
    \centering
    \includegraphics[width=\textwidth]{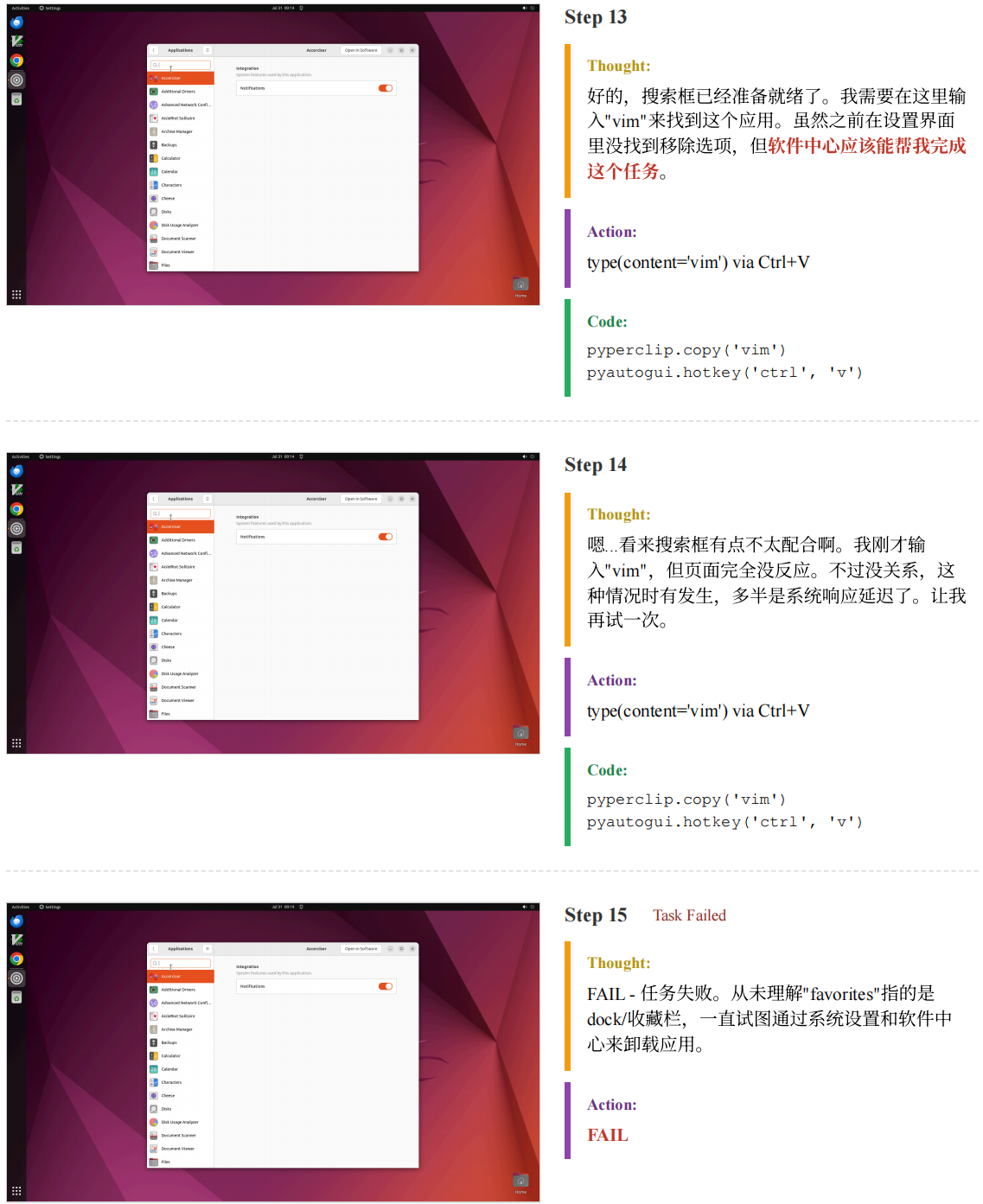}
\end{figure*}
\begin{figure*}[t]
    \centering
    \includegraphics[width=\textwidth]{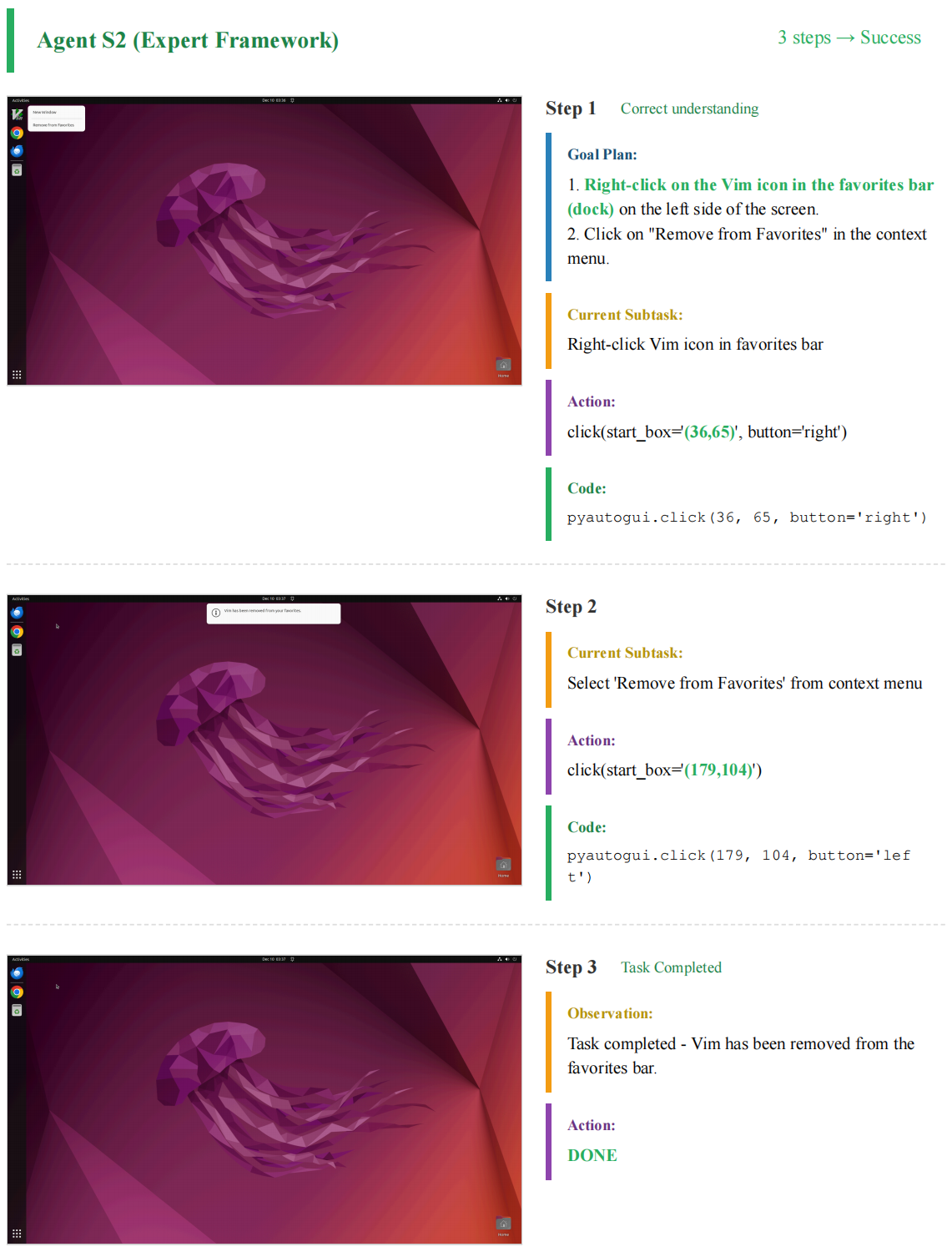}
\end{figure*}

% \input{body/checklist}
%%%%%%%%%%%%%%%%%%%%%%%%%%%%%%%%%%%%%%%%%%%%%%%%%%%%%%%%%%%%

\end{document}